\documentclass[lettersize,journal]{IEEEtran}
\usepackage{amsmath,amsfonts}
\usepackage{algorithmic}
\usepackage{algorithm}
\usepackage{array}
\usepackage[T1]{fontenc}
\usepackage{textcomp}
\usepackage{stfloats}
\usepackage{url}
\usepackage{verbatim}
\usepackage{graphicx}
\usepackage{cite}
\usepackage{color}
\usepackage{multirow}
\usepackage{booktabs} 
\usepackage{color,xcolor}
\usepackage{bbm}
\usepackage{pifont}
\usepackage{subcaption}
\usepackage{bm}
\usepackage{makecell}
\usepackage{hyperref}
 
\begin{document}

\title{GSStream: 3D Gaussian Splatting based Volumetric Scene Streaming System}

\author{
        Zhiye~Tang,
        Qiudan~Zhang,~\IEEEmembership{Member,~IEEE,}
        Lei~Zhang, ~\IEEEmembership{Member,~IEEE,}
        Junhui~Hou, ~\IEEEmembership{Senior Member,~IEEE,}
        You Yang, ~\IEEEmembership{Senior Member,~IEEE,}
        Xu Wang,~\IEEEmembership{Member,~IEEE}
\thanks{
This work was supported in part by the National Natural Science Foundation of China under Grant 62371310, in part by the Shenzhen Science and Technology Program (JCYJ20241202124415021), in part by the Guangdong Basic and Applied Basic Research Foundation under Grant 2023A1515011236. \textit{(Corresponding author: Xu Wang.)}

Zhiye Tang, Qiudan Zhang, Lei Zhang and Xu Wang are with the College of Computer Science and Software Engineering, Shenzhen University, Shenzhen, 518060, China. Email: (tangzhiye2022@foxmail.com, qiudanzhang@szu.edu.cn, leizhang@szu.edu.cn, wangxu@szu.edu.cn).

Junhui Hou is with Department of Computer Science, City University of Hong Kong, Kowloon, Hong Kong SAR, China. Email: (jh.hou@cityu.edu.hk).

You Yang is with School of Electronic Information and Communications, Huazhong University of Science and Technology, Wuhan 430074, China. Email: (yangyou@hust.edu.cn).

}
}

\markboth{}%
{Shell \MakeLowercase{\textit{et al.}}: A Sample Article Using IEEEtran.cls for IEEE Journals}


\maketitle

\begin{abstract}
Recently, the 3D Gaussian splatting (3DGS) technique for real-time radiance field rendering has revolutionized the field of volumetric scene representation, providing users with an immersive experience. But in return, it also poses a large amount of data volume, which is extremely bandwidth-intensive. Cutting-edge researchers have tried to introduce different approaches and construct multiple variants for 3DGS to obtain a more compact scene representation, but it is still challenging for real-time distribution. In this paper, we propose GSStream, a novel volumetric scene streaming system to support 3DGS data format. Specifically, GSStream integrates a collaborative viewport prediction module to better predict users' future behaviors by learning collaborative priors and historical priors from multiple users and users' viewport sequences and a deep reinforcement learning (DRL)-based bitrate adaptation module to tackle the state and action space variability challenge of the bitrate adaptation problem, achieving efficient volumetric scene delivery. Besides, we first build a user viewport trajectory dataset for volumetric scenes to support the training and streaming simulation. Extensive experiments prove that our proposed GSStream system outperforms existing representative volumetric scene streaming systems in visual quality and network usage. Demo video: \url{https://youtu.be/3WEe8PN8yvA}.
\end{abstract}

\section{Introduction}

Lately, 3DGS has significantly transformed the landscape of volumetric scene representation, enabling the achievement of exceptionally high-fidelity real-time rendering~\cite{kerbl20233d}. 3DGS explicitly models a volumetric scene through the aggregation of numerous 3D Gaussians, offering enhanced flexibility and detail compared to traditional representations. However, this substantial advantage in rendering quality is accompanied by a notable increase in data volume, which poses challenges for practical applications, particularly in scenarios requiring efficient storage and transmission~\cite{10870258}. To mitigate this issue, extensive research efforts have been directed towards optimizing various aspects of the 3DGS framework. In a recent advancement, Lu \textit{et al.} introduced Scaffold-GS~\cite{lu2024scaffold}, which innovatively employs learnable anchors instead of directly training on individual 3D Gaussians. In this approach, each anchor is initially derived from a structure-from-motion (SfM) point cloud and aligned to a voxel grid, establishing a structured foundation for the scene representation. During both the training and rendering phases, these anchors dynamically generate multiple surrounding neural 3D Gaussians on the fly through the use of multi-layer perceptrons (MLPs). This process results in significant memory savings, as it reduces the need to store vast numbers of individual Gaussians. For example, representing an indoor scene measuring 3 meters by 8 meters in high quality using Scaffold-GS requires only a few hundred megabytes of storage, which is a considerable improvement over previous methods. Despite these advancements, the storage requirements still present a substantial obstacle for streaming such high-quality volumetric scenes under the constraints of current network infrastructure, necessitating further optimization and innovation in streaming technologies.

\begin{figure}
    \subcaptionbox{Original}{
        \includegraphics[width=0.45\linewidth]{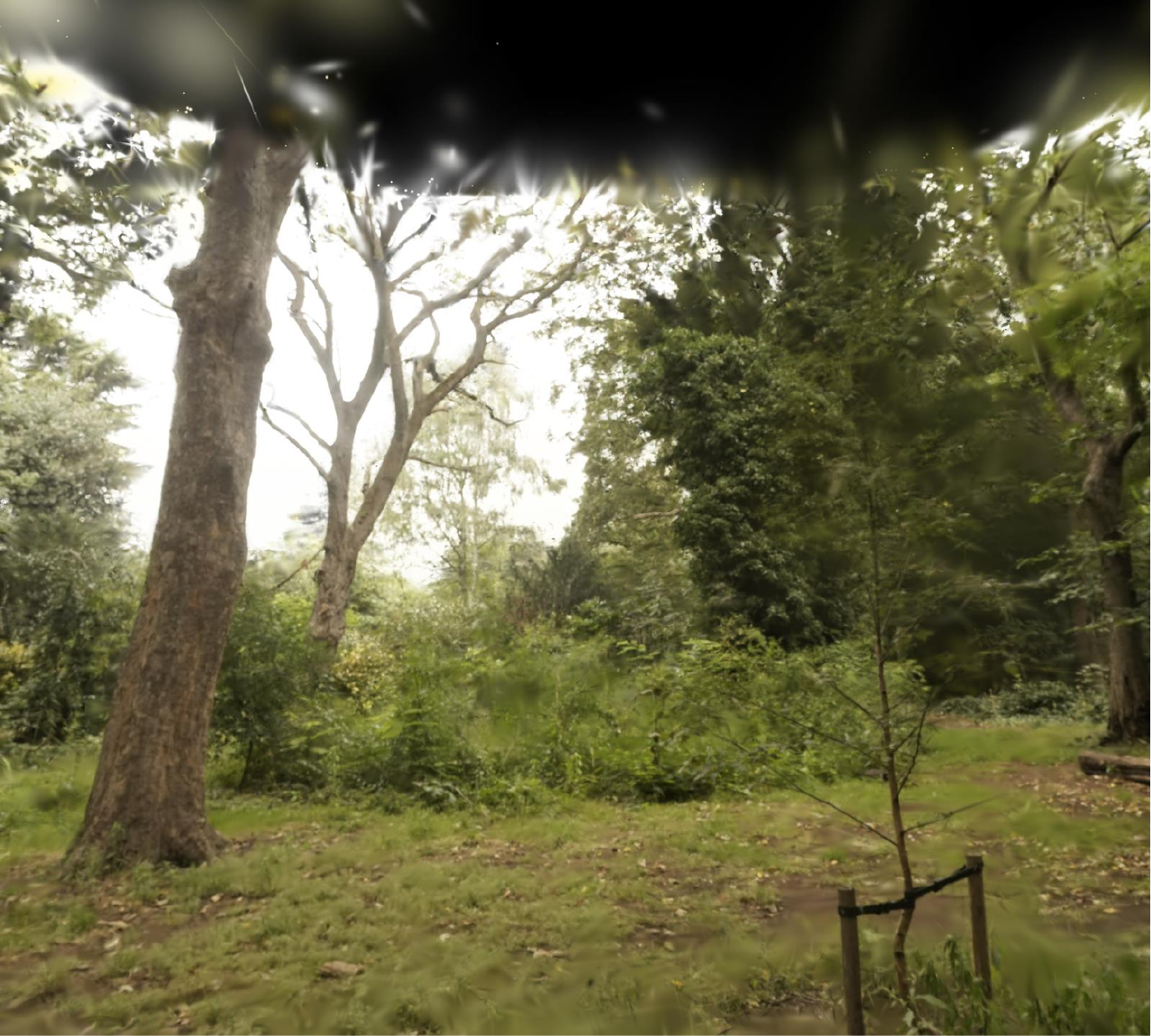}
    }
    \subcaptionbox{CaV3~\cite{liu2023cav3}}{
        \includegraphics[width=0.45\linewidth]{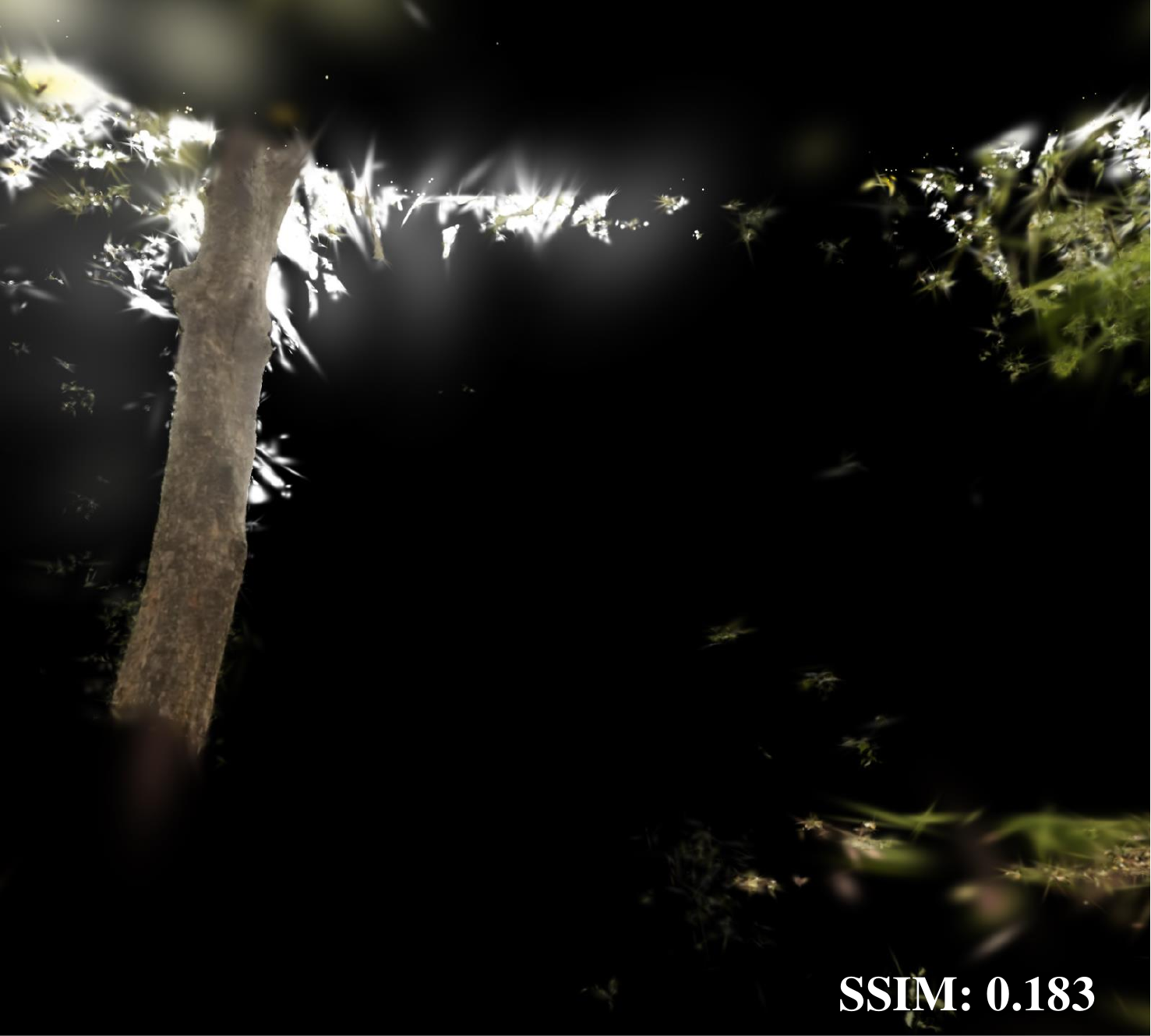}
    }\\
    \subcaptionbox{GS3D~\cite{kellogg2023gaussiansplats3d}}{
        \includegraphics[width=0.45\linewidth]{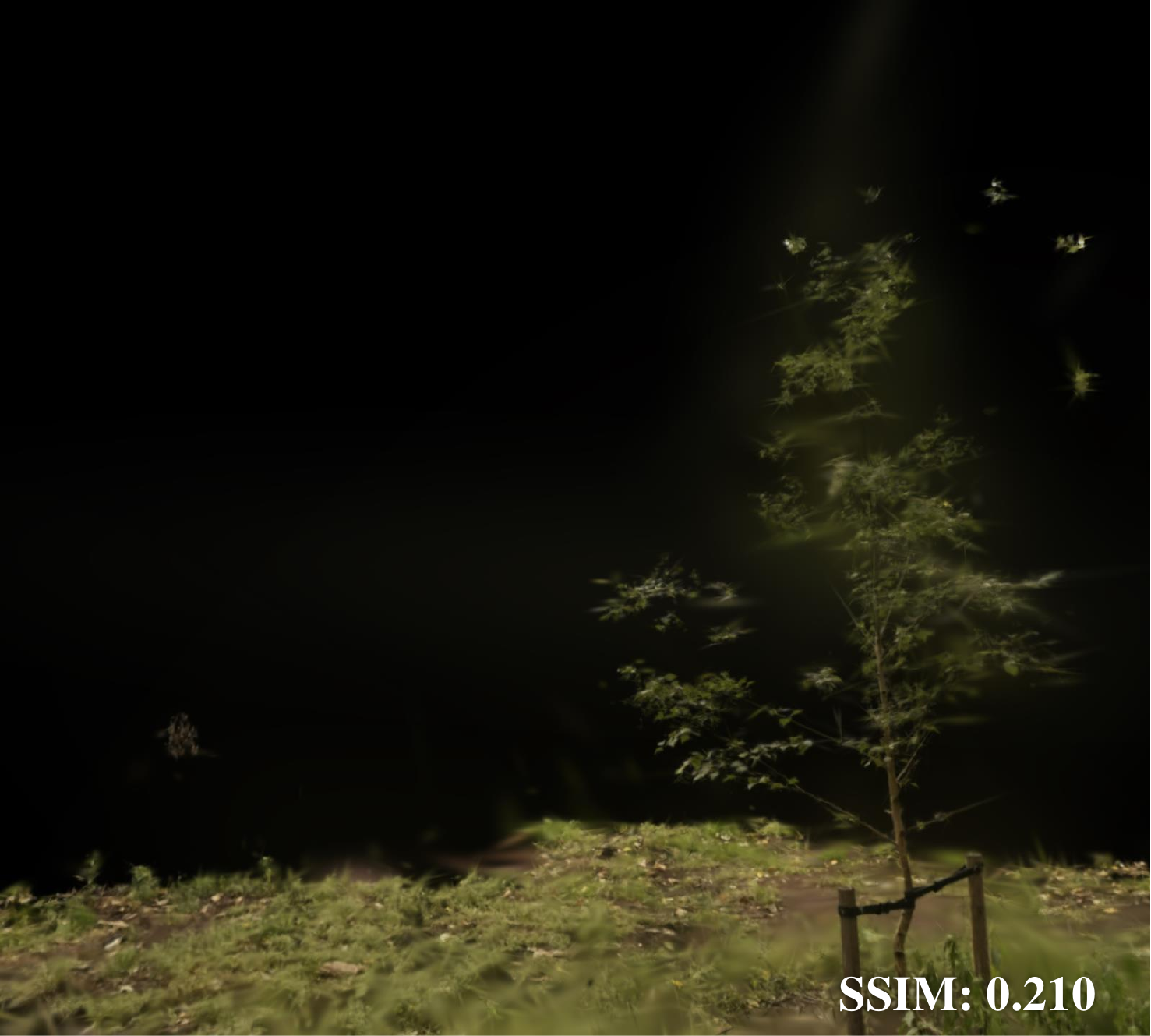}
    }
    \subcaptionbox{Ours}{
        \includegraphics[width=0.45\linewidth]{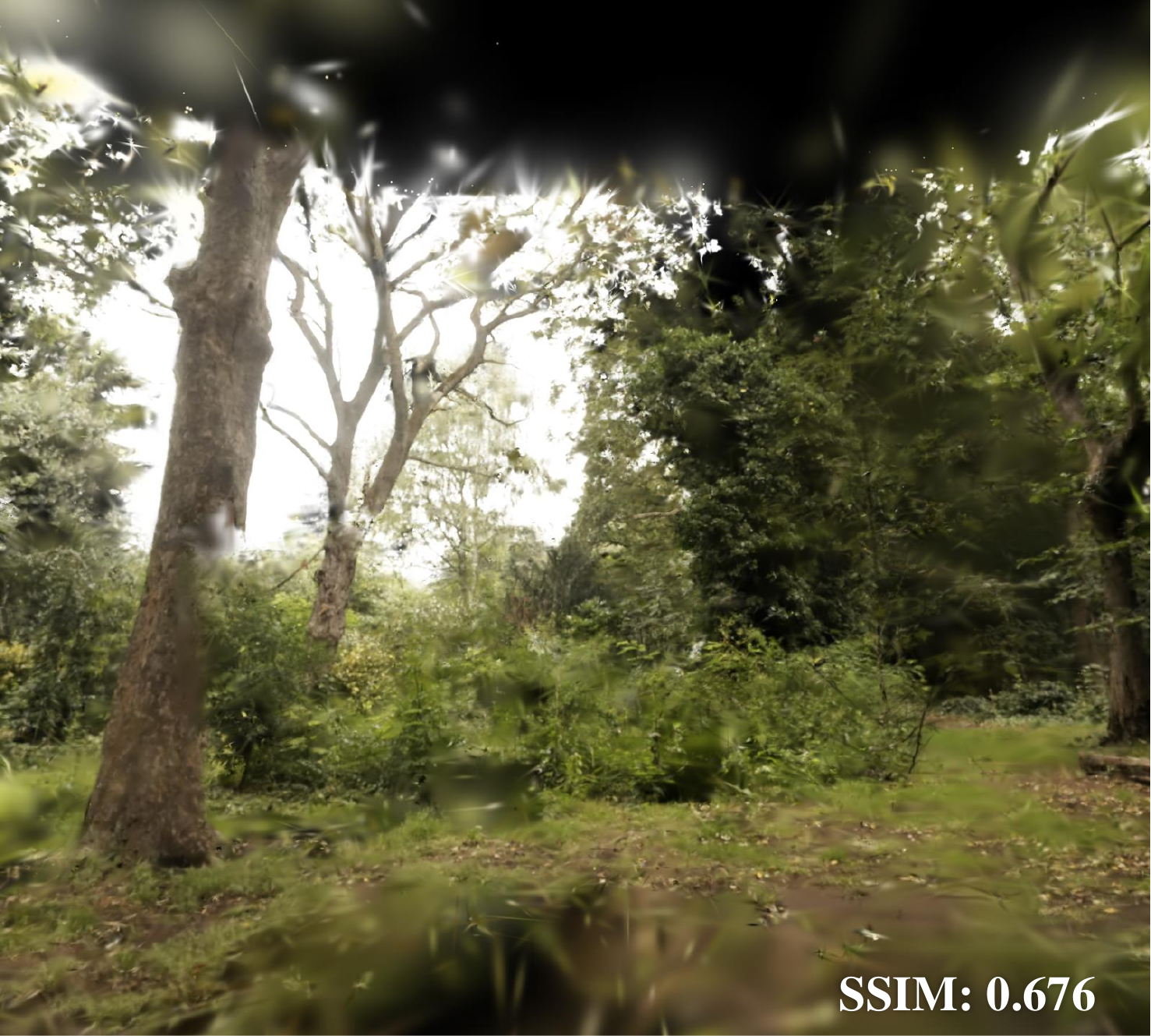}
    }
    \caption{Visualization effect of different volumetric scene streaming systems. Our proposed GSStream achieves the best visual performance among the state-of-the-art (SOTA) systems with the help of the collaborative viewport prediction module and the DRL-based bitrate adaptation module. Note that these examples are rendered on the \textit{stump} scene under a network constraint of 120Mbps, at the 10th second (300th frame) of the viewing process.}
    \label{fig:examples}
\end{figure}

From the perspective of data formatting, 3DGS-derived representations (\textit{e.g.}, Scaffold-GS) can be viewed as an evolution of traditional point clouds. A conventional point cloud typically represents a 3D scene as a collection of discrete points, where each point encompasses attributes such as 3D spatial coordinates $(x,y,z)$, color information, and normal vectors, among others~\cite{schwarz2018emerging}. Similarly, Scaffold-GS organizes its data by storing anchors as clusters of vectors, with each vector encapsulating both feature and geometric attributes, including center coordinates and scaling coefficients. This similarity in storage structure implies that established point cloud scene streaming methodologies are inherently compatible with 3DGS scenes, allowing for the adaptation of existing streaming techniques to the 3DGS framework. Nonetheless, the challenge of efficiently streaming point cloud content has been a focal point of extensive research, exploring multiple dimensions and strategies. The prevailing approach in this domain involves the simultaneous application of viewport prediction and tile-based bitrate adaptive streaming to minimize the volume of data transmitted~\cite{han2020vivo}. In practice, this involves partitioning a volumetric scene into a grid of non-overlapping tiles, each of which is individually encoded or downsampled into multiple representations corresponding to different quality levels. Rather than transmitting all tiles at the highest quality, a bitrate adaptation strategy is employed to selectively determine the quality level of each tile based on its relevance to the user's current and predicted future viewports. For instance, Zhang \textit{et al.}~\cite{zhang2022trans} proposed Trans-RL and adopted a sequence prediction paradigm where a single Transformer model is utilized to predict the user’s viewport trajectories. Subsequently, DRL is introduced to address the bitrate adaptation challenge, determining the optimal quality level for each tile to balance visual fidelity and bandwidth usage. However, a significant limitation of this approach is the assumption that all volumetric scenes are divided into an identical number of tiles, which is rarely the case in practical applications. Consequently, existing methods encounter three primary challenges: (1) As a novel form of volumetric scene representation, there is a dearth of comprehensive user viewport trajectory datasets specifically tailored for 3DGS scenes, hindering the development of effective viewport prediction models; (2) Current viewport prediction techniques often treat viewport trajectories across different users in a uniform manner, failing to account for individual differences in viewing behavior and personal preferences, which can lead to suboptimal prediction accuracy; and (3) The full potential of DRL algorithms in addressing the bitrate adaptation problem has yet to be realized, as existing approaches have not fully exploited the capabilities of DRL to dynamically and intelligently manage bitrate allocation based on varying network conditions and user behaviors.

In this paper, we introduce \textbf{GSStream}, a comprehensive 3D \textbf{G}aussian \textbf{S}platting-based volumetric scene \textbf{Stream}ing system designed to tackle the aforementioned challenges head-on. The GSStream system harnesses the capabilities of Scaffold-GS~\cite{lu2024scaffold} to represent volumetric scenes, thereby enhancing the visual quality experienced by users through head-mounted devices (HMDs). Recognizing that users exhibit diverse behavioral inertia when engaging with splatted Gaussians, GSStream incorporates a collaborative viewport prediction algorithm. This algorithm not only predicts user viewports but also extracts and utilizes user-specific embeddings derived from the collected dataset in a collaborative manner, thereby leveraging individual viewing patterns to improve prediction accuracy and relevance. Additionally, GSStream addresses the variability in the number of tiles across different volumetric scenes by treating the tiles decomposed from each scene as an unordered set. To effectively manage bitrate adaptation under these conditions, GSStream models the problem as a Markov decision process (MDP)~\cite{sutton2018reinforcement} and employs a deep deterministic policy gradient (DDPG)-based~\cite{silver2014deterministic,lillicrap2015continuous} algorithm to derive optimal bitrate allocation policies. For validation, we have constructed a realistic user viewport trajectory dataset tailored for 3DGS content, encompassing 15 diverse volumetric scenes that include both indoor and outdoor scenarios. We developed a specialized user viewport data collection platform to gather behavioral data from 32 subjects representing various industries, ensuring a broad and representative dataset. Our experimental results demonstrate that the proposed GSStream system outperforms existing SOTA volumetric scene streaming systems in terms of both visual quality and network efficiency, underscoring the effectiveness of our integrated approach in enhancing the user experience while optimizing resource utilization.

The contributions of this work can be summarized as follows:
\begin{itemize}
    \item We present the first 3DGS-based volumetric scene streaming system, \textbf{GSStream}, which seamlessly integrates a \textbf{C}ollaborative \textbf{V}iewport \textbf{P}rediction (CVP) module with a \textbf{D}RL-based \textbf{B}itrate \textbf{A}daptation (DBA) module. This integration allows for a more accurate estimation of users' future viewing behaviors and effectively addresses the challenges posed by variability in state and action spaces inherent in the bitrate adaptation problem.
    \item We develop the first comprehensive user viewport trajectory dataset specifically for 3DGS content, encompassing 15 diverse volumetric scenes that include a range of indoor and outdoor scenarios. This dataset was constructed by designing and implementing a user viewport data collection platform, which was utilized to capture detailed behavioral data from 32 participants across different industries, thereby ensuring the dataset's diversity and applicability.
    \item Through rigorous experimental evaluations, we demonstrate that our proposed GSStream system significantly enhances visual quality and optimizes network usage compared to existing SOTA volumetric scene streaming systems. These results validate the efficacy of our system in delivering superior performance in real-world streaming scenarios.
\end{itemize}

\section{Related Works}

\subsection{3D Scene Representations and Rendering}

There have been various explorations on representing volumetric scenes, \textit{e.g.}, mesh, point cloud. Point cloud format is the most popular among them since it is flexible, similar, and easy for reconstruction~\cite{liu2021point}. Its adaptability to capture fine geometric details, as well as its straightforward generation process from common 3D scanning or multi-view reconstruction pipelines, makes the point-based format a fundamental choice for many computer vision tasks. 
Therefore, the topic of point-based rendering has long been discussed since 1985~\cite{levoy1985use}, in which points are used as primitives for detailed rendering. 
Modern methods usually replace points with surfels estimated based on point cloud data~\cite{pfister2000surfels} and use splatting operations to rasterize and combine multiple overlapping surfels either linearly or using neural networks~\cite{bui2018point,meshry2019neural} to obtain the final rendering result. Such approaches efficiently handle geometry and appearance while maintaining a level of simplicity that benefits many real-time applications.

With the emergence of differentiable rendering techniques, neural radiance field (NeRF)~\cite{mildenhall2021nerf} and 3DGS~\cite{kerbl20233d} have both attracted increasing interest. NeRFs are neural network models trained to predict novel views in 3D scenes according to certain given views, effectively learning a volumetric representation that captures scene geometry and appearance. Each pre-trained model can be seen as an implicit representation of one certain scene, in which color and density are output by a network given a spatial coordinate and viewing direction. Recently, 3DGS has been notable for its pure explicit representation and real-time rendering. It proposed using 3D Gaussians as rendering primitives, maintains the storage structure as a point cloud, and has brought new possibilities to the field of volumetric scene representation. 
This explicit formulation enables direct manipulation of the scene’s geometry without relying on a learned neural model, offering users more interpretability and flexibility when editing or processing 3D data while facilitating real-time rendering.

Starting from a SfM point cloud, the position $\mu$ of each point $i$ in a 3DGS is treated as the position of a 3D Gaussian $G_i$:
\begin{equation}
    G_i(x)=e^{-\frac{1}{2}(x-\mu_i)^T\Sigma_i^{-1}(x-\mu_i)},
\end{equation}
where $x$ is a random position within a 3D space (\textit{i.e.}, domain of definition), while $\Sigma_i$ denotes the covariance matrix describing the pose and scale of the 3D Gaussian, formulated by a scaling matrix $S_i$ and a rotation matrix $R_i$ as
\begin{equation}
    \Sigma_i=R_iS_iS_i^TR_i^T,
\end{equation}
ensuring $\Sigma_i$ is a positive semi-definite matrix. During training supervised by image datasets, 3DGS learns an opacity value $\alpha_i$ and a group of Spherical Harmonics coefficients for each 3D Gaussian $G_i$ to model its color. The abovementioned variables will then be rasterized onto a 2D plane by employing $\alpha$-blending to ensure high-fidelity rendering. Following the framework proposed by~\cite{kerbl20233d}, Scaffold-GS~\cite{lu2024scaffold} introduced a more storage-efficient anchor-based framework by replacing 3D Gaussians as anchors. Specifically, each anchor $i$ consists of a location $\bm{x}_i\in\mathbb{R}^3$ and attributes $\mathcal{A}_i=\left\{\bm{f}_i\in\mathbb{R}^{D^f},\bm{s}_i\in\mathbb{R}^{D^s},\bm{o}_i\in\mathbb{R}^{D^o}\right\}$, with each representing anchor feature, scaling and offsets respectively. In practical rendering, $\bm{f}_i$ is inputted into $\text{MLP}_g$ to generate a group of neural Gaussians surrounding anchor $i$, whose shapes and locations are determined by anchor location $\bm{x}_i$, scaling $\bm{s}_i$ and offsets $\bm{o}_i$. Such a design demonstrates how volumetric primitives can be learned and manipulated in a way that is both memory-efficient and highly adaptable to diverse scene structures.

Compared with the raw point cloud format, the 3DGS format is significantly different. With extra attributes, 3DGS can provide rendered images with substantially higher fidelity and is streaming-friendly, with far-reaching application prospects. By explicitly encoding information within 3D Gaussians, the pipeline enables a wider range of scene manipulations, such as more precise control over geometry and appearance, as well as the potential for advanced lighting or material effects. 
However, such an advantage comes at the expense of a larger data volume. In traditional point clouds, each point is typically described by spatial coordinates, color, and normal vector, occupying 27 bytes of memory~\cite{schwarz2018emerging}. 3DGS represents a volumetric scene as a set of 3D Gaussians, with attribute information such as center coordinates and covariance matrix, thus consuming 248 bytes per point. For a typical indoor scene with millions of 3D Gaussians, the data volume can be up to 800MB, which is challenging for efficient delivery under current network conditions. Such memory demands can be particularly prohibitive in scenarios where fast, real-time updates or remote streaming of high-fidelity 3D content is necessary.

To this end, some researchers have also focused on developing compact representations for 3DGS. 
In~\cite{lee2023compact}, Lee \textit{et al.} designed a learnable mask strategy to reduce the amount of 3D points in a 3DGS scene, greatly restricting the data volume. Morgenstern \textit{et al.}~\cite{morgenstern2023compact} proposed a highly parallel algorithm to arrange Gaussian parameters into 2D grids for a more compact representation, thereby enabling more efficient memory usage and potentially faster data transfer. Sun \textit{et al.}~\cite{sun20243dgstream}, from another perspective, proposed 3DGStream, utilizing a neural transformation cache to model the dynamic information of objects in 3DGS scenes, which is a streaming-friendly dynamic 3DGS representation. This method reduces the bandwidth required to continuously update dynamic scenes, offering a pathway to practical applications such as virtual reality or telepresence systems. 
This paper mainly focuses on streaming optimization using the anchor-based Scaffold-GS format, aiming to balance storage efficiency and rendering quality for real-time and remote 3D experiences.

\begin{figure*}
    \centering
    \includegraphics[width=\textwidth]{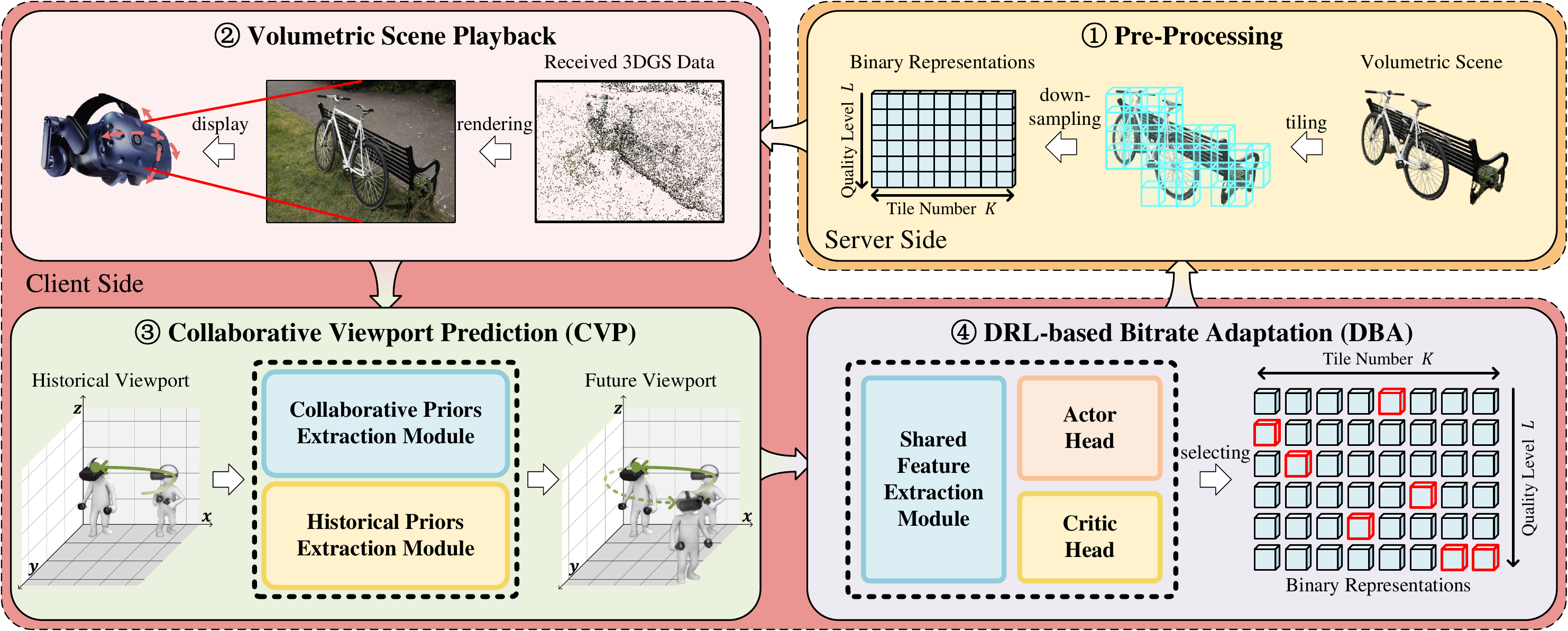}
    \caption{Overview of the proposed GSStream system. \ding{172} In the pre-processing phase, the volumetric scene is tiled into $K$ tiles, and each is downsampled into representations in $L$ quality levels. \ding{173} Representations are selectively transmitted to the client side to reconstruct a 3DGS scene for display. \ding{174} User's viewports captured by HMD are utilized to predict future viewports. \ding{175} The predicted future viewports are then utilized to generate the representation selection strategy for the next time slot.}
    \label{fig:overview}
\end{figure*}

\subsection{Volumetric Scene Streaming}

To efficiently transmit volumetric scenes, a common approach is to segment the scene content into multiple tiles, as a viewer typically only focuses on the region of the point cloud scene that lies within the field of view (FoV). By dividing the scene into these manageable chunks, it becomes more flexible to address diverse network conditions and personalized user requirements. The decomposed tiles can be independently encoded or downsampled and then selectively transmitted according to various conditions, such as the user's dynamically changing viewports, the inherent characteristics of each tile (\textit{e.g.}, size, volume or number of Gaussians), and \textit{etc.} to optimize the user's quality of experience (QoE) under stringent bandwidth constraints. By adopting this selective transmission approach, streaming systems can effectively avoid transmitting unnecessary data outside a viewer's FoV, thus reducing bandwidth consumption and promoting more efficient resource utilization.

Numerous attempts have been made to stream volumetric data. These efforts can be broadly classified into two categories. The first category is dedicated to building more efficient volumetric content delivery strategies based on simulated viewports, which help researchers anticipate user behavior under various usage scenarios or network conditions. For instance, Van der Hooft \textit{et al.}~\cite{van2019towards} modified dynamic adaptive streaming over HTTP (DASH)~\cite{stockhammer2011dynamic} for volumetric video streaming and incorporated bitrate adaptation for different objects, effectively separating foreground and background content. In~\cite{park2019rate,wang2022qoe}, researchers created tile-wise metrics based on quality levels and user viewports for each representation, then applied greedy algorithms to select the most suitable representation that balanced bandwidth demands and visual quality. Building on this idea, Li \textit{et al.} treated point cloud streaming as a time-bound optimization problem by using branch-and-bound techniques in~\cite{li2020joint}, and later in~\cite{li2022optimal} they developed another hybrid visual saliency streaming method to further improve the perceptual quality. More recently,~\cite{10129903,10816615} proposed DRL-empowered approaches for advanced point cloud content delivery.

Another development direction is the design of comprehensive streaming systems that integrate viewport prediction and bitrate adaptation into one framework. For example, ViVo~\cite{han2020vivo} employed linear regression for viewport prediction and introduced a streaming approach that considers tile distance, occlusion, and visibility, thereby reducing the wasteful transmission of low-priority regions. More recently, CaV3~\cite{liu2023cav3} leveraged Transformers~\cite{vaswani2017attention} to fuse priors learned from historical user behavior patterns and volumetric scene representations, aiming to better predict users' future viewports. In tandem with this prediction method, they also introduced the approach from Yusuf \textit{et al.}~\cite{yusuf2020cache}, which formulated the volumetric scene streaming problem as a cache-assisted multi-armed bandit (MAB)~\cite{slivkins2019introduction} problem and tackled it with a reinforcement learning algorithm. In~\cite{zhang2022trans}, Zhang \textit{et al.} proposed Trans-RL, which forecasted user viewports via a Transformer and formulated the bitrate adaptation problem as an MDP. Despite these advances, certain assumptions, such as each scene having the same number of tiles, simplify the problem space and leave unresolved issues regarding variable state and action spaces.

In conclusion, among the mainstream volumetric scene streaming methods, the user viewport information and the potential of DRL are not fully leveraged. Existing schemes either rely on simplistic viewport assumptions or use constrained strategies that do not adequately account for the dynamics of user interaction in immersive environments. To this end, we propose \textbf{GSStream}, a novel 3DGS-based volumetric scene streaming system that can efficiently transmit volumetric media content while delivering superior performance and better QoE compared to other state-of-the-art systems.

\section{System Design}

As shown in Figure~\ref{fig:overview}, our proposed GSStream system can be decomposed into four decoupled parts, each of which addresses a distinct stage in the volumetric streaming pipeline while working cooperatively to ensure high performance and user immersion. During pre-processing, the volumetric scenes are meticulously subdivided into non-overlapping cubic tiles, resulting in downsampled representations at $L$ different quality levels. When the user watches volumetric scenes via an HMD, the instant viewport information is collected and routed to the CVP module, which aggregates these real-time viewing parameters and prepares them for prediction. GSStream predicts the user's future viewports by leveraging both the historical viewport sequence and learned user-specific embeddings. The predicted viewport is then used to generate a representation selection strategy, ensuring that each cubic tile is streamed at a suitable quality level aligned with the user's immediate and near-future needs. Afterward, the server-side transmits the selected representations in accordance with the prediction, thereby achieving progressive streaming while optimizing network usage and preserving an immersive viewing experience.

\begin{figure}
    \centering
    \subcaptionbox*{Subject ID 05}{
        \includegraphics[width=0.1\textwidth]{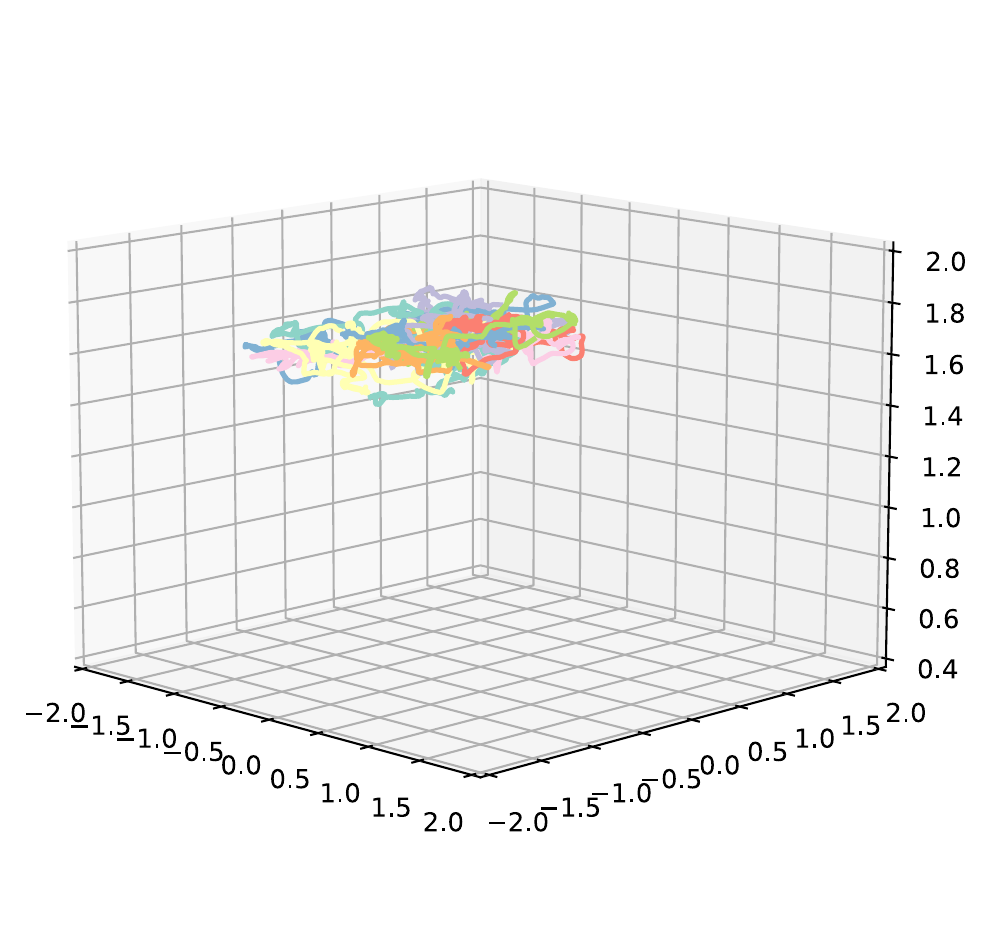}
        \includegraphics[width=0.1\textwidth]{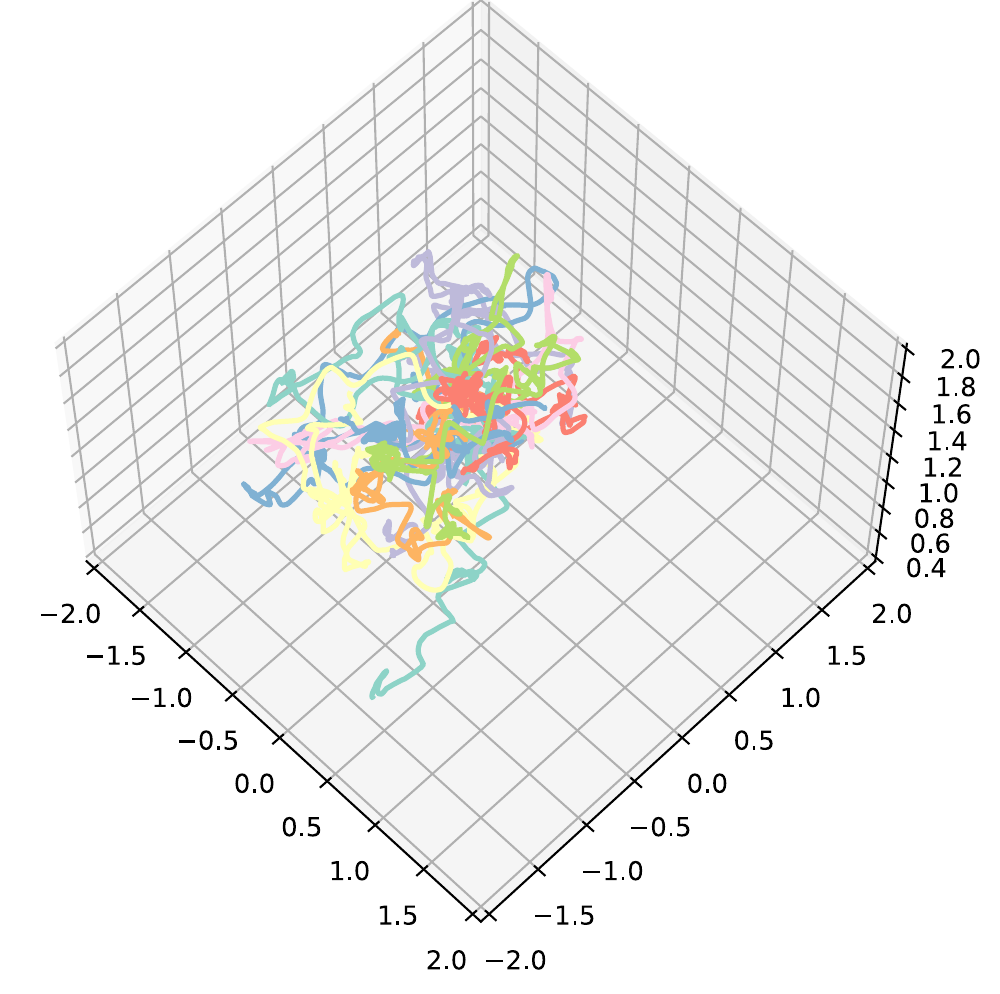}
    }
    \subcaptionbox*{Subject ID 07}{
        \includegraphics[width=0.1\textwidth]{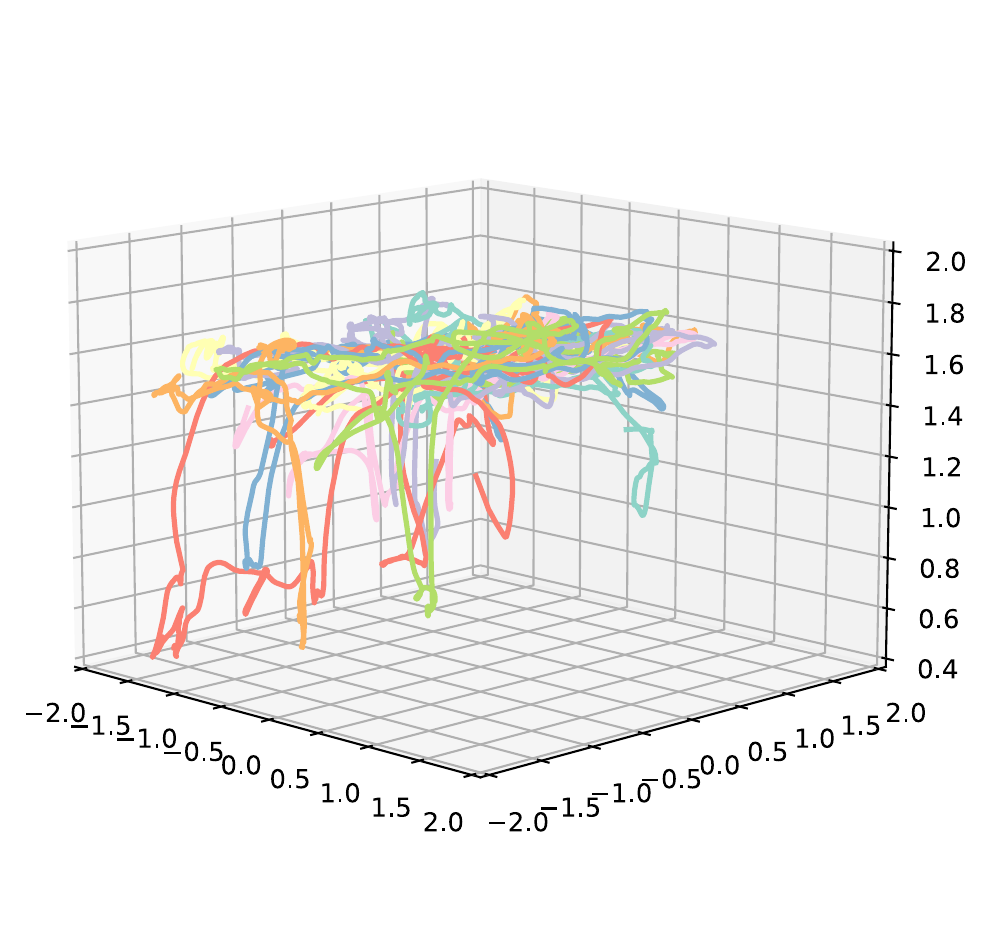}
        \includegraphics[width=0.1\textwidth]{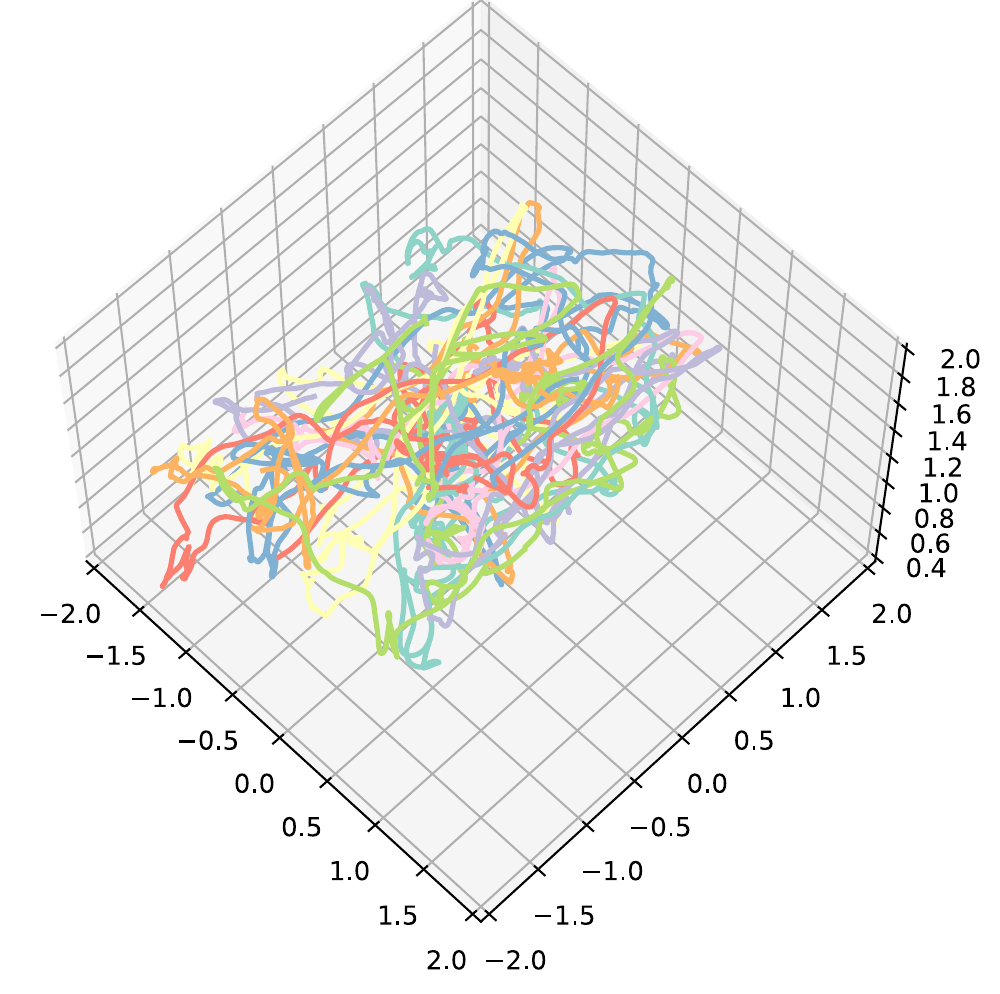}
    }
    \subcaptionbox*{Subject ID 16}{
        \includegraphics[width=0.1\textwidth]{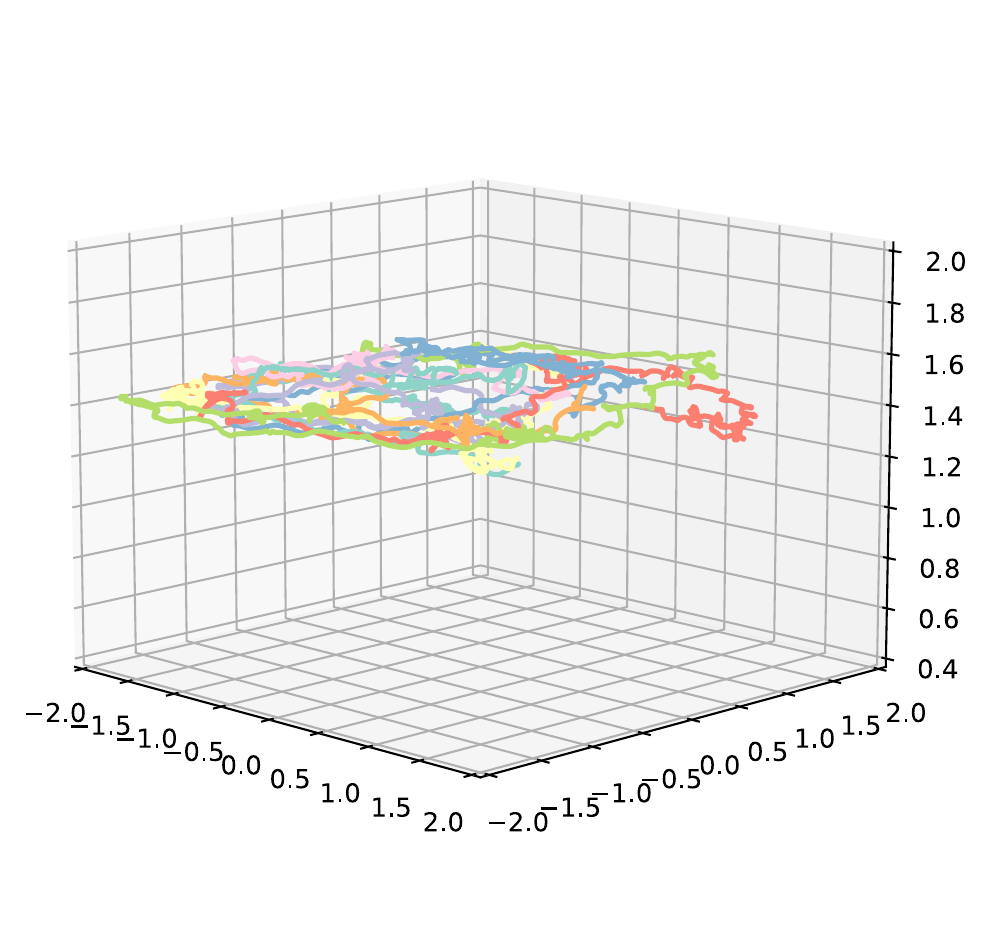}
        \includegraphics[width=0.1\textwidth]{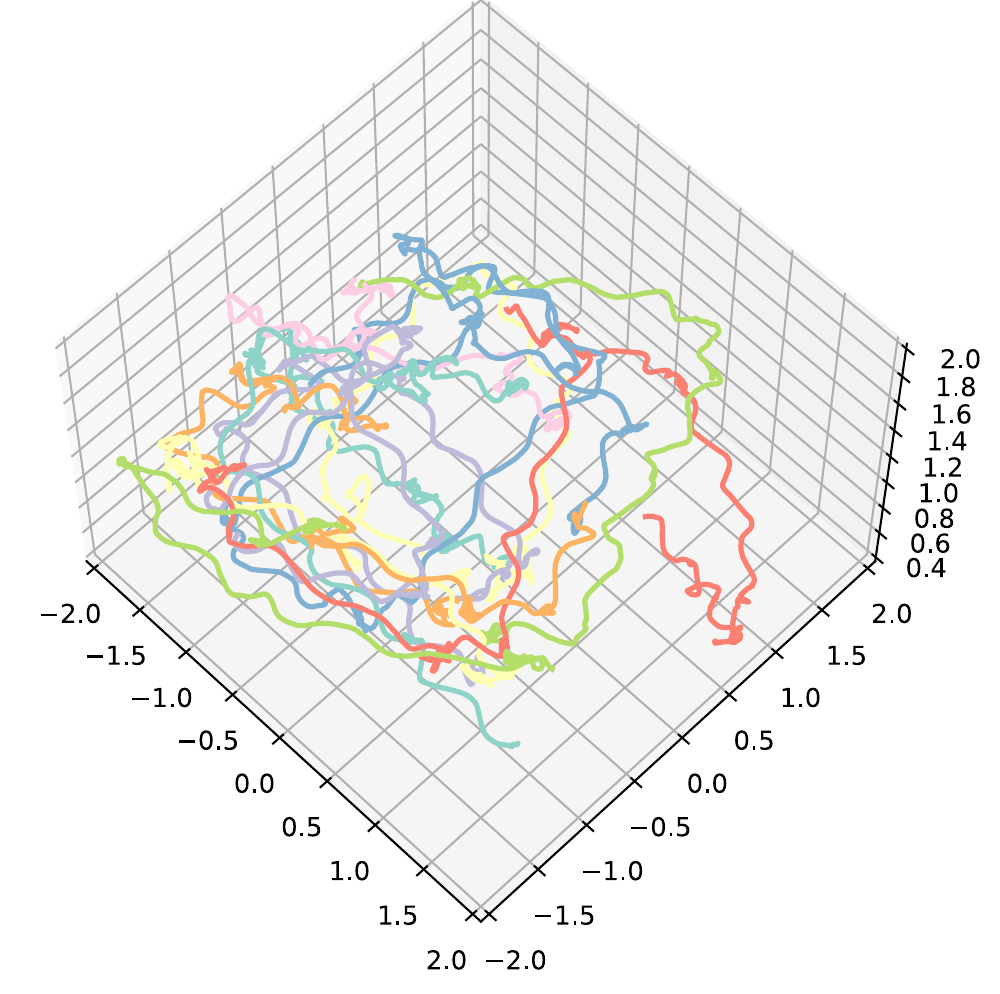}
    }
    \subcaptionbox*{Subject ID 22}{
        \includegraphics[width=0.1\textwidth]{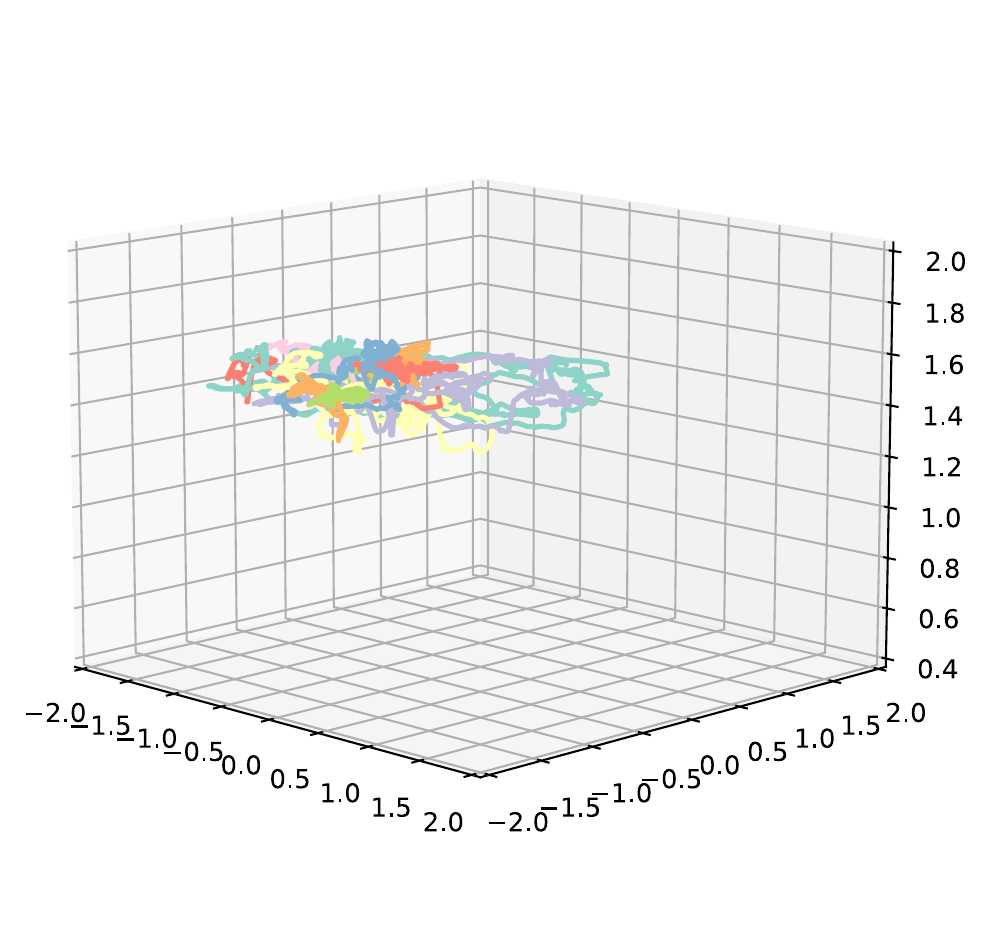}
        \includegraphics[width=0.1\textwidth]{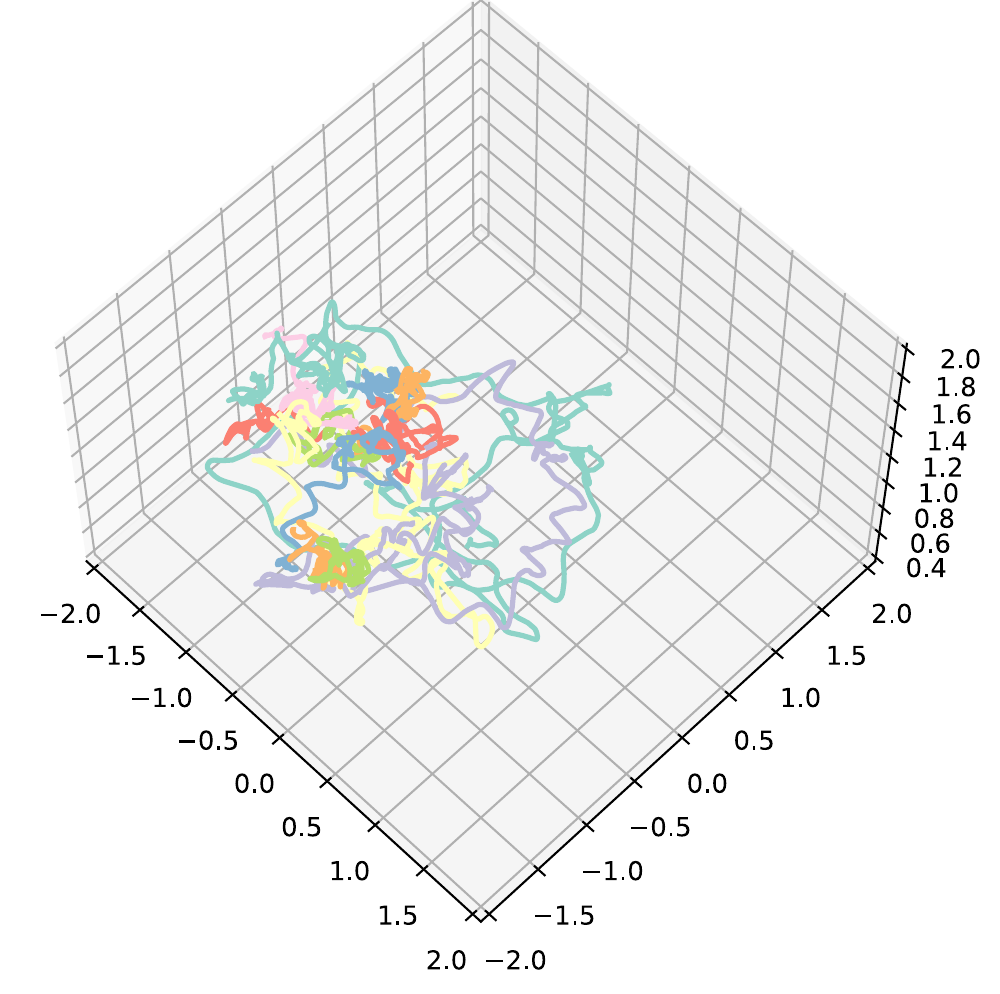}
    }
    \caption{Viewport trajectories captured from four of the subjects. Different subjects have different behavioral characteristics in viewing volumetric scenes. For example, subjects like 07 and 16 prefer to surround the scenes randomly, while subjects like 05 and 22 tend to stand in fixed positions with fewer movements.}
    \label{fig:subjects}
\end{figure}

\subsection{Pre-Processing}

In the pre-processing phase, the entire volumetric scene is divided into several non-overlapping cubic spaces (referred to as \textit{tiles}), which are subsequently downsampled into multiple representations for the purpose of selective transmission. Due to the inherently uneven distribution of 3D Gaussians in 3D space, the number of points contained within each tile can vary. Similarly, the total number of tiles into which the volumetric scene is divided is also not fixed and can change depending on the characteristics of the scene. To facilitate computations, we treat each tile as a 3D region and represent the position of the $k$-th tile by its center coordinates, denoted as $(x_k, y_k, z_k)$. For convenience, let us assume that the entire volumetric scene is divided into a total of $K$ tiles. A voxel grid filter is then applied to downsample each tile into representations at $L$ different quality levels. Specifically, for any two quality levels $l_1$ and $l_2$, where $1 \leq l_1 < l_2 \leq L$, the following relationships hold:

\begin{equation}
    \begin{aligned}
        &b_{k,l_1} \leq b_{k,l_2}, \\
        &p_{k,l_1} \leq p_{k,l_2}, \\
        \text{s.t.}\ &1 \leq l_1 < l_2 \leq L,
    \end{aligned}
\end{equation}

where $b_{k,l}$ denotes the number of bits and $p_{k,l}$ represents the number of points contained in the $l$-th level representation of tile $k$. The pre-processed tile representations are then selectively transmitted to the user based on the predicted viewport sequence generated by the CVP module and the corresponding output produced by the DBA module. This selective transmission ensures that the system can maintain a smooth data flow while still delivering an immersive and satisfactory viewing experience to the user.

\begin{figure}
    \centering
    \includegraphics[width=\linewidth]{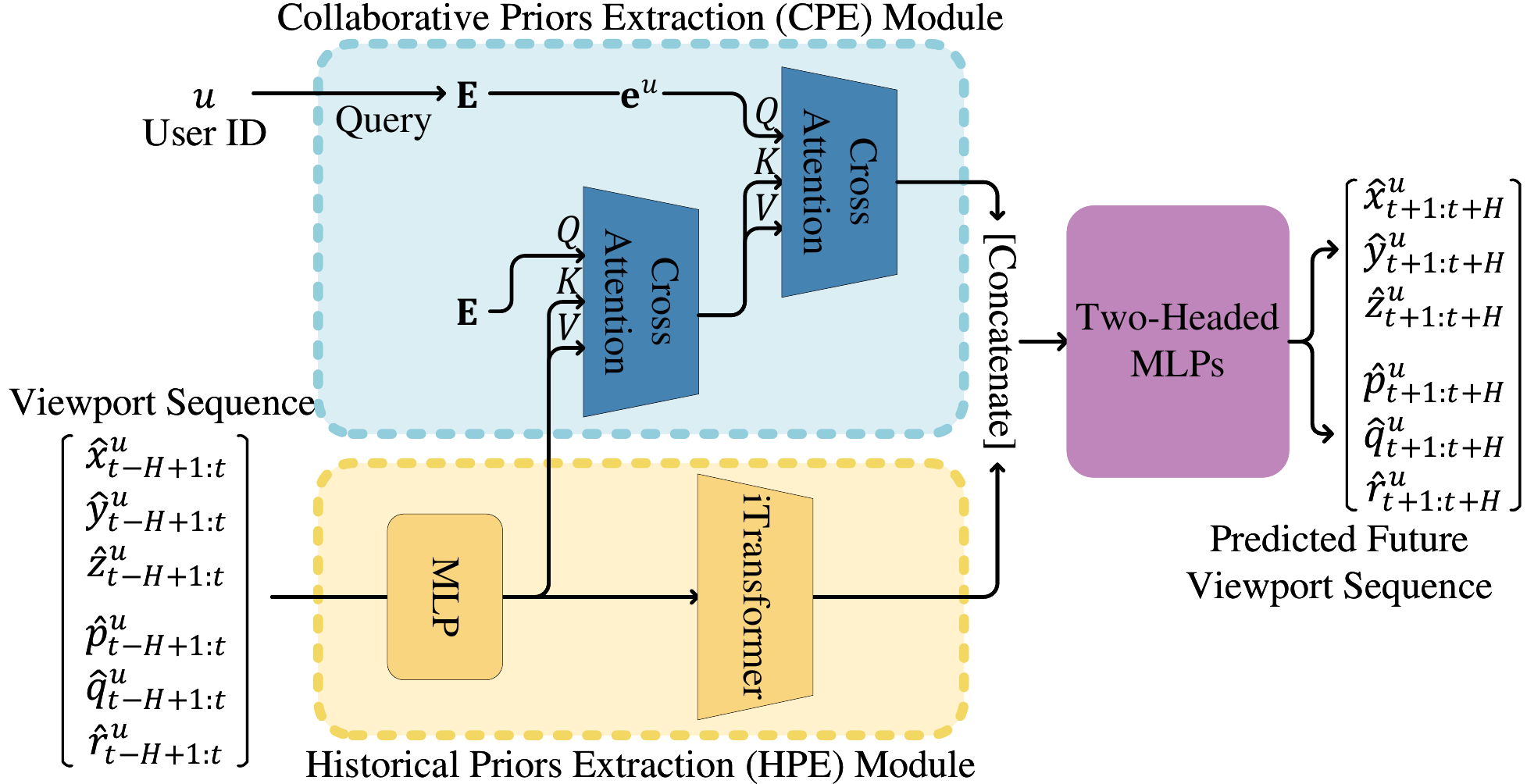}
    \caption{Illustration of the CVP module. The network utilizes both historical information and collaborative information.}
    \label{fig:cvp}
\end{figure}

\subsection{Collaborative Viewport Prediction}

\subsubsection{Motivation}

As a nascent form of volumetric scene representation, there is a notable lack of real user viewing behavior datasets specifically for scenes derived from 3DGS (3D Graphics Systems). This scarcity poses challenges for accurately modeling and understanding user interactions. To bridge this gap, we developed a user behavior data collection platform and successfully constructed a comprehensive dataset from 32 subjects with diverse backgrounds. When compared to existing user behavior datasets based on point cloud representations, our findings reveal that the high-fidelity rendering supported by 3DGS-derived scenes leads to more active user engagement. This increased activity better captures the unique behavioral characteristics of each user. For instance, compared to the point cloud video-based user behavior dataset introduced by~\cite{hu2023understanding}, the mean-variance of users' positional distribution in our dataset is 9$\times$ higher (0.45 vs. 0.08). This significant difference highlights the enhanced variability and richness of user interactions facilitated by 3DGS-derived representations. The qualitative analysis further supports these results, as shown in Fig.~\ref{fig:subjects}, which illustrates the distinct behavioral patterns of each subject. These differences underscore the effectiveness of our dataset in reflecting individual user behaviors. Consequently, we raise an important question: \textit{Can we better predict users' future behavior based on their historical behavior sequences by proposing a method for collaboratively learning user group behavior?} The answer is yes. We introduce a Collaborative Visual Prediction (CVP) module, depicted in Fig.~\ref{fig:cvp}, which effectively leverages both historical and collaborative information to accurately predict the future behavior of the current user.

\subsubsection{Overview}

As depicted in Figure~\ref{fig:cvp}, the proposed CVP module is designed to enhance the accuracy of viewport predictions by integrating both user-specific characteristics and their historical interaction data. It is important to recognize that individual users exhibit varying degrees of behavioral inertia when navigating through volumetric scenes, meaning that their viewing patterns and preferences can differ substantially. Despite this variability, existing viewport prediction methods often process viewport trajectories from diverse users in a uniform manner, thereby overlooking the intricate relationships between unique user attributes and their corresponding viewing behaviors. To address this gap, the proposed network architecture is composed of two primary components: the \textbf{C}ollaborative \textbf{P}riors \textbf{E}xtraction (CPE) module and the \textbf{H}istorical \textbf{P}riors \textbf{E}xtraction (HPE) module. The CPE module is responsible for collaboratively learning user embeddings by leveraging data from multiple users, thereby capturing shared patterns and individual differences. Meanwhile, the HPE module focuses on extracting insights from the historical viewport sequences of each user, enabling the model to understand temporal dynamics and predict future viewports based on past behavior. The entire network takes as input the user identifier $u$ along with the associated historical viewport sequence $\mathbf{v}^u_{t-H+1:t}$. Specifically, a viewport $\mathbf{v}^u_t$ for user $u$ at a given time slot $t$ is represented by a vector defined as
\begin{equation}
    \mathbf{v}^u_t=\left(x^u_t,y^u_t,z^u_t,p^u_t,q^u_t,r^u_t\right)^T,
    \label{eq:viewport}
\end{equation}
where the components $x^u_t$, $y^u_t$, and $z^u_t$ correspond to the spatial position coordinates of the user's virtual camera, while $p^u_t$, $q^u_t$, and $r^u_t$ denote the orientation angles of the virtual camera at time $t$. At each discrete time slot $t$, the instantaneous viewport $\mathbf{v}^u_t$, as captured by the HMD, is centralized on the client side to construct a comprehensive historical viewport sequence. Utilizing this sequence, the network is then capable of predicting future viewports with higher precision, thereby providing a more personalized and responsive user experience.

\subsubsection{Collaborative Priors Extraction}

The CPE module aims to extract collaborative priors by learning from the viewport data of multiple users simultaneously. This collaborative learning approach enables the module to identify common viewing patterns and unique user-specific behaviors, thereby enriching the overall prediction model. The module incorporates two distinct attention mechanisms, each designed to capture different aspects of user interactions. The attention mechanisms operate based on the standard attention function, which involves a query $q$, a key $k$, and a value $v$, mathematically expressed as
\begin{equation}
    \text{Attention}\left(\mathbf{Q},\mathbf{K},\mathbf{V}\right)=\text{softmax}\left(\frac{\mathbf{Q}\mathbf{K}^T}{\sqrt{d_K}}\right)\mathbf{V},\label{eq:attention}
\end{equation}
where the transformed matrices are given by
\begin{equation}
    \mathbf{Q}=q\mathbf{W}_q,\quad \mathbf{K}=k\mathbf{W}_k,\quad \mathbf{V}=v\mathbf{W}_v.
\end{equation}
In this context, $\mathbf{W}_q$, $\mathbf{W}_k$, and $\mathbf{W}_v$ are learnable weight matrices that project the query, key, and value vectors into a suitable feature space. To facilitate user-specific learning, an embedding matrix $\mathbf{E}$ is established, where each column vector $\mathbf{e}^u$ serves as a learnable embedding unique to user $u$. This embedding matrix is initially fed into the first cross-attention module as the query, while the historical viewport sequence acts as both the key and the value. This configuration allows the model to capture temporal attention, effectively learning how each user's viewing behavior evolves over time. The output from this first attention layer is subsequently passed to a second cross-attention module. In this stage, the query is specifically the user embedding $\mathbf{e}^u$, which enables the model to learn inter-user attention, thereby understanding how a particular user's behavior relates to the behaviors of other users. This two-tiered attention mechanism ensures that the CPE module can effectively leverage both temporal dynamics and collaborative information across multiple users to generate robust user embeddings that are instrumental for accurate viewport prediction.

\subsubsection{Historical Priors Extraction}

The HPE module is dedicated to extracting historical priors from the input sequence of past viewports, thereby providing the model with contextual information that is crucial for predicting future viewports. This module comprises two main components: an MLP and an iTransformer module~\cite{liu2023itransformer}. The MLP serves as an initial projection layer, transforming each viewport vector at a given time slot into a higher-dimensional feature space. Following the MLP, the iTransformer module is employed to model the temporal dependencies within the historical viewport sequence, which has been demonstrated to achieve SOTA performance in temporal sequence prediction tasks. In the context of the proposed network, the iTransformer leverages the enhanced feature representations from the MLP to predict future viewport behaviors in the time domain. By doing so, the HPE module not only encapsulates the sequential nature of user interactions but also enhances the model's capacity to anticipate future viewing directions and positions based on past behavior, thereby contributing to more accurate and personalized viewport predictions.

\textbf{Feature Merge:} To synthesize the rich information learned from both the CPE and HPE modules, we employ a series of MLPs to effectively merge these distinct priors. Specifically, the outputs generated by the CPE and HPE modules are first concatenated, creating a unified feature vector that encapsulates both collaborative and historical insights. This concatenated vector is then passed through an MLP, which serves to integrate and transform the combined features into a deep viewport representation. Following this integration, another MLP is utilized to decode the deep viewport features into actionable predictions by bifurcating into two separate MLPs: one dedicated to predicting future camera positions and the other focused on predicting camera rotations.

The CVP module is trained using the captured user viewport trajectory dataset. The training process employs a loss function defined as the mean absolute error (MAE) between the predicted viewports $\hat{\mathbf{v}}^u_{t+1:t+H}$ and the actual ground truth viewports $\mathbf{v}^u_{t+1:t+H}$. By minimizing the MAE during training, the CVP module is able to produce highly accurate viewport predictions that closely align with the users' true viewing behaviors.

\subsection{DRL-based Bitrate Adaptation}

To achieve efficient 3DGS content delivery, we propose a novel DBA module that adaptively determines the optimal transmitted representations at each discrete time slot $t$ based on comprehensive tile statistics, the available network bandwidth of the current user, and the user's anticipated future viewports as predicted by the CVP module. In this section, our proposed bitrate adaptation strategy is specifically defined and analyzed within a single-user scenario, allowing us to omit the superscript $u$ that denotes the user in previous sections, thereby facilitating a more concise and streamlined presentation.

\begin{figure*}
    \centering
    \includegraphics[width=\textwidth]{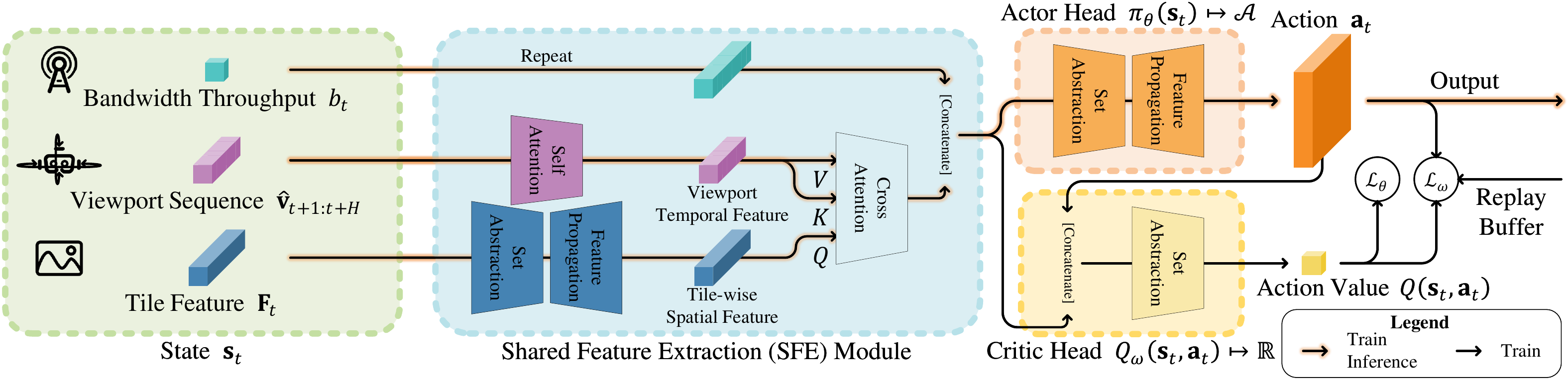}
    \caption{The illustration of the whole architecture of the proposed DRL-based bitrate adaptation algorithm. The proposed algorithm is based on DDPG, which is under the Actor-Critic framework~\cite{konda1999actor}, with the actor head $\pi(\mathbf{s}_t)$ outputting an action $\mathbf{a}_t$, and the critic head outputting the corresponding action value $Q\left(\mathbf{s}_t,\mathbf{a}_t\right)$.}
    \label{fig:ddpg}
\end{figure*}

\subsubsection{MDP Formulation}\label{sec:MDP}

To address the bitrate adaptation challenge across discrete time slots utilizing DRL methods, we model the problem as an MDP. Specifically, we characterize the state space by integrating several key components: the bandwidth throughput $b_t$ at each discrete time slot $t$, the predicted sequence of future viewports denoted as $\hat{\mathbf{v}}_{t+1:t+H}$, and the comprehensive feature information of all tiles present within the volumetric scene, represented by $\mathbf{F}_t$. As a result, the DBA module within the GSStream system functions as an intelligent decision-maker. It is responsible for selecting an appropriate action $\mathbf{a}_t\in\mathcal{A}$ based on the current state $\mathbf{s}_t\in\mathcal{S}$, with the objective of maximizing the received reward $r_t\in\mathbb{R}$. Here, $\mathcal{S}$ and $\mathcal{A}$ denote the state and action spaces, respectively. In greater detail, the definitions of the state space, action space, and the corresponding reward function are elaborated as follows:

\textbf{State Space:} The observed state $\mathbf{s}_t$ at each time slot $t$ is defined as a combination of the current bandwidth throughput $b_t$ (measured in bits per time slot), the predicted future user viewport sequence $\hat{\mathbf{v}}_{t+1:t+H}$ and the feature information of tiles within the volumetric scene $\mathbf{F}_t=\left(\mathbf{f}^1_t,\mathbf{f}^2_t,\ldots,\mathbf{f}^K_t\right)^T$ with $K$ denoting the number of tiles within a certain volumetric scene, as
\begin{equation}
    \mathbf{s}_t=\left(b_t,\hat{\mathbf{v}}_{t+1:t+H},\mathbf{F}_t\right).
\end{equation}
Notably, $\mathbf{f}^k_t$ is a feature vector that describes tile $k$ at time slot $t$. For each tile $k$, the corresponding $\mathbf{f}^k_t$ is defined as
\begin{equation}
    \mathbf{f}^k_t=\left(x_k,y_k,z_k,\frac{p_{k,t}^{\text{prev}}}{p_{k,L}},\frac{s_{k,t}^{\text{prev}}}{s_{k,L}},\frac{l^\text{prev}_{k,t}}{L}\right)^T.
\end{equation}
in which $s_{k,l}$ is an importance score of the $l$-th representation of tile $k$, while $p_{k,t}^{\text{prev}}$, $s_{k,t}^{\text{prev}}$ and $l^\text{prev}_{k,t}$ denote the number of points, the importance score and the quality level of the tiles that have already been transmitted (\textit{or} selected) in previous time slots respectively, with $L$ denoting the highest quality level. Aiming at the splatting nature of the rendering technique of Scaffold-GS, the importance of each anchor is indeed represented by the base scaling factors. To this end, we implement the importance score for each tile as a combination:
\begin{equation}
    s_{k,l}=\sum_{i=1}^{n_{k,l}}\prod(S_i),
\end{equation}
where $n_{k,l}$ is the number of anchors within the $l$-level representation of tile $k$, while $\prod(S_i)$ denoting the production of the scaling factors of anchor $i$ as the base volume of the associated neural Gaussians.

\textbf{Action Space:} In our formulation, the action space is defined as $\mathcal{A}=\mathbb{R}^{K\times L}$. At each time slot, entry $a^{k,l}_t\in\mathbf{a}_t$ denotes the preference score of tile $k$ to be transmitted in selected level $l$ at time slot $t$. Rather than determining the quality level of each tile directly by the DRL method, GSStream selects the representations according to the action $\mathbf{a}_t$ in a greedy way, forming a set
\begin{equation}
\begin{aligned}
    X_t=\left\{x_{t}^{k,l}\right\}\ \  
    s.t.\ \ x_{t}^{k,l}\in\left\{0,1\right\}
\end{aligned},
\end{equation}
in which each item $x_{t}^{k,l}=1$ denotes that tile $k$ is selected to transmitted in level $l$ at time slot $t$, otherwise no. The set $X_t$ is constructed by greedily setting $x_t^{k,l}=1$ with the highest $a_t^{k,l}$ until
\begin{equation}
    \sum_k^K\sum_l^L \left(b_{k,l}-b_{k,t}^\text{prev}\right)\cdot x_{t}^{k,l}>b_t,
\label{eq:greedy}
\end{equation}
to ensure a smooth and progressive streaming. The set $X_t$ also satisfies
\begin{equation}
    \sum_l^L x_{t}^{k,l}\leq1,\ \forall 1\leq k\leq K.
\end{equation}

\textbf{Reward Function:} A reward function is a metric that can evaluate each action $\mathbf{a}_t$ according to the state $\mathbf{s}_t$. In our design, the DRL algorithm participates in decision-making indirectly by determining the set $X_t$. Therefore, the reward function $r_t\left(\mathbf{s}_t,\mathbf{a}_t\right)$ is strongly associated with the set $X_t$, as
\begin{equation}
\begin{aligned}
    r_t(\mathbf{s}_t,\mathbf{a}_t)
    &=\omega_1\sum_{k}^K y_t^k \sum_l^L\left(\frac{p_{k,l}-p_{k,t}^\text{prev}}{p_{k,L}-p_{k,t}^\text{prev}}\right)\cdot x_{t}^{k,l}\\
    &+\omega_2\sum_{k}^K y_t^k \sum_l^L\left(\frac{s_{k,l}-s_{k,t}^\text{prev}}{s_{k,L}-s_{k,t}^\text{prev}}\right)\cdot x_{t}^{k,l}\\
    &-\omega_3\max\left\{t_\text{cost}-\Delta t,0\right\},
\end{aligned}\label{eq:reward}
\end{equation}
in which $y_t^k\in Y_t$ satisfying
\begin{equation}
    Y_t=\left\{y_t^k\right\}\ \ s.t.\ \ y_t^k\in\{0,1\}.
\end{equation}
In the set $Y_t$, each item $y_t^k=1$ denotes that tile $k$ will appear in the user's FoV within $H$ time slots in the future, according to the predicted viewport sequence $\hat{\mathbf{v}}_{t+1:t+H}$, vice versa. Moreover, $\omega_1$, $\omega_2$ and $\omega_3$ are hyperparameters. In the reward function, term $p_{k,l}-p_{k,t}^\text{prev}$ is the number of points in tile $k$ that will be incrementally transmitted at time slot $t$, and term $p_{k,L}-p_{k,t}^\text{prev}$ is the number of the remaining untransmitted points. Inspired by the idea of the Signed Distance Field (SDF)~\cite{oleynikova2016signed}, in which the sign of a distance indicates the collision relationship between a tile and the user's view frustum, the first term in $r(\mathbf{s}_t,\mathbf{a}_t)$ acts as a progressive streaming reward, encouraging the network to transmit the tile representations closer to the user's FoV according to the value of $y_t^k$ and with higher incremental content. Similarly, the second term promotes the willingness of the network to transmit tiles with higher incremental importance score, while the last term is a delay time penalty with $\Delta t$ as the time slot duration, and
\begin{equation}
    t_\text{cost}=\frac{1}{b_t}\sum_k^K\sum_l^L \left(b_{k,l}-b_{k,t}^\text{prev}\right)\cdot x_{t}^{k,l}.
\end{equation}

Therefore, the state transition function can be formulated as $P\left(\mathbf{s}_{t+1},r_t\vert\mathbf{s}_t,\mathbf{a}_t\right)$.

\subsubsection{DDPG-based Algorithm}

The proposed DDPG network architecture is depicted in Figure~\ref{fig:ddpg}. It is important to note that the MDP defined in Section~\ref{sec:MDP} operates within a continuous action space, making it well-suited for resolution using a DDPG-based approach. DDPG is a DRL algorithm that effectively integrates the Actor-Critic framework as introduced by Konda and Tsitsiklis~\cite{konda1999actor}. Instead of employing two separate networks to independently learn the policy function $\pi$ and the action value function $Q$, our design incorporates a \textbf{S}hared \textbf{F}eature \textbf{E}xtraction (SFE) block, along with dedicated actor and critic heads. This shared feature extraction promotes efficiency and coherence between the policy and value estimations.

The SFE block is a pivotal component of our DDPG network, designed to process and integrate various features effectively. It includes a self-attention module that captures temporal features from the predicted viewport sequence $\hat{\mathbf{v}}_{t+1:t+H}$. Additionally, the block utilizes a \textbf{S}et \textbf{A}bstraction (SA) module, and a \textbf{F}eature \textbf{P}ropagation (FP) module, as proposed by Qi \textit{et al.}~\cite{qi2017pointnet++}, to extract tile-wise spatial features. The roles of the SA and FP modules are further elaborated in Section~\ref{sec:spacevariability}. Subsequently, the SFE block merges the temporal and spatial features using a cross-attention module, where the tile-wise spatial feature acts as the query and the viewport temporal feature serves as both the key and value, as described in Equation~(\ref{eq:attention}). The resulting feature is spatially aligned and concatenated with the bandwidth throughput $b_t$, thereby generating a high-level feature for each tile $k$.

The high-level features produced by the SFE block are then fed into the actor and critic heads. In the actor head, we again apply the SA and FP modules to ensure accurate estimation of the tile-wise preference scores $\mathbf{a}_t$. Each entry $a^{k,l}_t$ represents the preference score for transmitting tile $k$ at level $l$ during time slot $t$. In contrast, the critic head focuses on estimating an overall action value function $Q(\mathbf{s}_t,\mathbf{a}_t)$ across all tiles in the scene. To achieve this, the critic head employs a single SA module for global feature abstraction and omits the FP module, streamlining the evaluation of action values.

The training process involves constructing two DDPG networks: the primary DDPG network $\left\{\pi_{\theta}(\mathbf{s}_t), Q_{\omega}(\mathbf{s}_t,\mathbf{a}_t)\right\}$, which interacts with the environment, and the target network $\left\{\pi_{\theta'}(\mathbf{s}_t), Q_{\omega'}(\mathbf{s}_t,\mathbf{a}_t)\right\}$, which aids in stabilizing training. Both networks are initialized with identical parameters. As the primary network interacts with the environment, experiences are stored in a replay buffer $\mathcal{R}$ as tuples $\left(\mathbf{s}_t, \mathbf{a}_t, r_t, \mathbf{s}'_t\right)$, where $\mathbf{s}'_t$ is the next state resulting from action $\mathbf{a}_t$ in state $\mathbf{s}_t$. During training, batches $\mathcal{B} \subseteq \mathcal{R}$ are sampled to update the networks. The critic network's gradient is computed as

\begin{equation}
    \nabla_\omega \mathcal{L}_{\omega} = \frac{1}{\left\vert\mathcal{B}\right\vert} \sum_{i=1}^{\left\vert\mathcal{B}\right\vert} 2 \left(r_i + \gamma Q_{\omega'}\left(\mathbf{s}'_i, \mathbf{a}'_i\right) - Q_{\omega}\left(\mathbf{s}_i, \mathbf{a}_i\right)\right),\label{eq:criticgradient}
\end{equation}

where $\mathbf{a}'_i = \pi_{\theta'}\left(\mathbf{s}'_i\right)$ is the action chosen by the target network in state $\mathbf{s}'_i$, and $\gamma$ is the discount factor. Simultaneously, the policy gradient for updating the actor network is approximated by

\begin{equation}
    \nabla_\theta \mathcal{L}_{\theta} \approx \frac{1}{\left\vert\mathcal{B}\right\vert} \sum_{i=1}^{\left\vert\mathcal{B}\right\vert} \nabla_\theta Q_{\omega}\left(\mathbf{s}_i, \mathbf{a}_i\right).\label{eq:actorgradient}
\end{equation}

This training regimen ensures that the actor improves its policy based on the feedback from the critic, leading to more effective decision-making over time.

In summary, the pseudo-code is shown in Algorithm~\ref{alg:algo}, with $\tau$ as the soft update factor and $\mathcal{N}$ as a random Gaussian noise to encourage the exploration.

\renewcommand{\algorithmicrequire}{\textbf{Input : }}
\renewcommand{\algorithmicensure}{\textbf{Output : }}
\begin{algorithm}
    \caption{DDPG-based Algorithm}
    \label{alg:algo}
    \begin{algorithmic}[1]
    \REQUIRE $\left\{\pi_{\theta}\left(\mathbf{s}\right),Q_{\omega}\left(\mathbf{s},\mathbf{a}\right)\right\}$, $\left\{\pi_{\theta'}\left(\mathbf{s}\right),Q_{\omega'}\left(\mathbf{s},\mathbf{a}\right)\right\}$, 
    
    $P\left(\mathbf{s}_{t+1},r_t|\mathbf{s}_t,\mathbf{a}_t\right)$
        \STATE \textbf{initialization}\ $\theta'\leftarrow\theta$, $\omega'\leftarrow\omega$, $\mathcal{N}$, $\mathcal{R}$;
        \FOR{episode $e=1\rightarrow E$}
            \STATE $\mathbf{s}_1\sim P\left(\mathbf{s}_1\right)$;
            \FOR{step $t=1\rightarrow T$}
                \STATE $\mathbf{a}_t\leftarrow\pi_\theta\left(\mathbf{s}_t\right)+\mathcal{N}$;
                \STATE $\mathbf{s}_{t+1},r_t\sim P\left(\mathbf{s}_{t+1},r_t|\mathbf{s}_t,\mathbf{a}_t\right)$;
                \STATE $\mathcal{R}\leftarrow\mathcal{R}\cup\left\{\left(\mathbf{s}_t,\mathbf{a}_t,r_t,\mathbf{s}_{t+1}\right)\right\}$;
            \ENDFOR
            \STATE sample batch $\mathcal{B}\subseteq\mathcal{R}$;
            \STATE update $\omega$ \textit{w.r.t.} Equation~(\ref{eq:criticgradient});
            \STATE update $\theta$ \textit{w.r.t.} Equation~(\ref{eq:actorgradient});
            \STATE $\theta'\leftarrow\tau\theta+(1-\tau)\theta'$;
            \STATE $\omega'\leftarrow\tau\omega+(1-\tau)\omega'$;
        \ENDFOR
    \end{algorithmic}
\end{algorithm}

\subsubsection{Space Variability}\label{sec:spacevariability}

In various volumetric scenes, the distribution of points within the 3D space is often non-uniform, and the bounding boxes do not entirely overlap. This inconsistency results in a variable number of tiles across different scenes. Moreover, this variability causes the state and action spaces in the MDP to fluctuate accordingly. Within DRL algorithms, this issue of spatial variability manifests itself in the form of inputs and outputs for the neural network that consists of tensors with varying dimensions.

One potential solution to this challenge is to implement neural network modules specifically designed to handle tensors with varying dimensions. For instance, consider the input tile feature $\mathbf{F}_t$. A straightforward approach would be to employ a Transformer or an attention-based module~\cite{vaswani2017attention} to effectively capture the features of each tile, treating $\mathbf{F}_t$ as a sequence with a variable length. However, this approach introduces two significant issues. Firstly, from a data structure perspective, each tile is represented by a vector $\mathbf{f}^k_t$, and $\mathbf{F}_t$ essentially constitutes an unordered set of vectors with varying lengths. As a result, the inherent ordering of $\mathbf{F}_t$ can interfere with the accurate extraction of features. Secondly, conventional Transformers and attention modules are not adept at emphasizing the spatial relationships between tiles, as they typically treat the spatial coordinates of each tile on par with other attributes, thereby neglecting the spatial dependencies. To overcome these limitations, we incorporate the SA and FP modules from~\cite{qi2017pointnet++} to effectively manage the variability in the state space, as depicted in Figure~\ref{fig:ddpg}. 

Within the SFE block, we consider $\mathbf{F}_t$ as an unordered collection of vectors. The sampling and grouping layers of the SA module are responsible for hierarchically aggregating features extracted from sampled tiles and their neighboring tiles, thereby constructing more sophisticated, high-level representations. Subsequently, the feature propagation mechanism ensures that these high-level features are accurately learned with respect to each individual tile. Consequently, the tile-wise spatial features illustrated in Figure~\ref{fig:ddpg} form an unordered set where each element is a vector representing the high-level embedding of its corresponding tile. This embedding reflects not only the intrinsic features of the tile but also its local and global relationships within the overall volumetric scene. In a similar fashion, we apply the SA and FP modules within both the actor head and the critic head to enhance the overall streaming performance of the system.

\begin{table}[htbp]
    \centering
    \begin{tabular}{c|ccccc}
        \toprule
        Scene & \textit{bicycle} & \textit{bonsai} & {\color{red}\textit{chtf}} & \textit{counter} & \textit{drjohnson}\\
        \#Tiles & 2,130 & 529 & 2,736 & 180 & 59\\
        \midrule
        Scene & \textit{flowers} & {\color{red}\textit{garden}} & \textit{kitchen} & \textit{playroom} & \textit{room}\\
        \#Tiles & 1,340 & 693 & 480 & 124 & 197\\
        \midrule
        Scene & \textit{sculpture} & {\color{red}\textit{stump}} & \textit{train} & \textit{treehill} & \textit{truck}\\
        \#Tiles & 1,471 & 11,601 & 1,267 & 2,326 & 1,741\\
        \bottomrule
    \end{tabular}
    \caption{The numbers of tiles of each scene. The 3 scenes marked {\color{red}red} construct the testing set while the remaining 12 scenes form the training set.}
    \label{tab:tiles}
\end{table}

\section{Experiments}

To assess the performance of GSStream against existing streaming systems, we conduct a series of experiments. These experiments were designed to simulate real user behaviors and adapt to dynamic network conditions, ensuring that our evaluation reflects practical usage scenarios.

\begin{figure}[htbp]
    \centering
    \subcaptionbox*{\textit{bicycle}}{
        \includegraphics[width=0.075\textwidth]{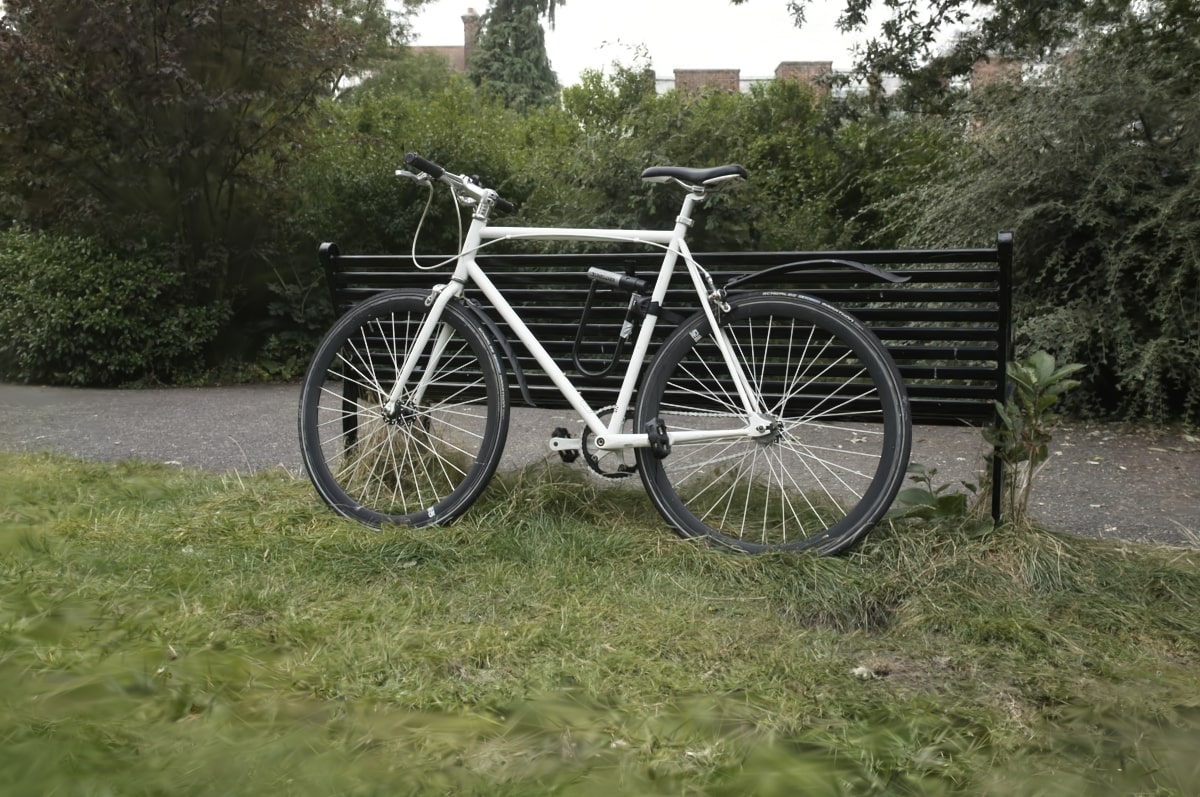}
    }
    \subcaptionbox*{\textit{bonsai}}{
        \includegraphics[width=0.075\textwidth]{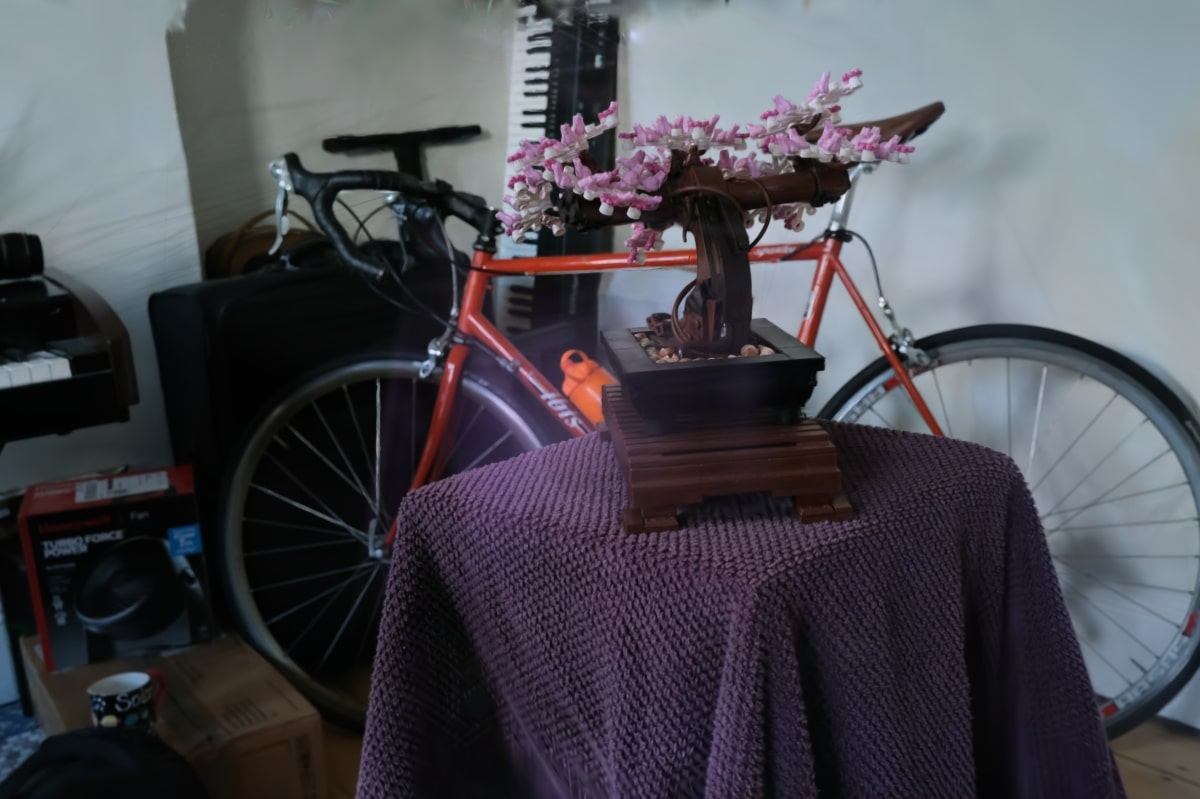}
    }
    \subcaptionbox*{\textit{{\color{red}{chtf}}}}{
        \includegraphics[width=0.075\textwidth]{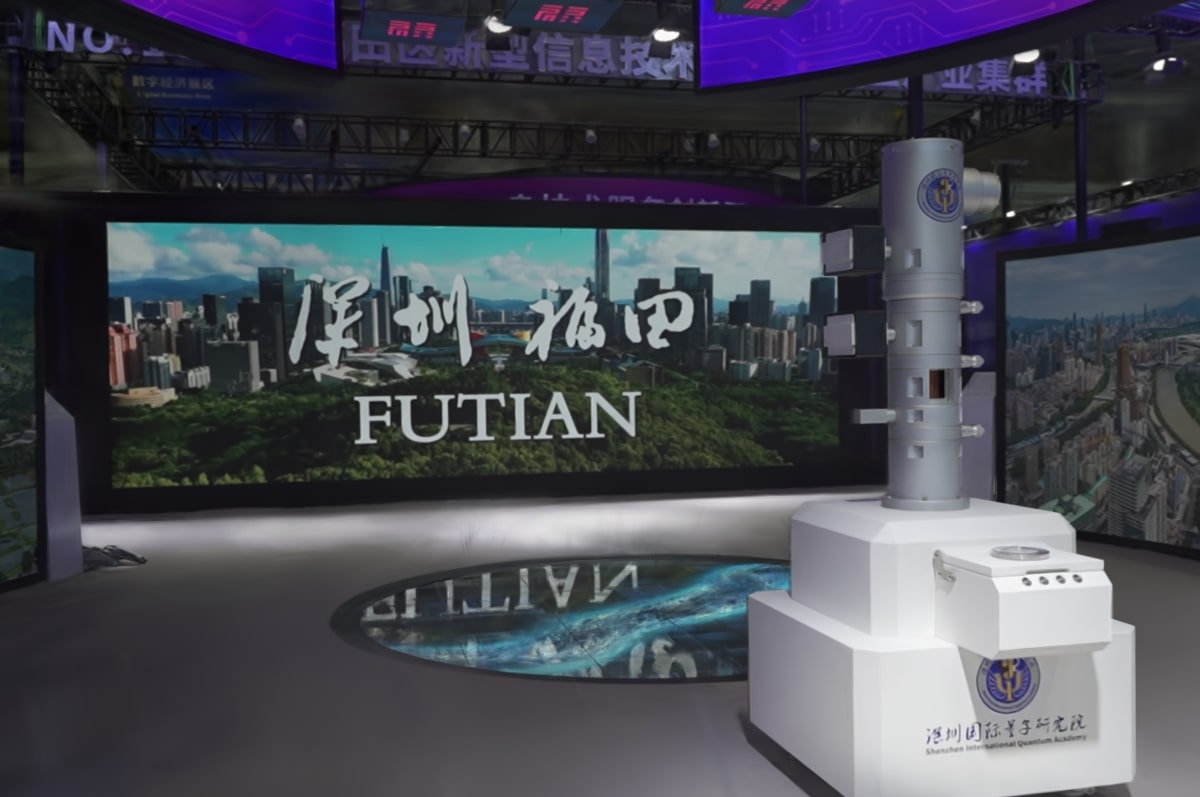}
    }
    \subcaptionbox*{\textit{counter}}{
        \includegraphics[width=0.075\textwidth]{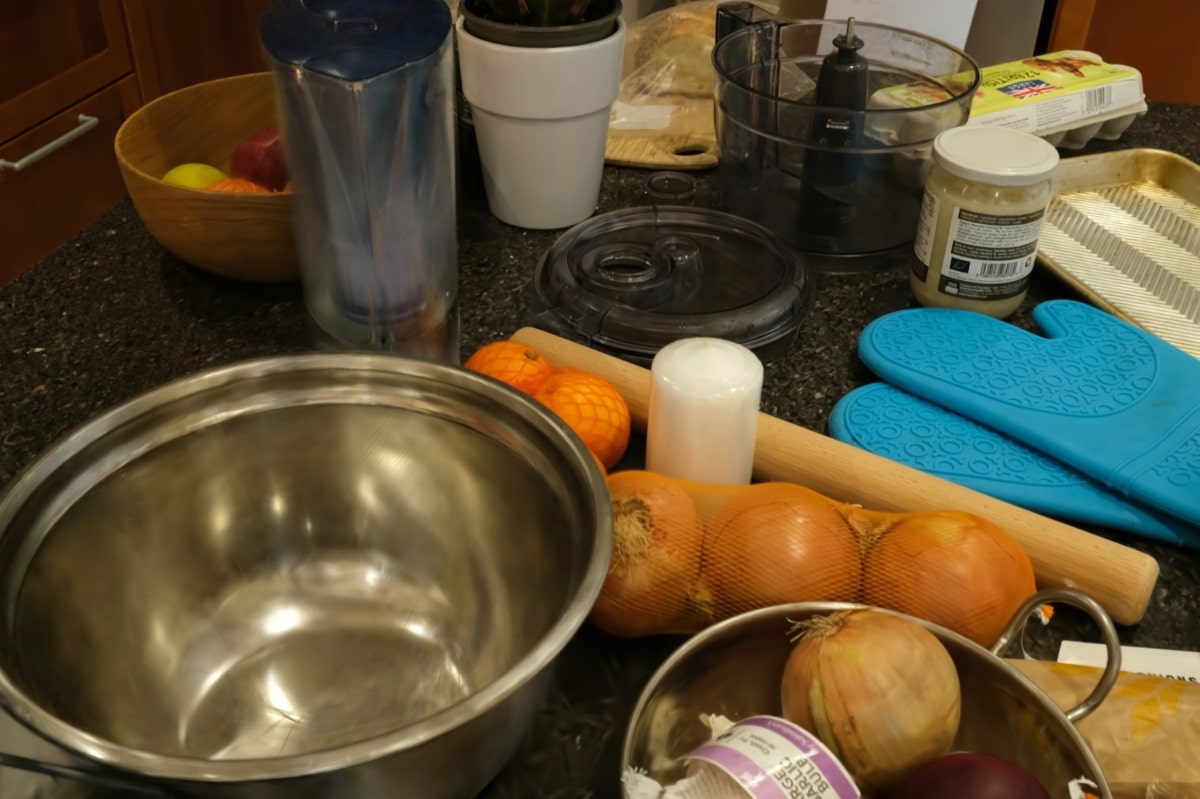}
    }
    \subcaptionbox*{\textit{drjohnson}}{
        \includegraphics[width=0.075\textwidth]{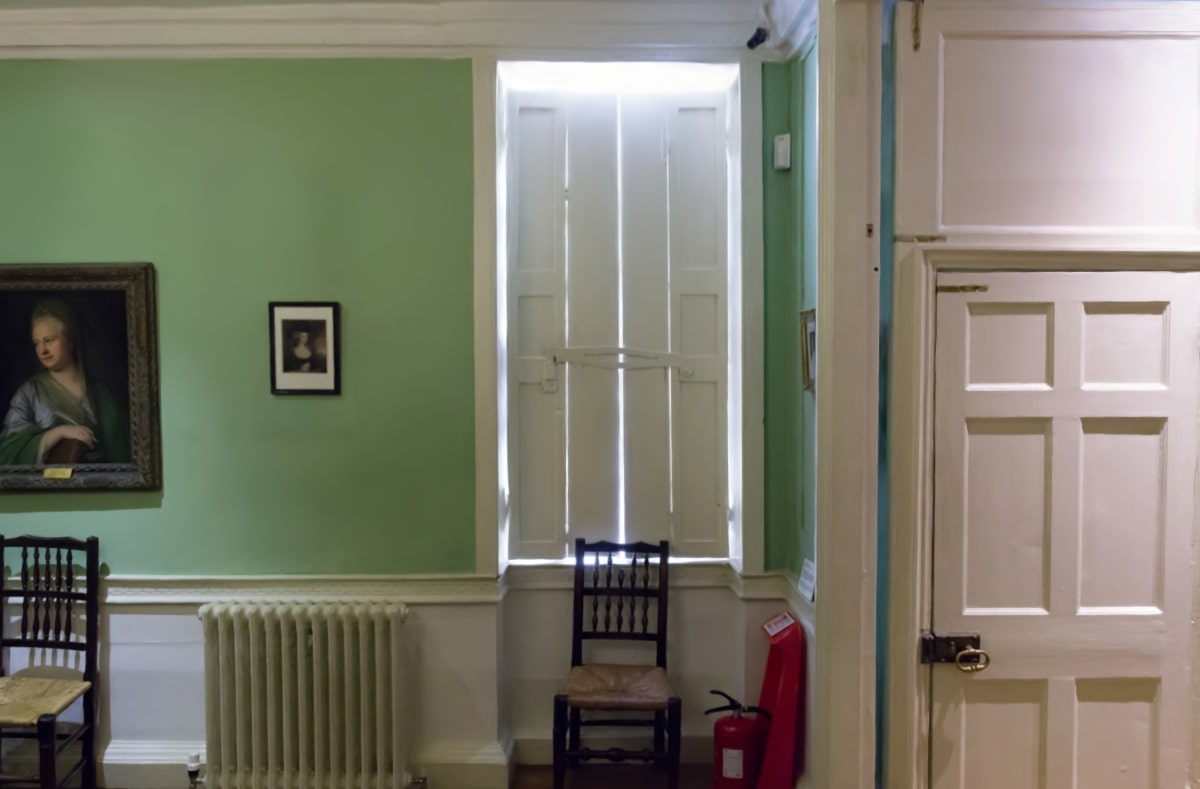}
    }\\
    \subcaptionbox*{\textit{flowers}}{
        \includegraphics[width=0.075\textwidth]{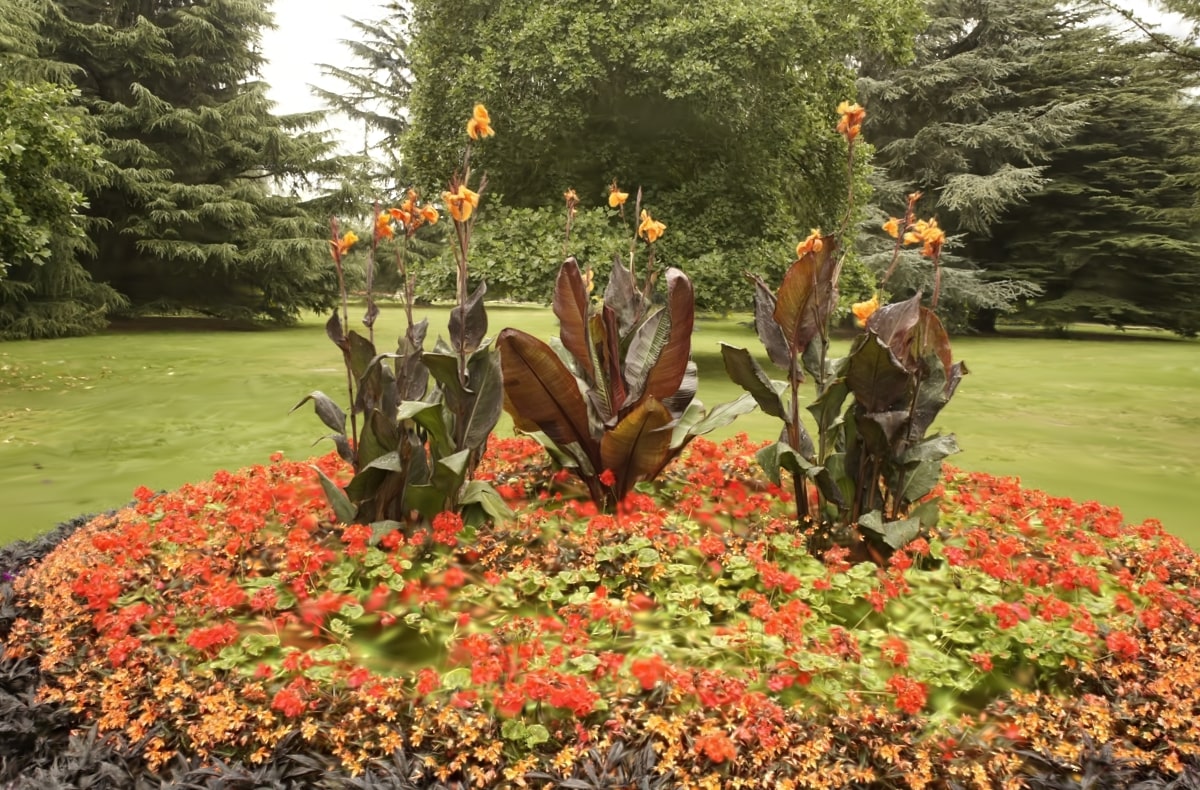}
    }
    \subcaptionbox*{\textit{{\color{red}{garden}}}}{
        \includegraphics[width=0.075\textwidth]{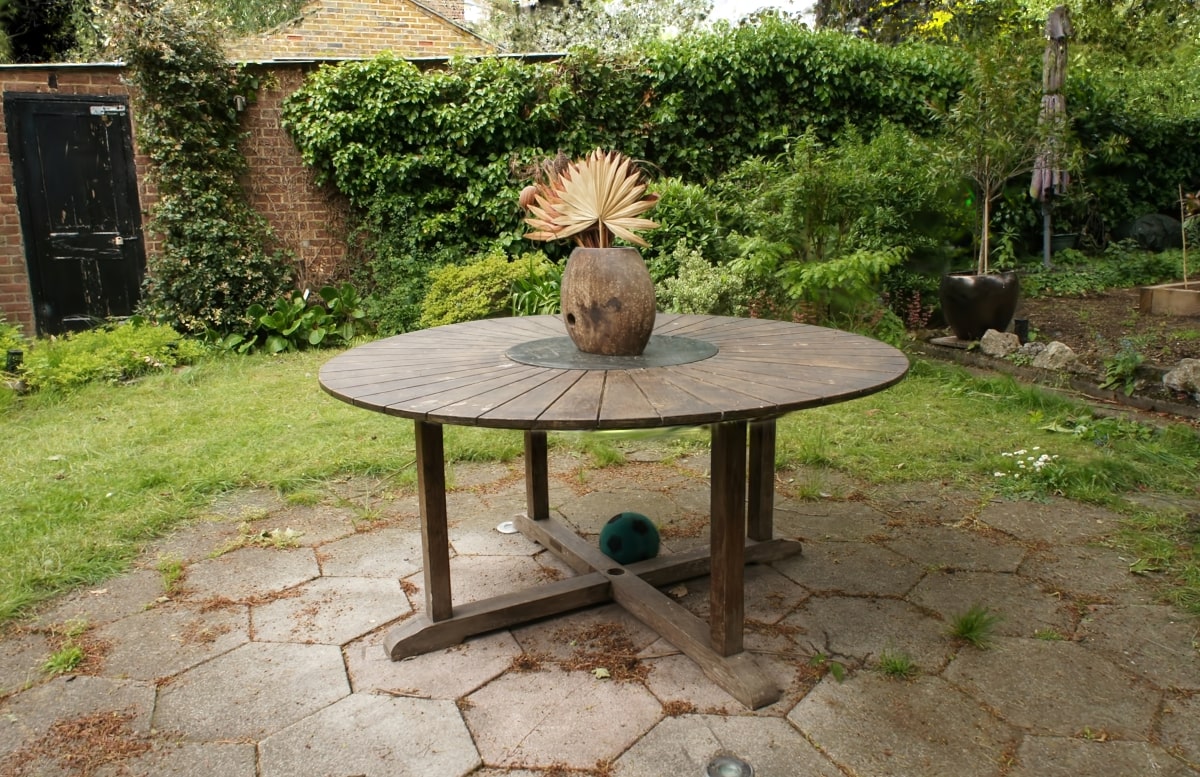}
    }
    \subcaptionbox*{\textit{kitchen}}{
        \includegraphics[width=0.075\textwidth]{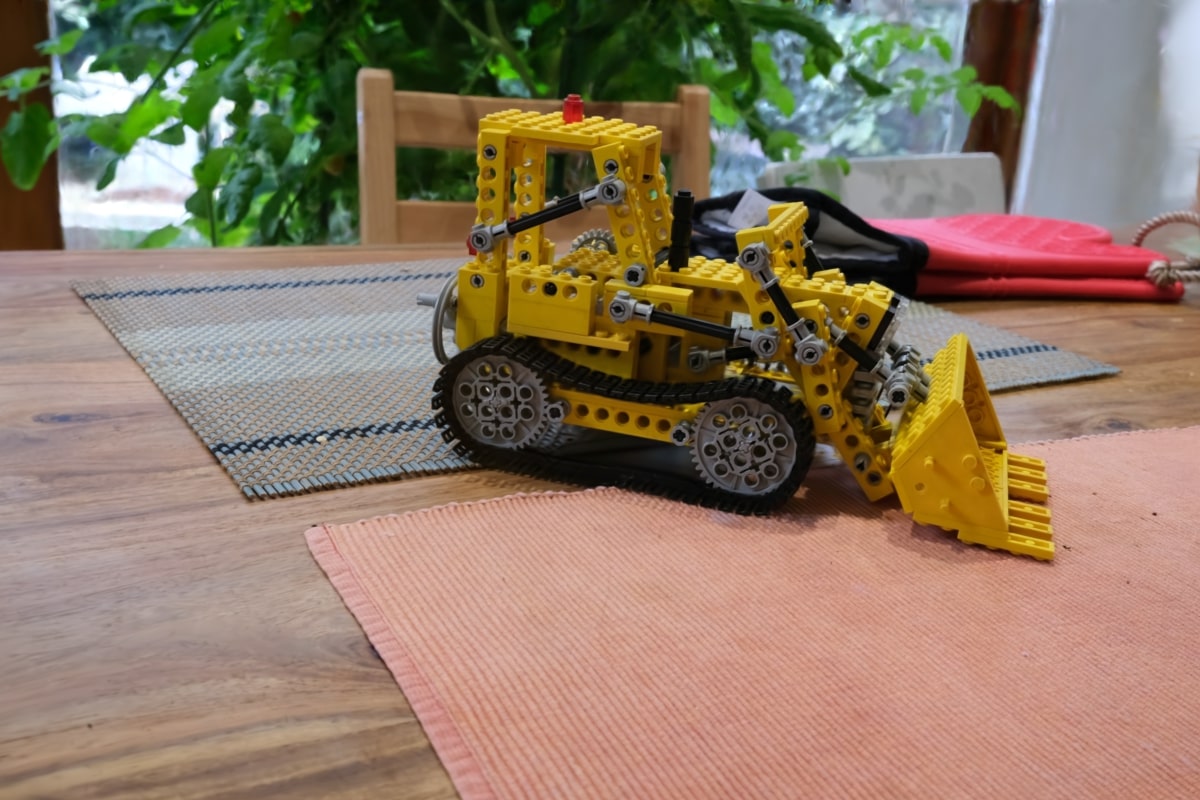}
    }
    \subcaptionbox*{\textit{playroom}}{
        \includegraphics[width=0.075\textwidth]{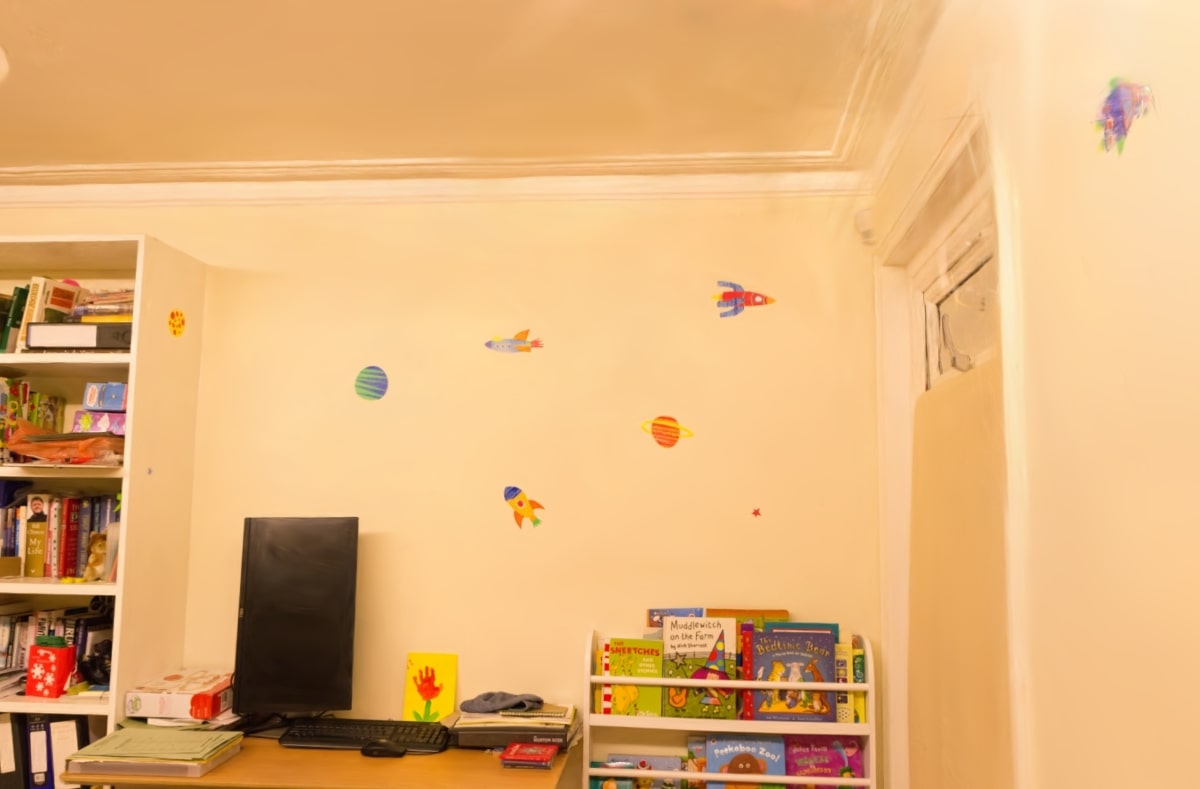}
    }
    \subcaptionbox*{\textit{room}}{
        \includegraphics[width=0.075\textwidth]{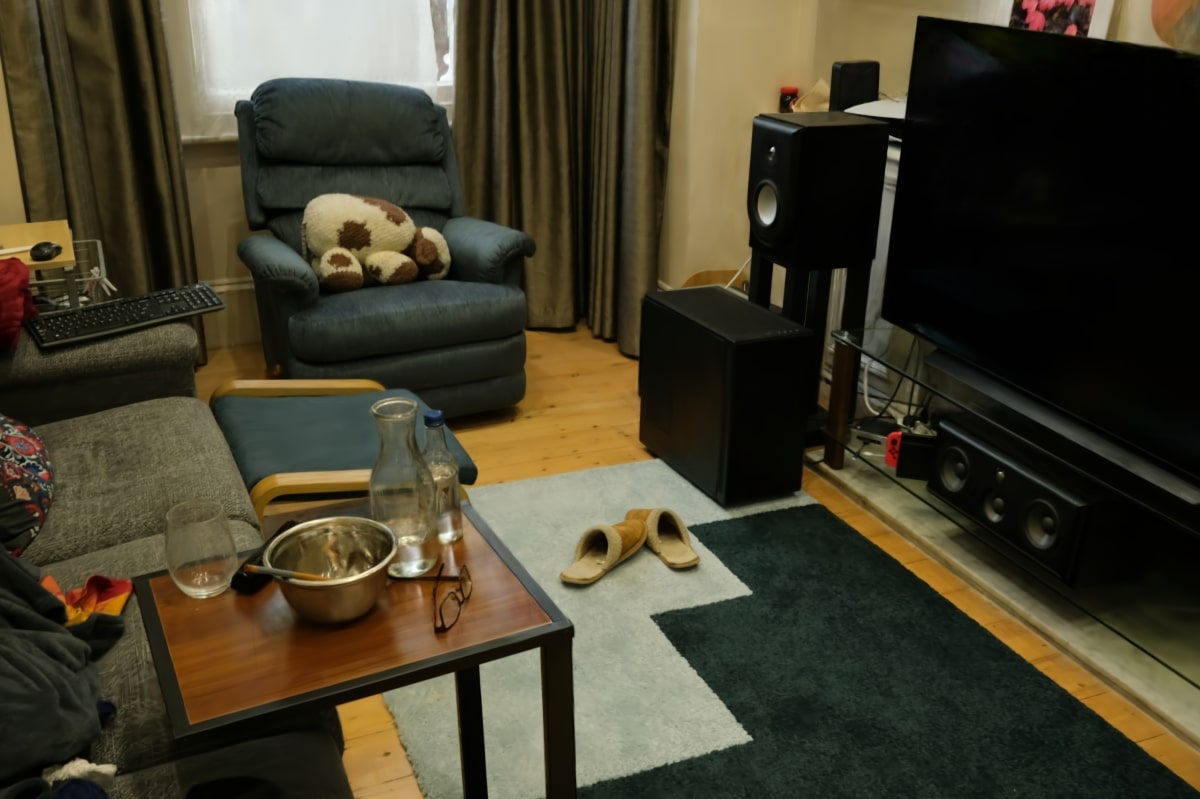}
    }\\
    \subcaptionbox*{\textit{sculpture}}{
        \includegraphics[width=0.075\textwidth]{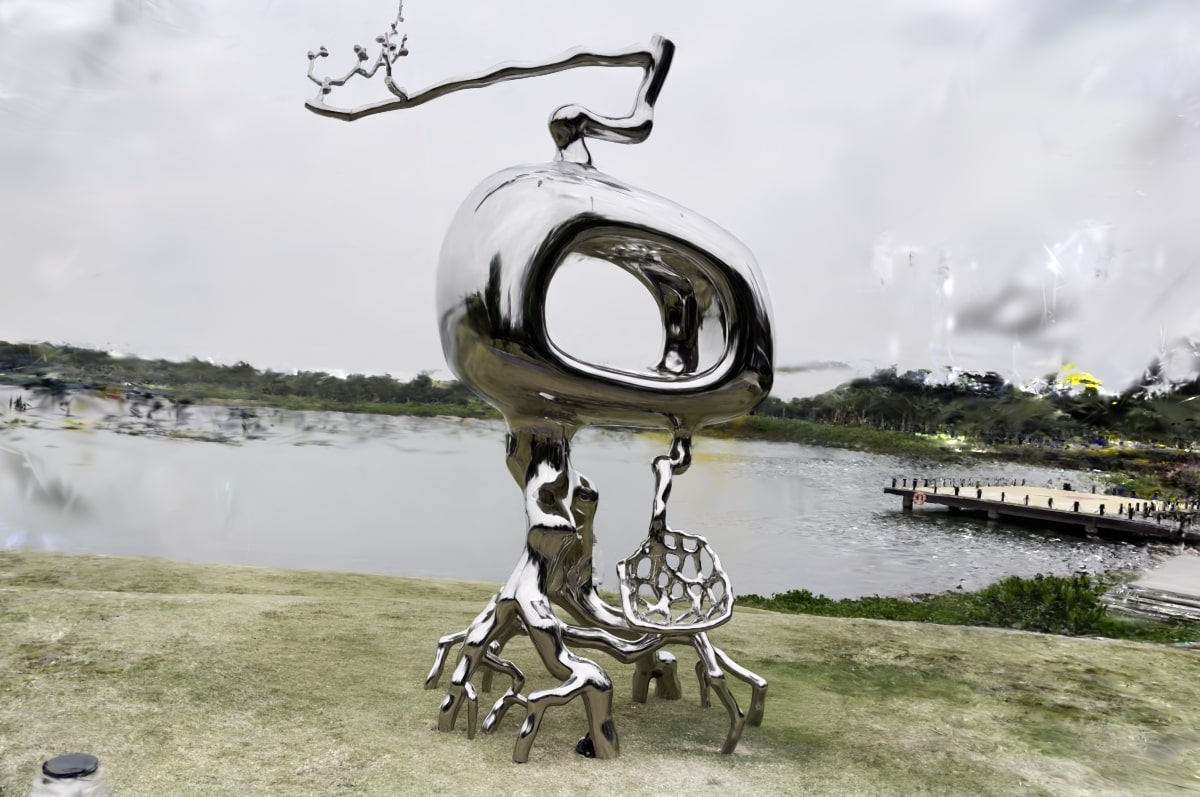}
    }
    \subcaptionbox*{\textit{{\color{red}{stump}}}}{
        \includegraphics[width=0.075\textwidth]{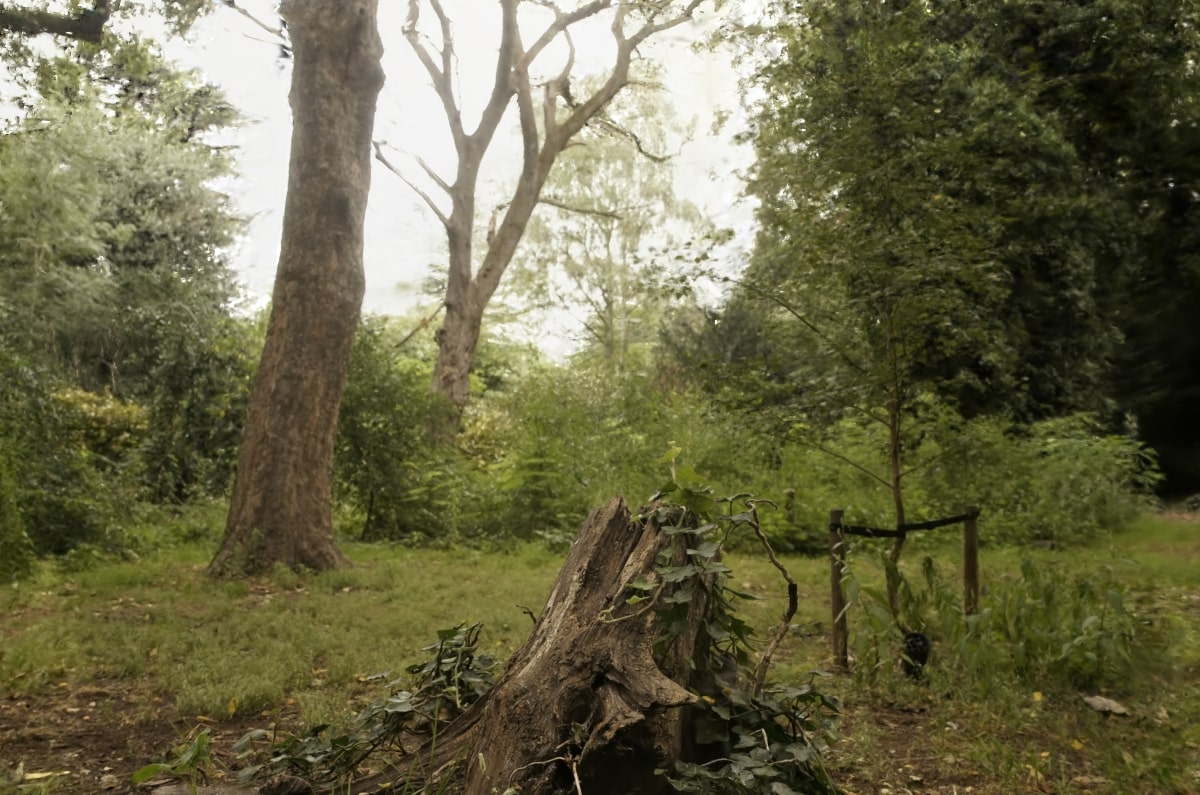}
    }
    \subcaptionbox*{\textit{train}}{
        \includegraphics[width=0.075\textwidth]{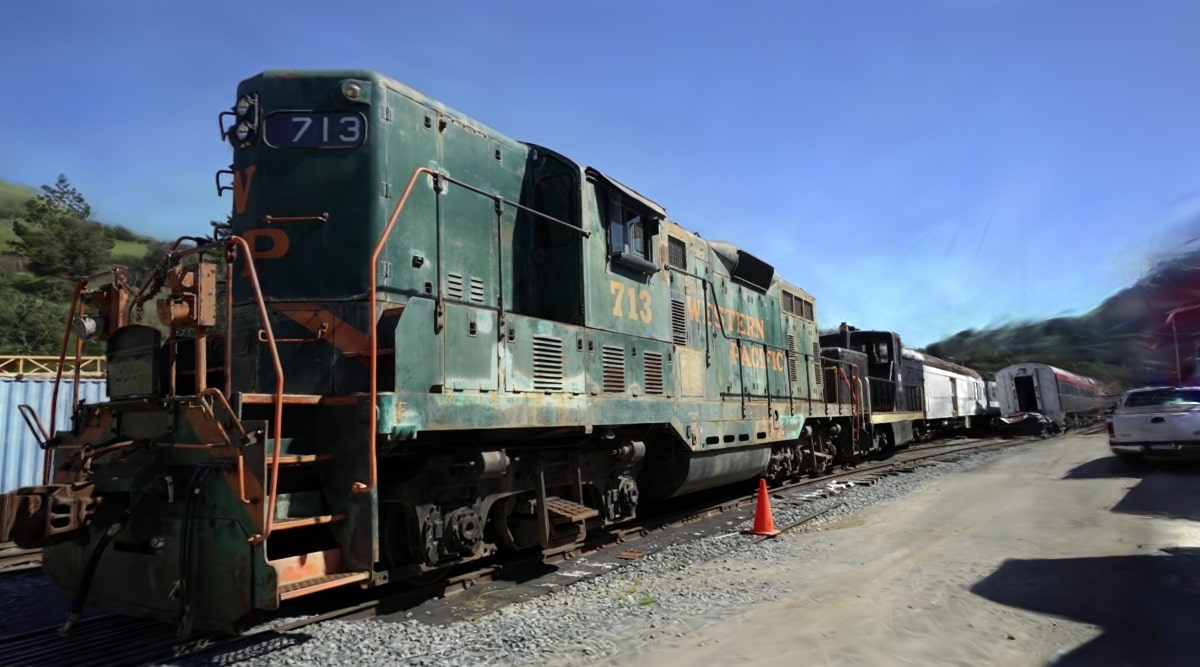}
    }
    \subcaptionbox*{\textit{treehill}}{
        \includegraphics[width=0.075\textwidth]{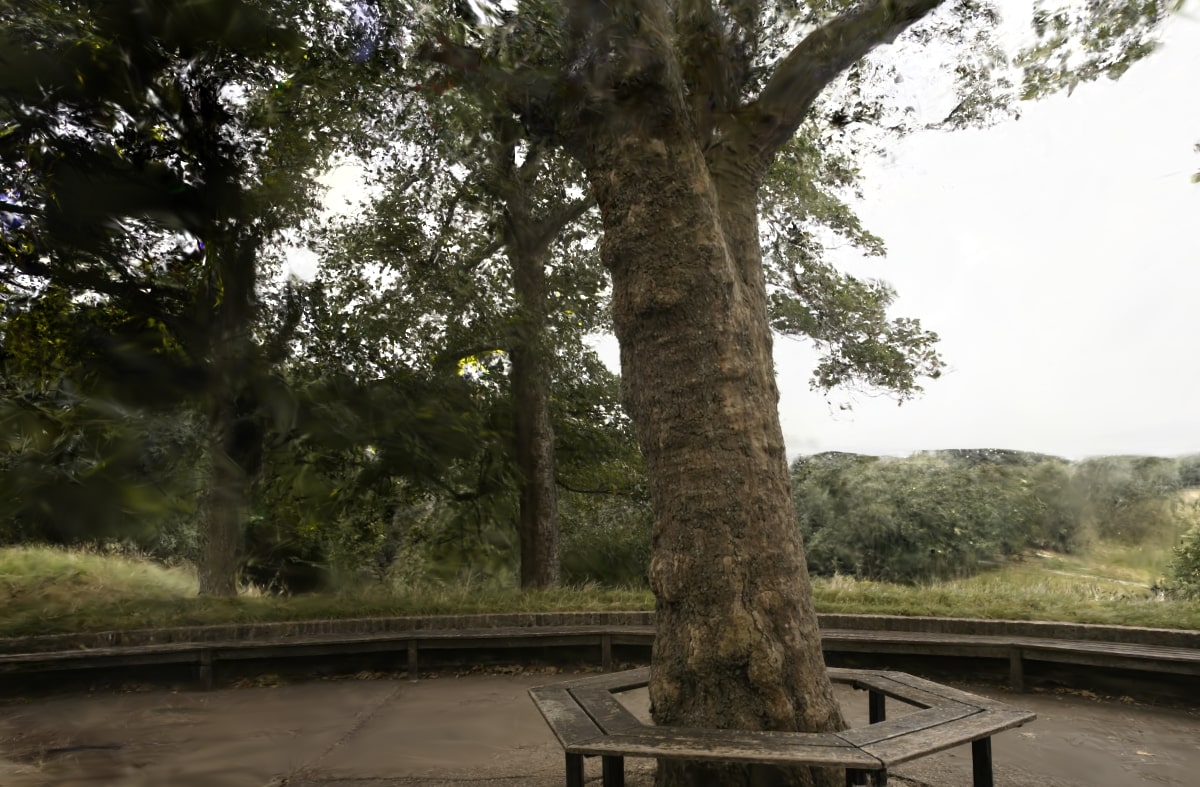}
    }
    \subcaptionbox*{\textit{truck}}{
        \includegraphics[width=0.075\textwidth]{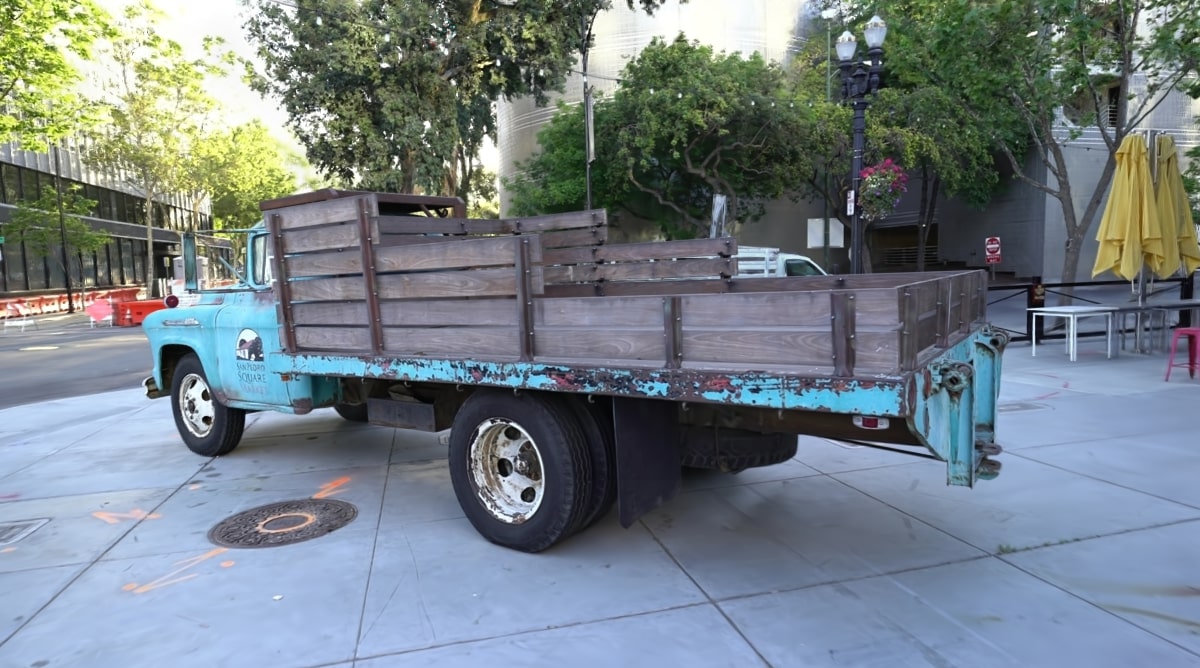}
    }
    \caption{Thumbnails of the volumetric scenes, with scenes marked {\color{red}{red}} forming the testing set, and the others forming the training set.}
    \label{fig:thumbnail}
\end{figure}

\subsection{Dataset}

We initiated our study by developing a user viewport trajectory collection platform specifically for 3DGS-based streaming tasks. A total of 32 subjects were recruited to provide viewport trajectory data. The dataset includes 15 volumetric scenes, with a balanced distribution of 8 outdoor and 7 indoor environments. Each subject was instructed to freely explore each volumetric scene for one minute, resulting in a dataset that captured over 864,000 frames. This extensive data collection ensures a realistic and comprehensive dataset, enhancing the validity of our experimental results. All volumetric scenes utilized in the experiments are presented as thumbnails in Figure~\ref{fig:thumbnail}, offering a visual overview of the scenes included in our study.

\subsubsection{Scene Setup}

Out of the 15 scenes utilized, 13 are selected from public-available datasets. Specifically, 2 from Tanks\&Temples~\cite{Knapitsch2017}, 2 from Deep Blending~\cite{DeepBlending2018}, and 9 from Mip-NeRF 360~\cite{barron2022mipnerf360}. The remaining 2 scenes were captured by our team to ensure a diverse set of environments. Each scene is pre-trained following the pipeline from~\cite{lu2024scaffold} and divided into tiles of size $3.2\text{m} \times 3.2\text{m} \times 3.2\text{m}$. The number of tiles per scene is detailed in Table~\ref{tab:tiles}. We create $L=6$ quality levels through voxel grid downsampling with voxel sizes of 16cm, 8cm, 4cm, 2cm, and 1cm, respectively, while the highest quality level retains all unsampled points.

\subsubsection{Dataset Setup}

\begin{figure}[htbp]
    \centering
    \subcaptionbox*{Subject 01}{
        \includegraphics[width=0.05\textwidth]{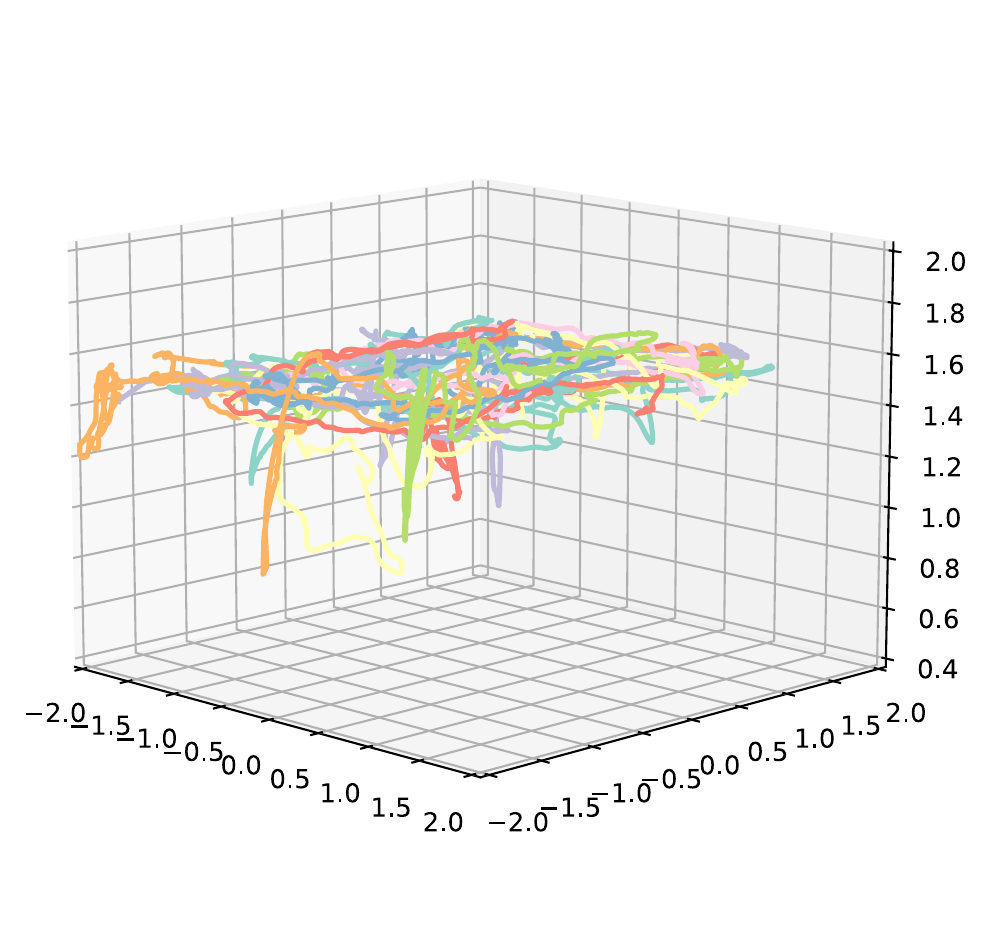}
        \includegraphics[width=0.05\textwidth]{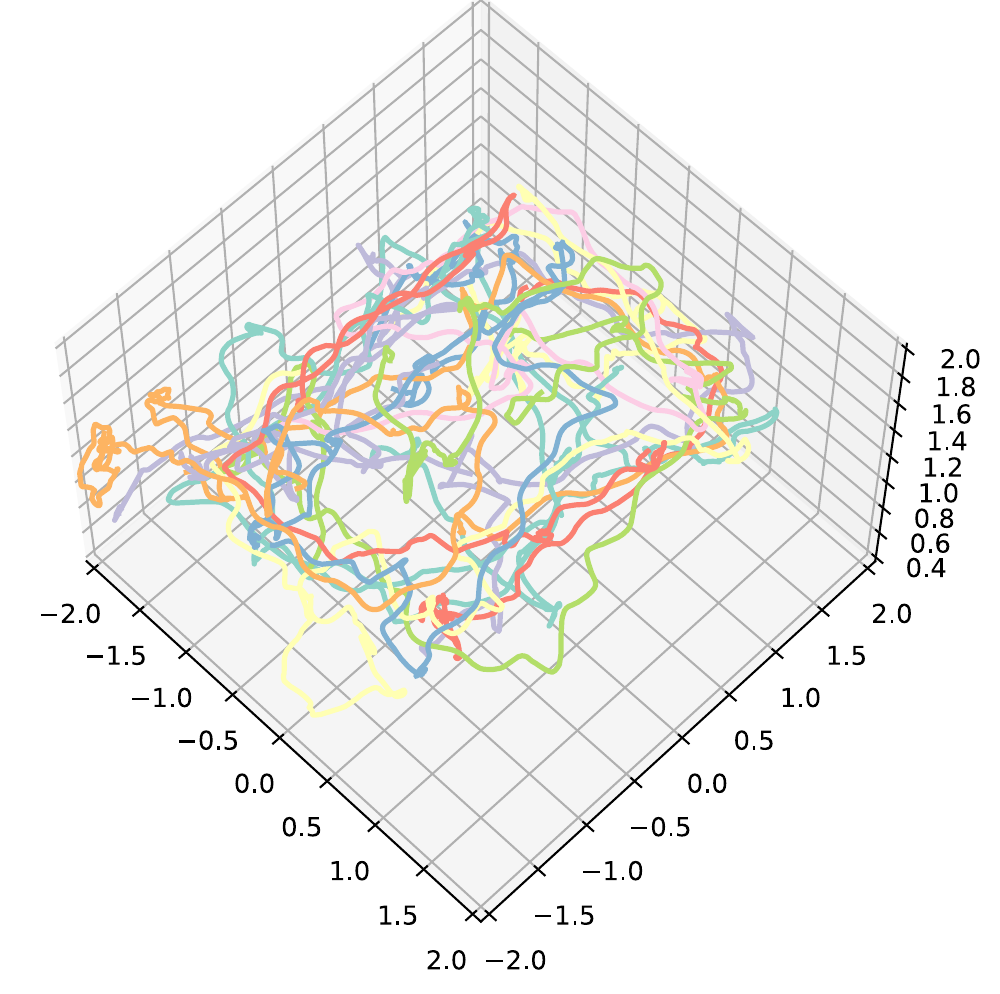}
    }
    \hspace{-10pt}
    \subcaptionbox*{Subject 02}{
        \includegraphics[width=0.05\textwidth]{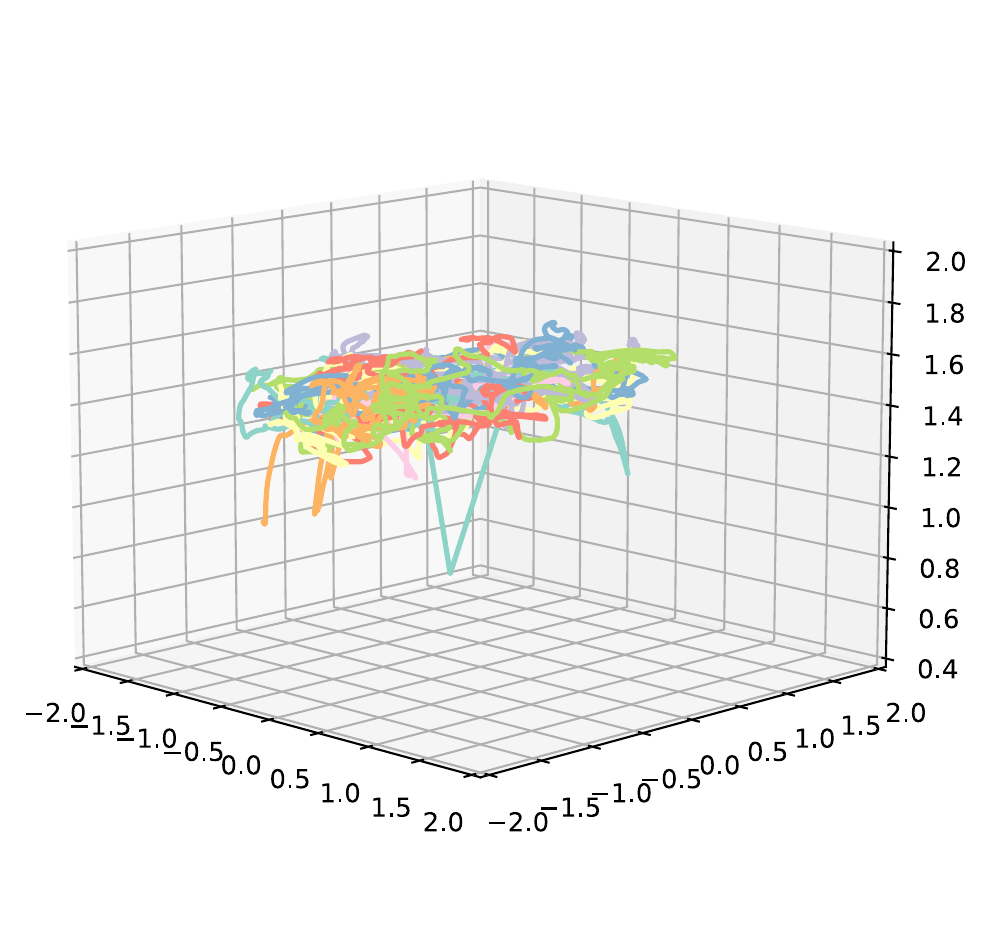}
        \includegraphics[width=0.05\textwidth]{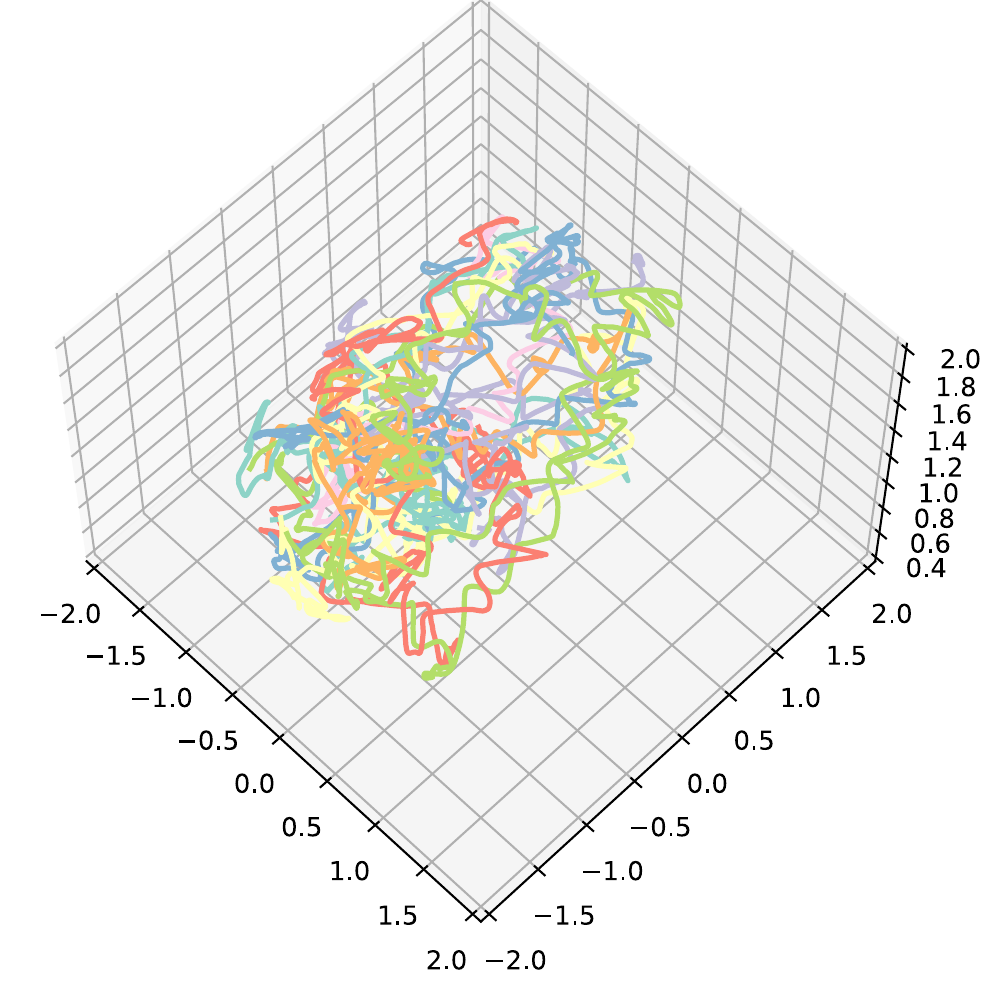}
    }
    \hspace{-10pt}
    \subcaptionbox*{Subject 03}{
        \includegraphics[width=0.05\textwidth]{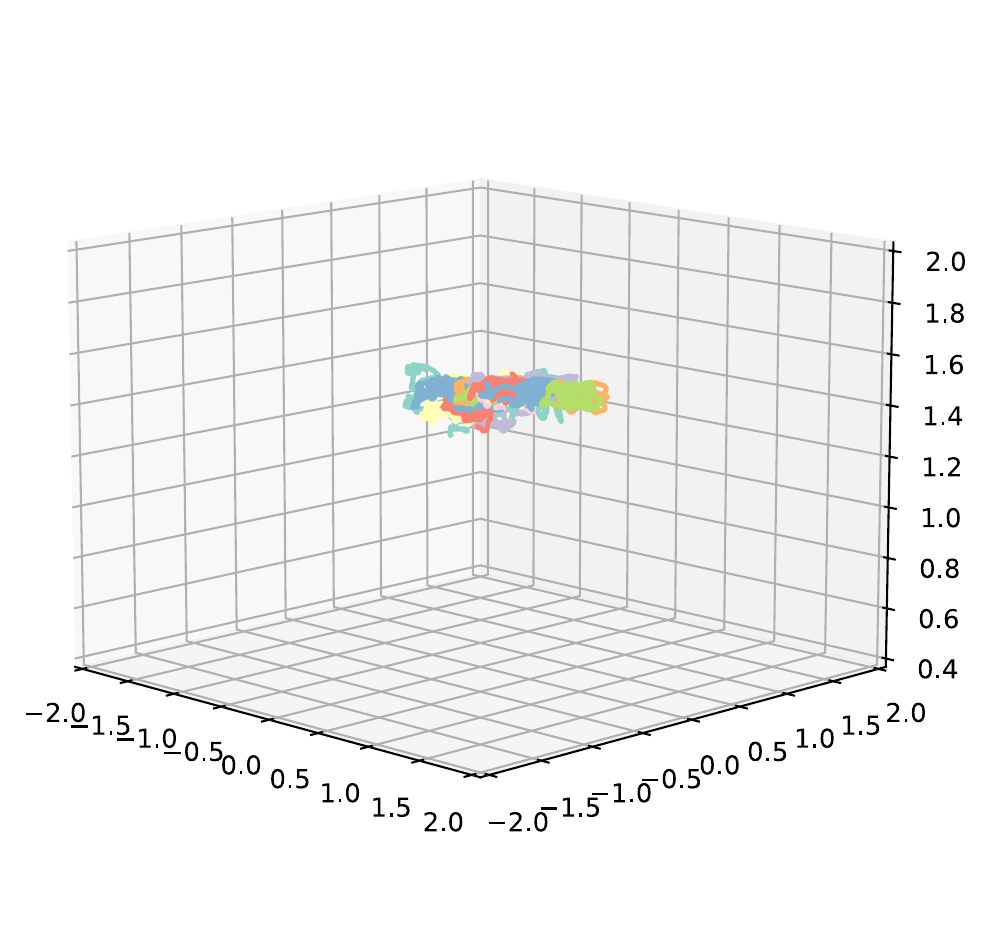}
        \includegraphics[width=0.05\textwidth]{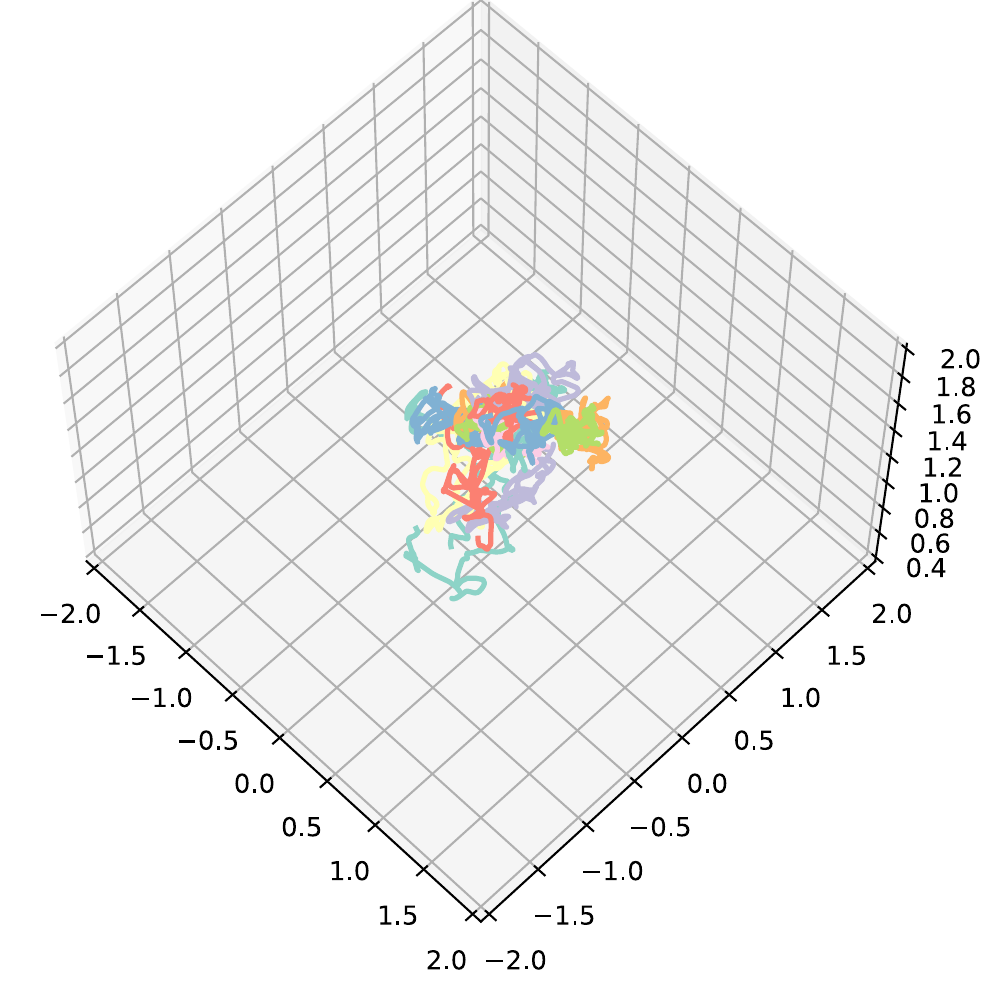}
    }
    \hspace{-10pt}
    \subcaptionbox*{Subject 04}{
        \includegraphics[width=0.05\textwidth]{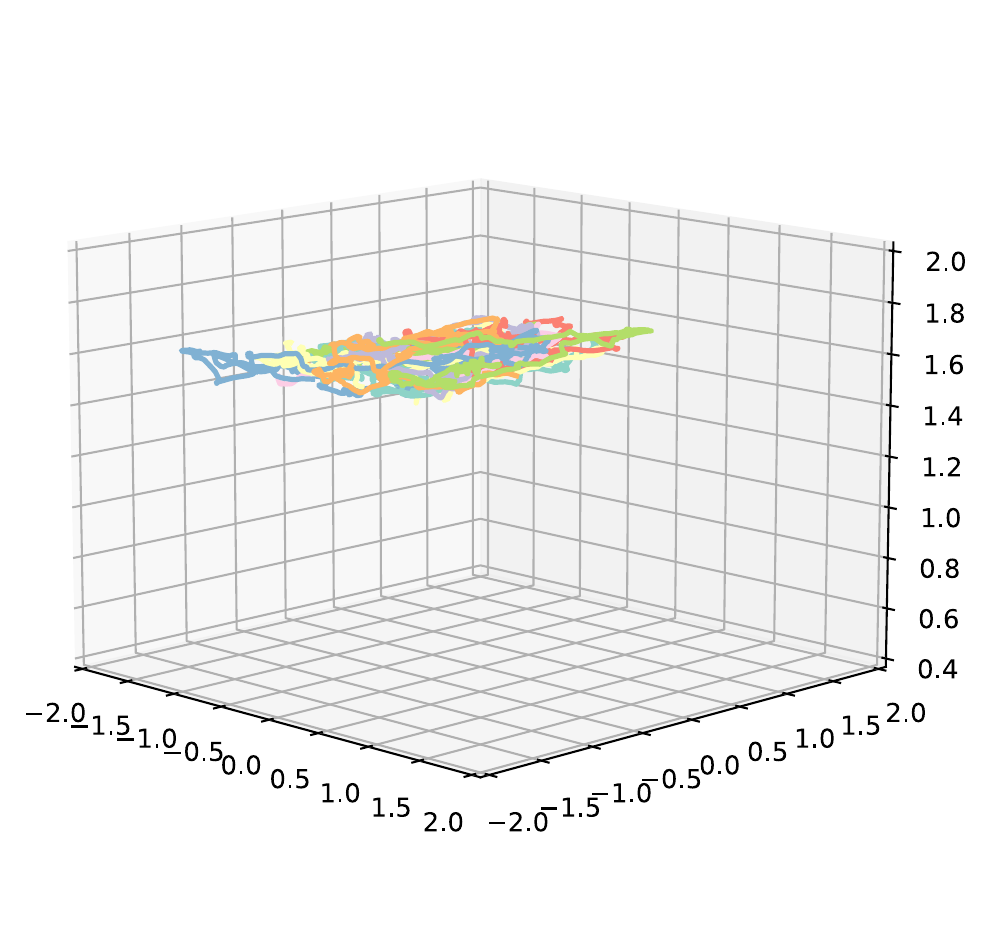}
        \includegraphics[width=0.05\textwidth]{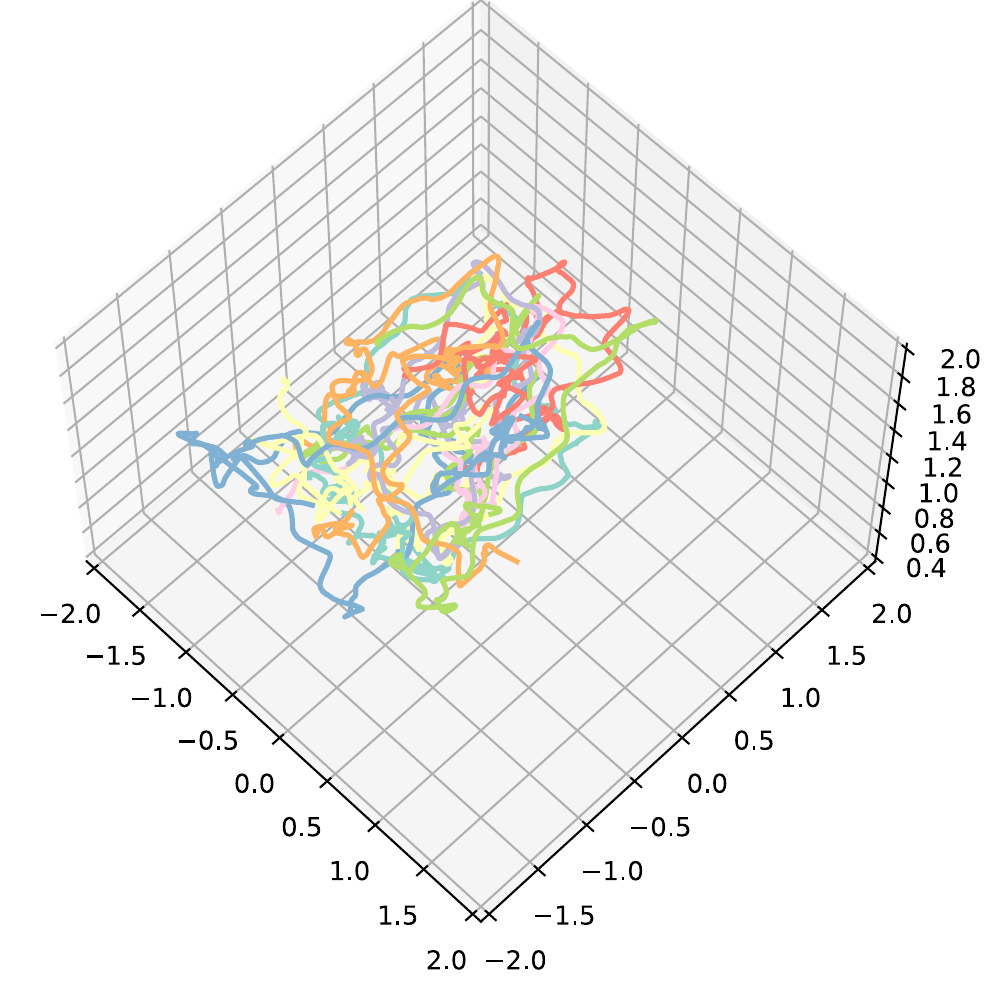}
    }

    \subcaptionbox*{Subject 05}{
        \includegraphics[width=0.05\textwidth]{figures/trajs/sideView05.pdf}
        \includegraphics[width=0.05\textwidth]{figures/trajs/topView05.pdf}
    }
    \hspace{-10pt}
    \subcaptionbox*{Subject 06}{
        \includegraphics[width=0.05\textwidth]{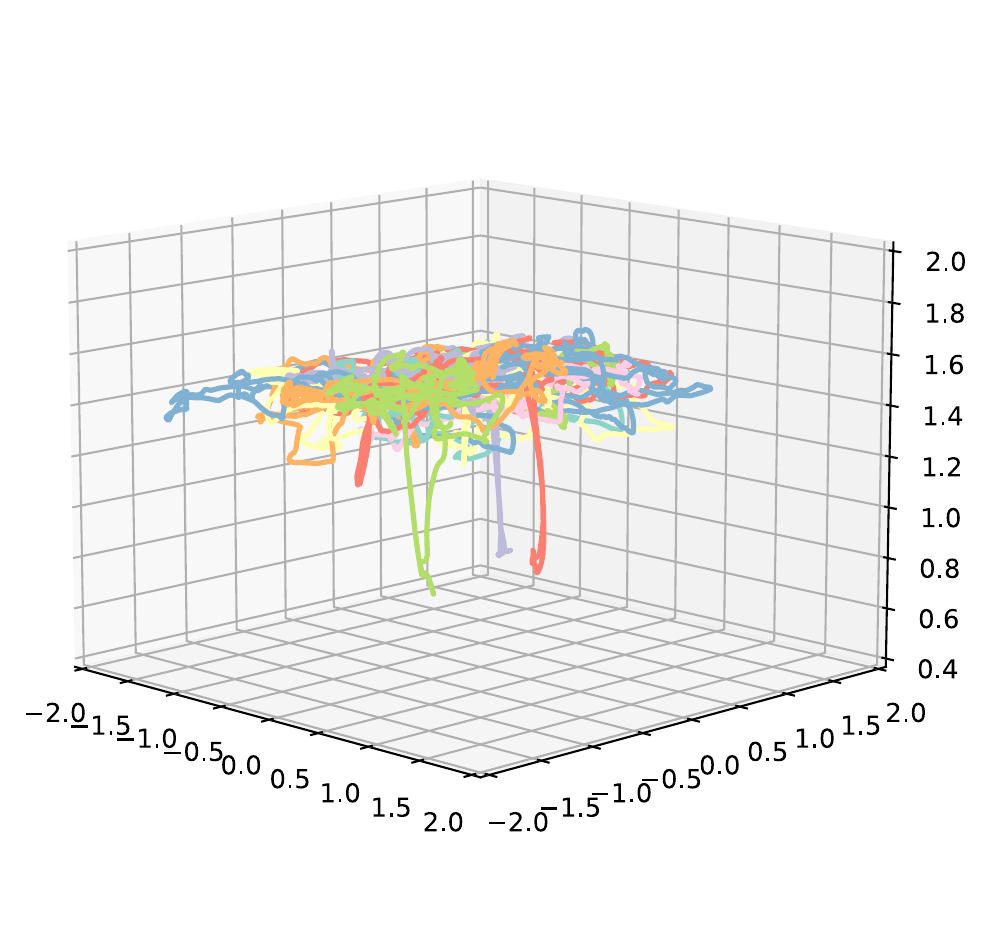}
        \includegraphics[width=0.05\textwidth]{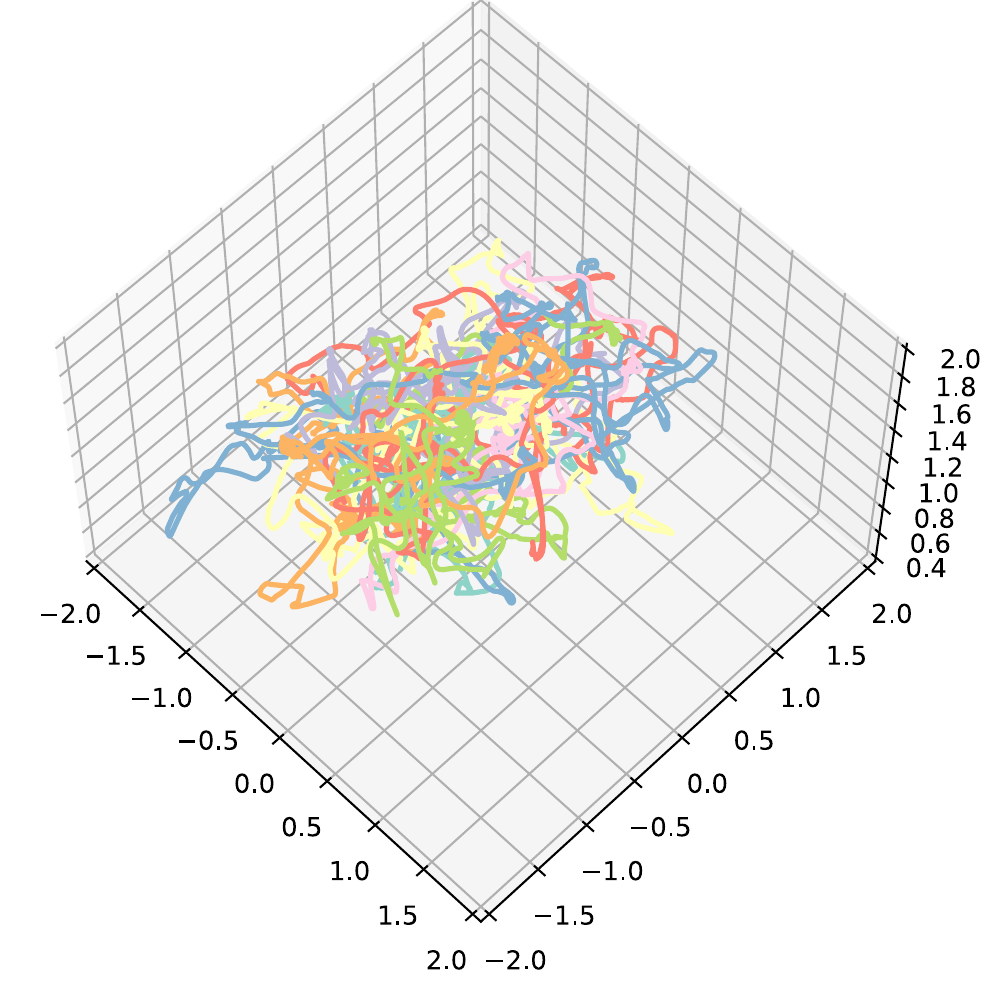}
    }
    \hspace{-10pt}
    \subcaptionbox*{Subject 07}{
        \includegraphics[width=0.05\textwidth]{figures/trajs/sideView07.pdf}
        \includegraphics[width=0.05\textwidth]{figures/trajs/topView07.pdf}
    }
    \hspace{-10pt}
    \subcaptionbox*{Subject 08}{
        \includegraphics[width=0.05\textwidth]{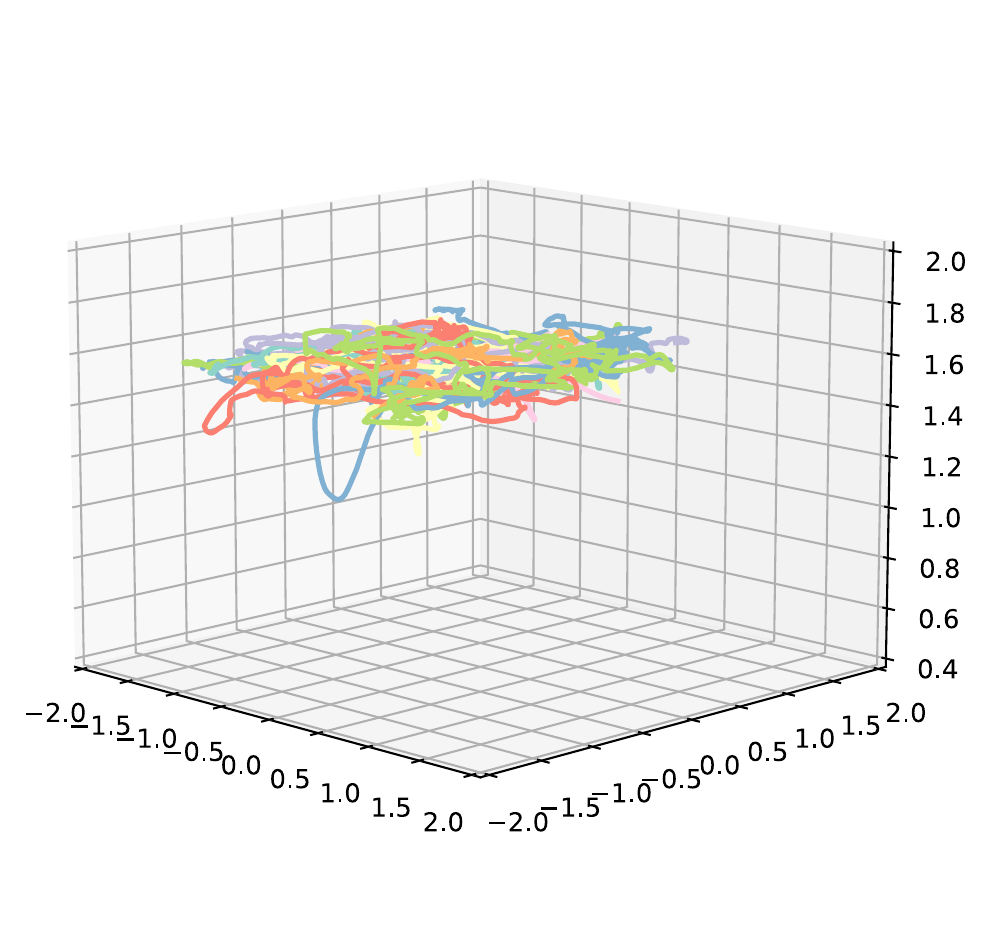}
        \includegraphics[width=0.05\textwidth]{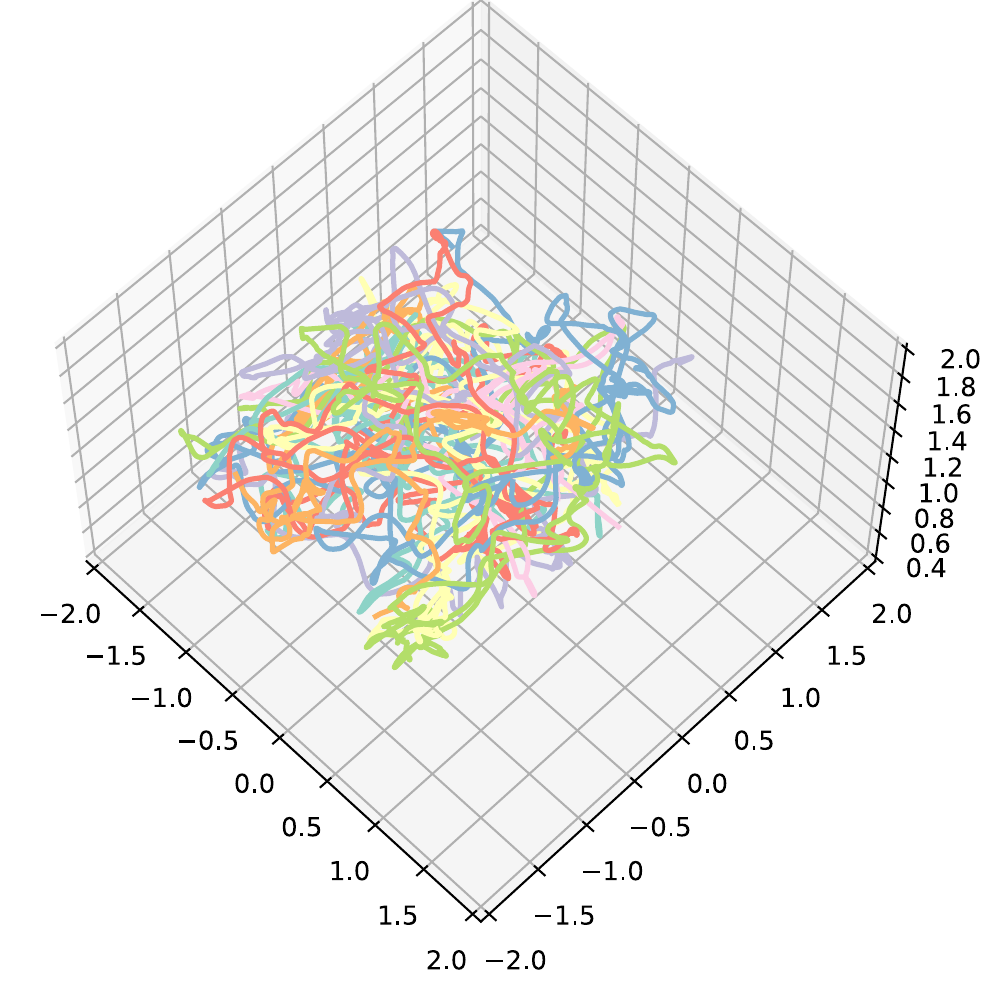}
    }

    \subcaptionbox*{Subject 09}{
        \includegraphics[width=0.05\textwidth]{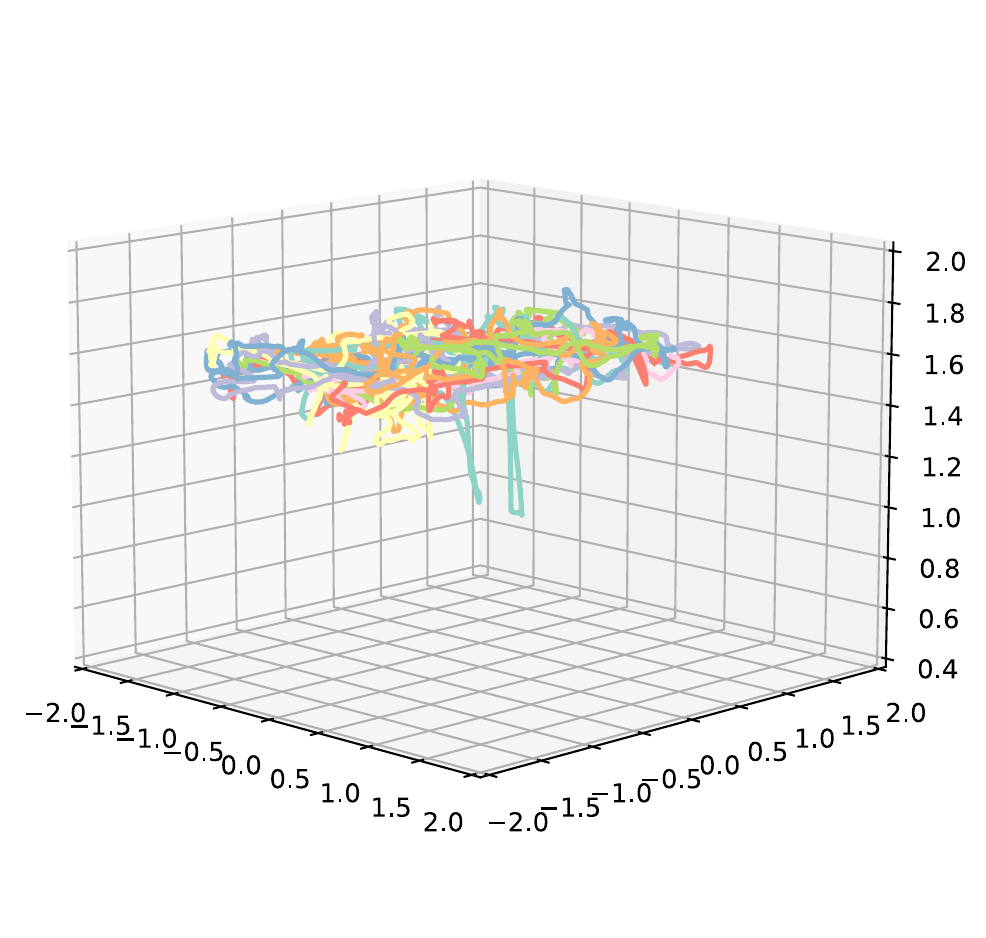}
        \includegraphics[width=0.05\textwidth]{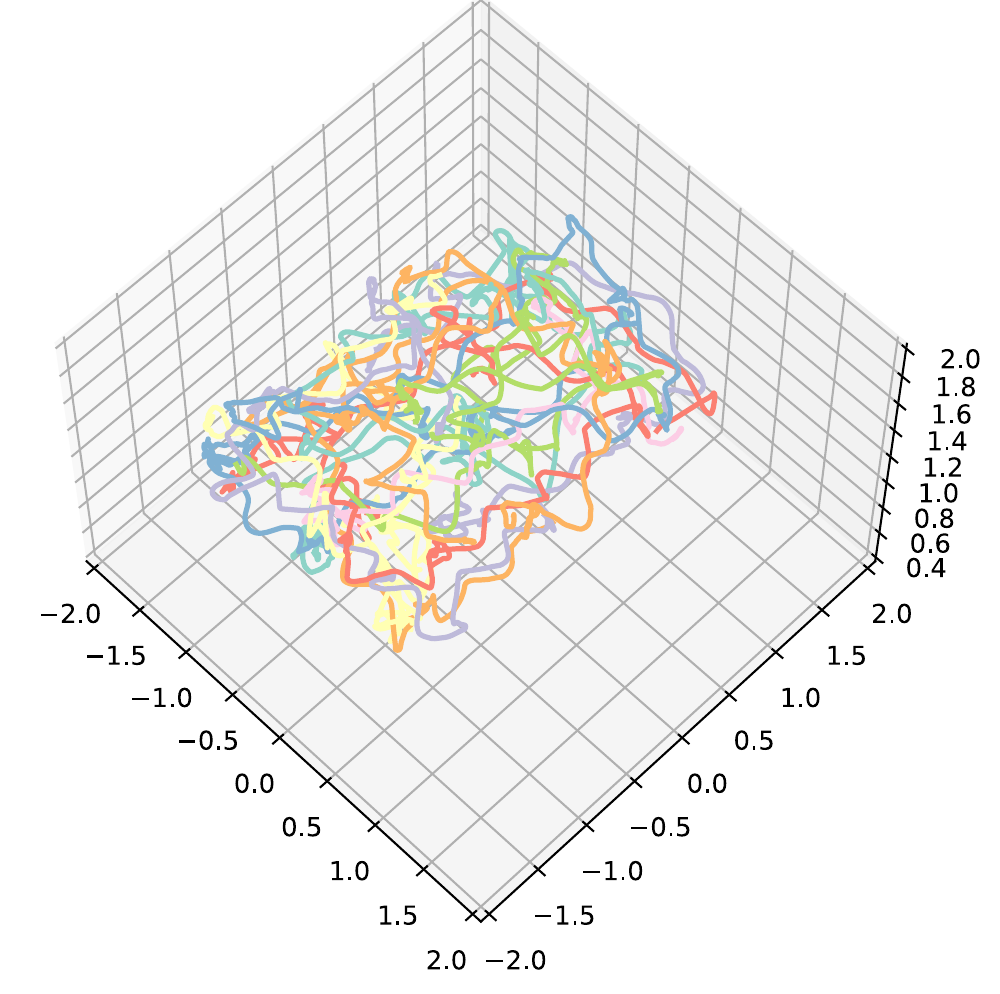}
    }
    \hspace{-10pt}
    \subcaptionbox*{Subject 10}{
        \includegraphics[width=0.05\textwidth]{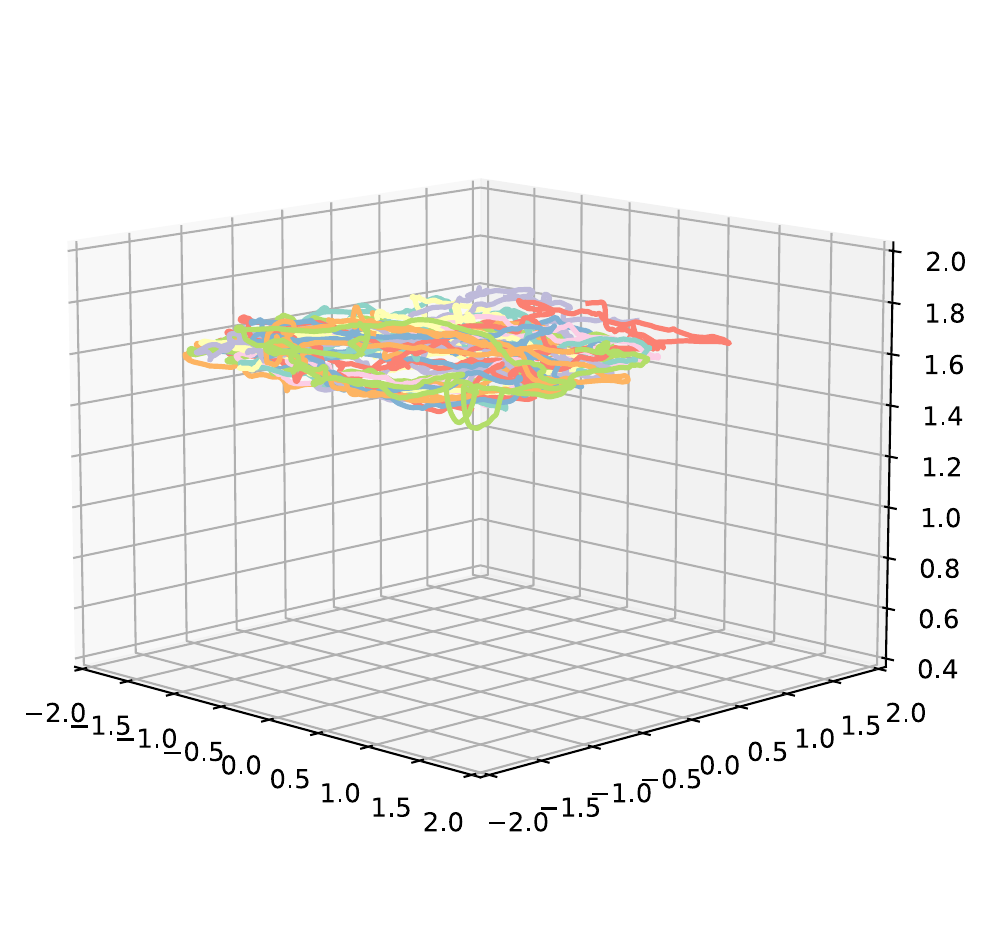}
        \includegraphics[width=0.05\textwidth]{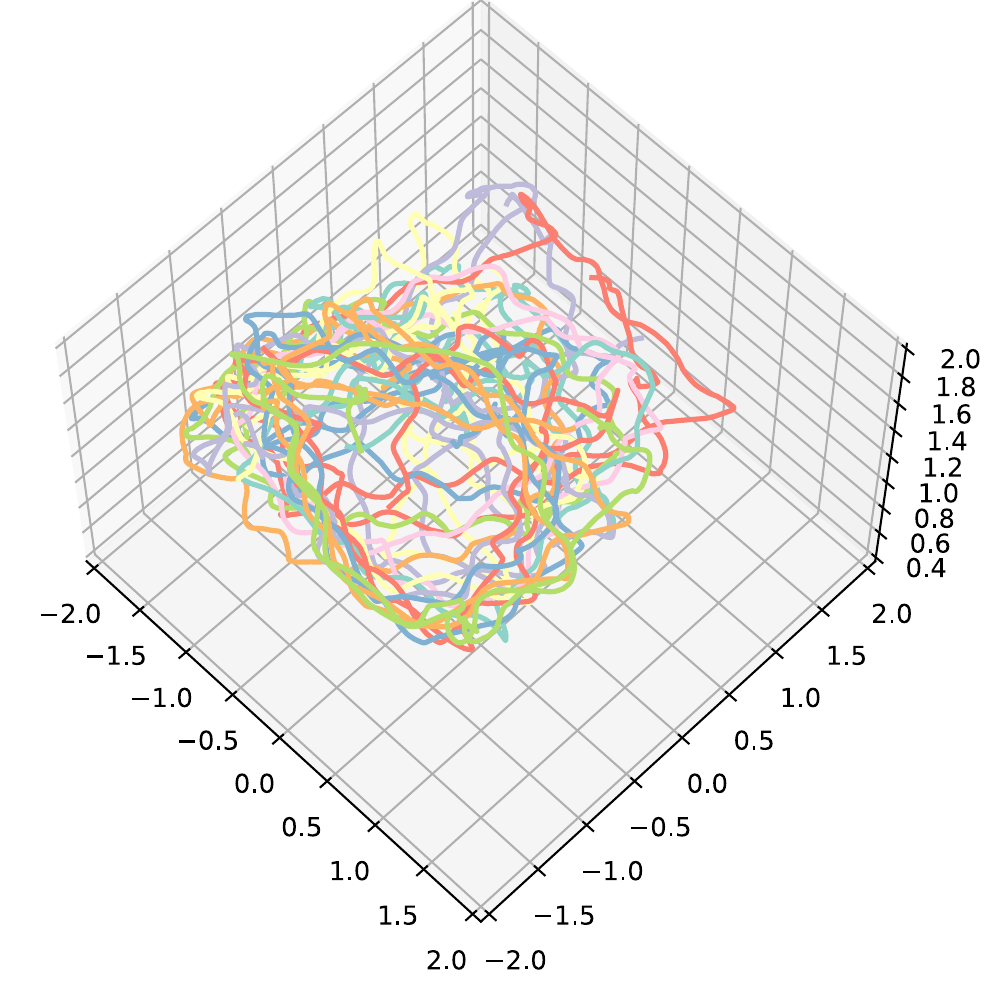}
    }
    \hspace{-10pt}
    \subcaptionbox*{Subject 11}{
        \includegraphics[width=0.05\textwidth]{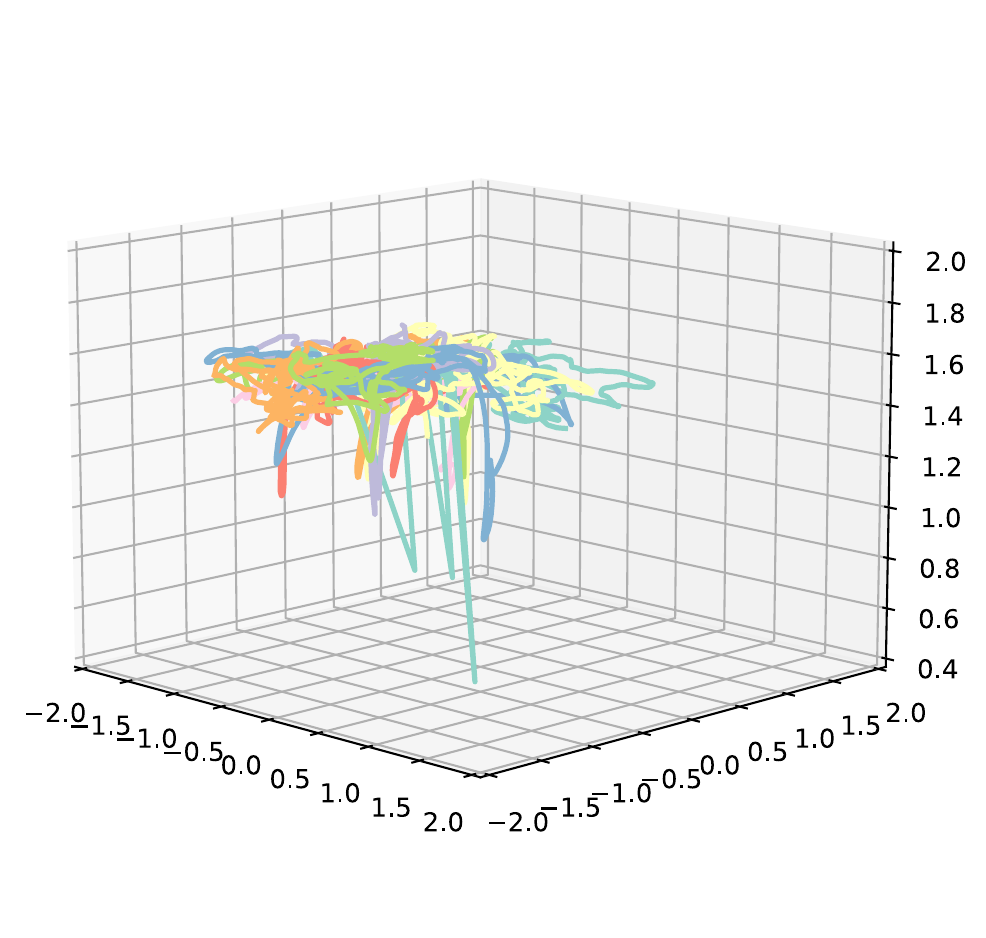}
        \includegraphics[width=0.05\textwidth]{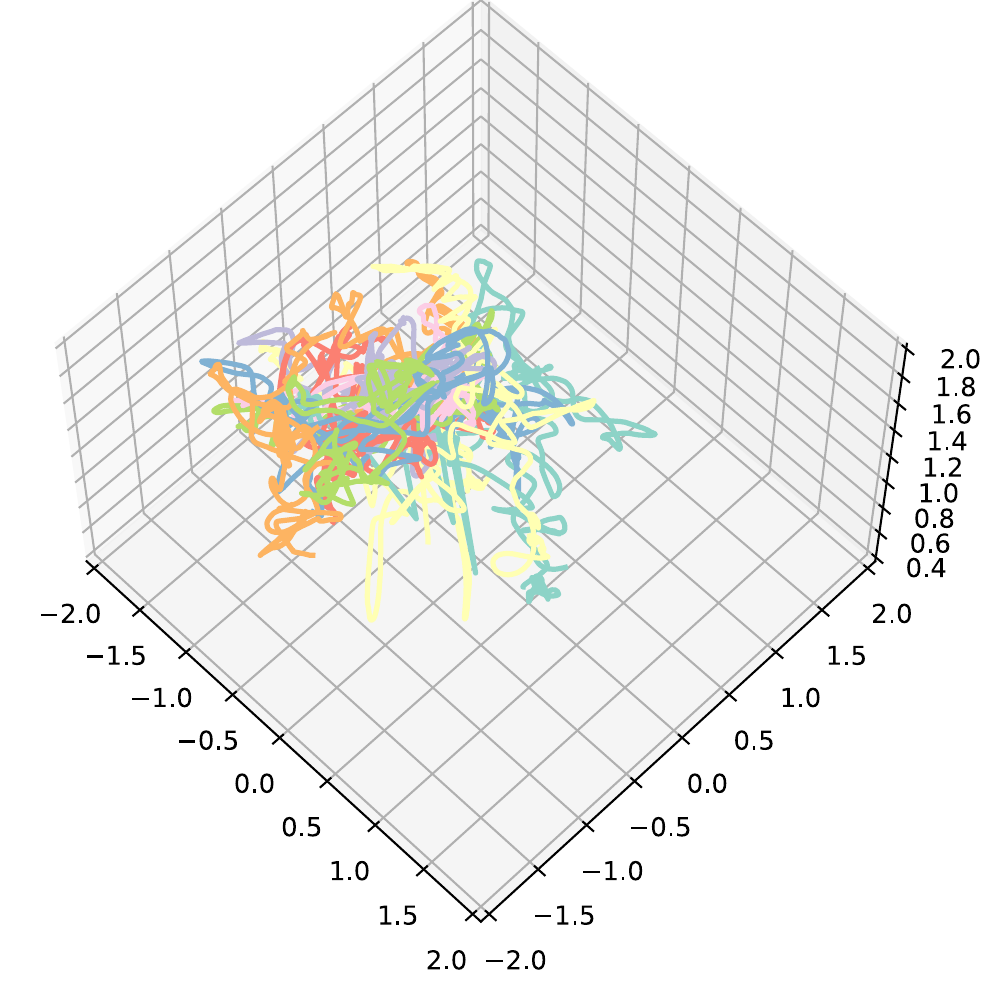}
    }
    \hspace{-10pt}
    \subcaptionbox*{Subject 12}{
        \includegraphics[width=0.05\textwidth]{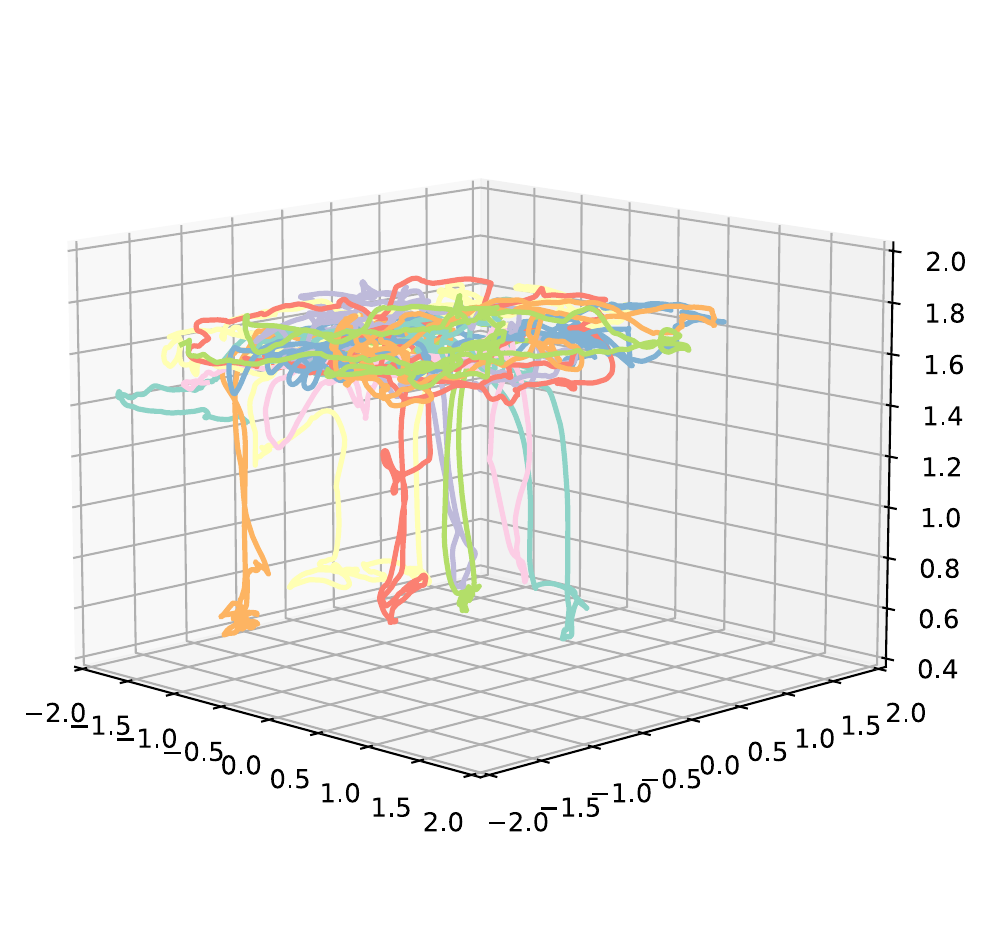}
        \includegraphics[width=0.05\textwidth]{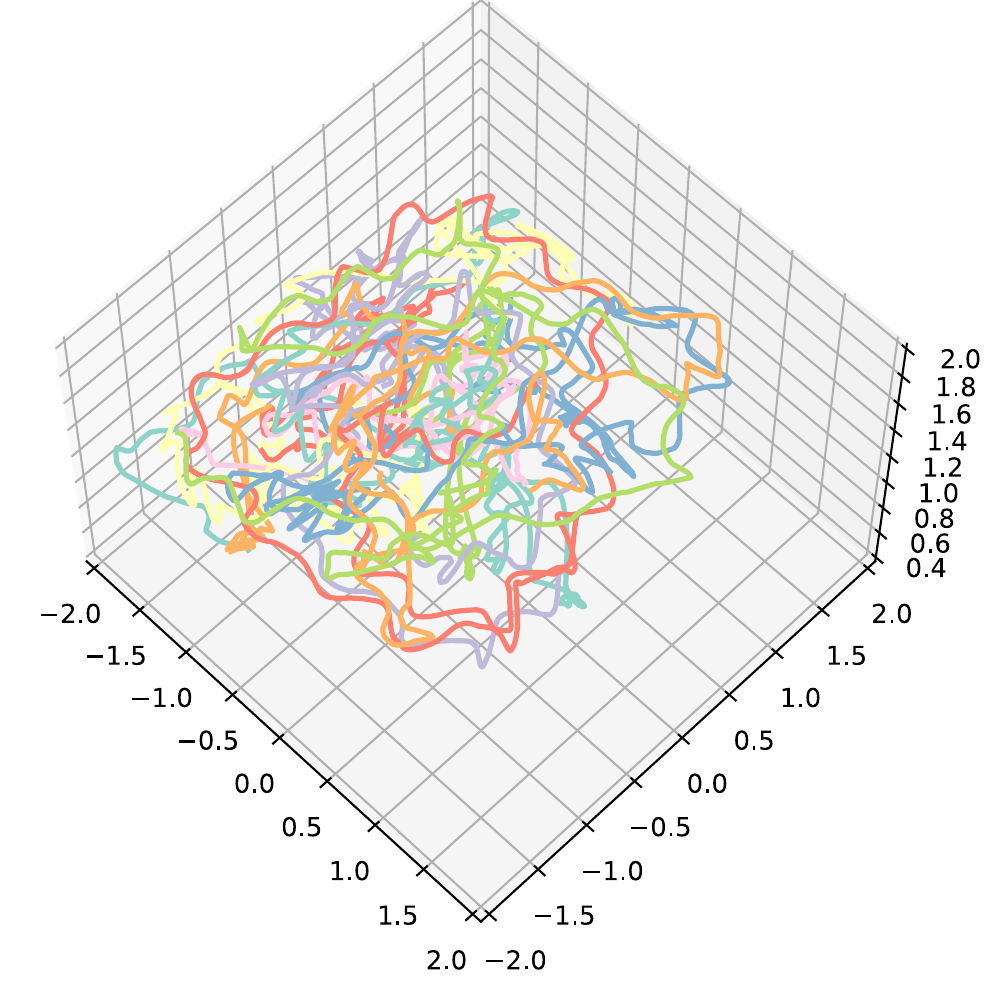}
    }

    \subcaptionbox*{Subject 13}{
        \includegraphics[width=0.05\textwidth]{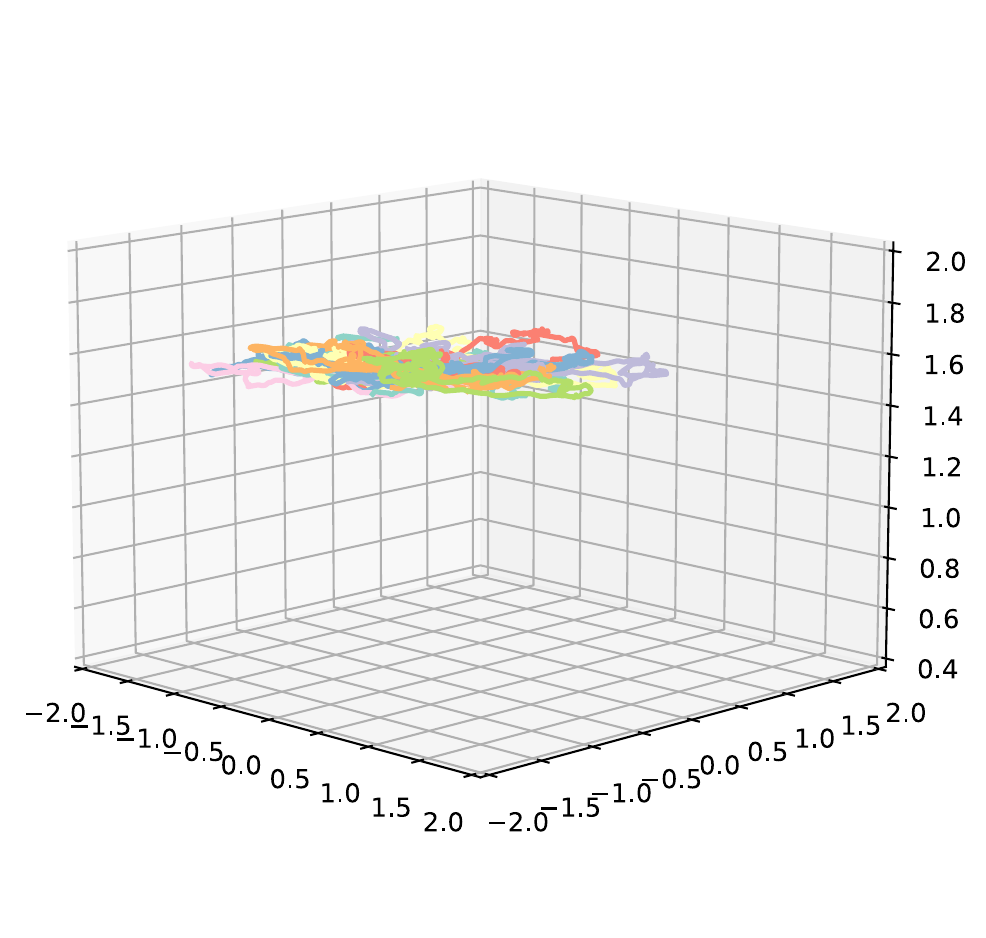}
        \includegraphics[width=0.05\textwidth]{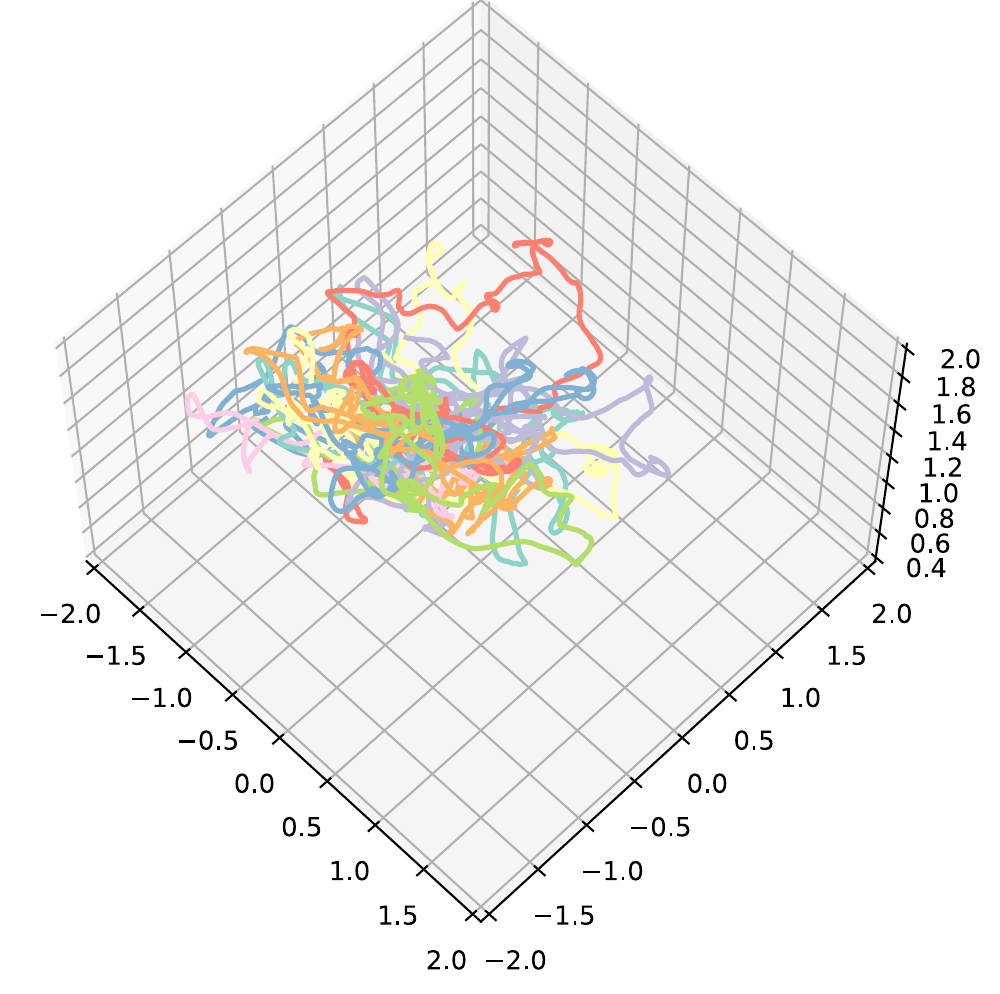}
    }
    \hspace{-10pt}
    \subcaptionbox*{Subject 14}{
        \includegraphics[width=0.05\textwidth]{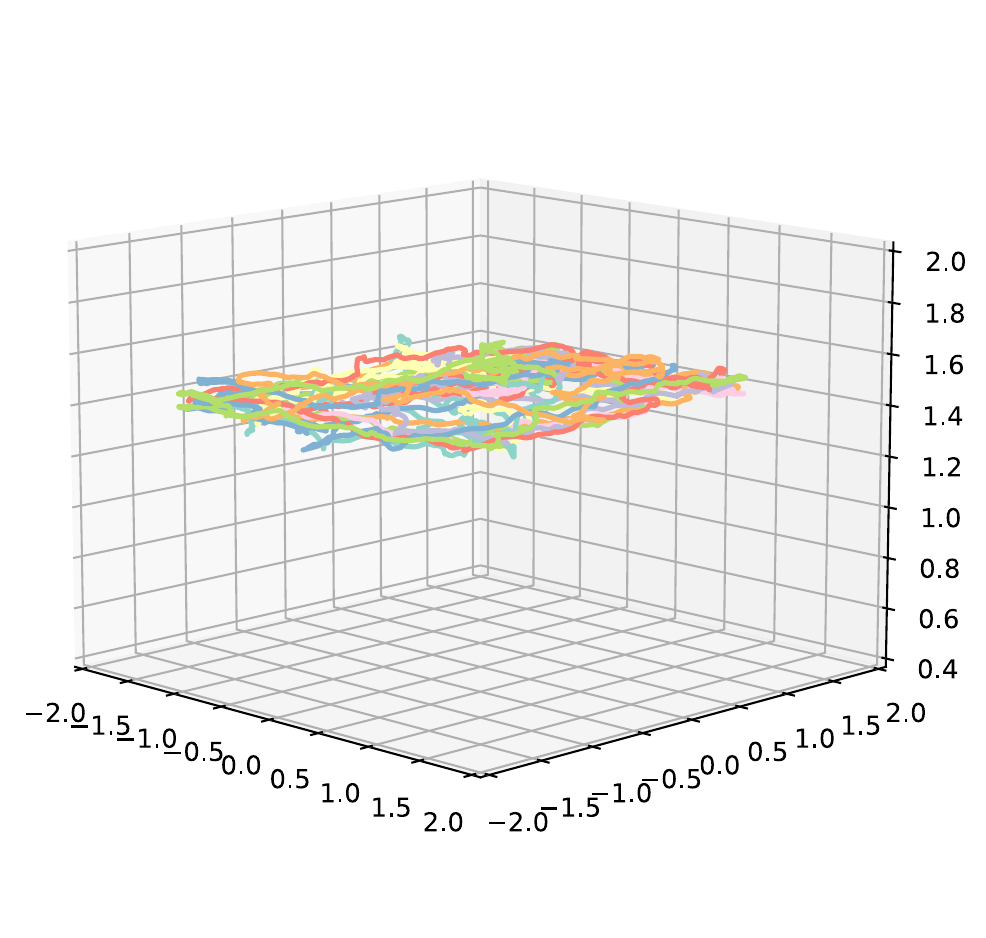}
        \includegraphics[width=0.05\textwidth]{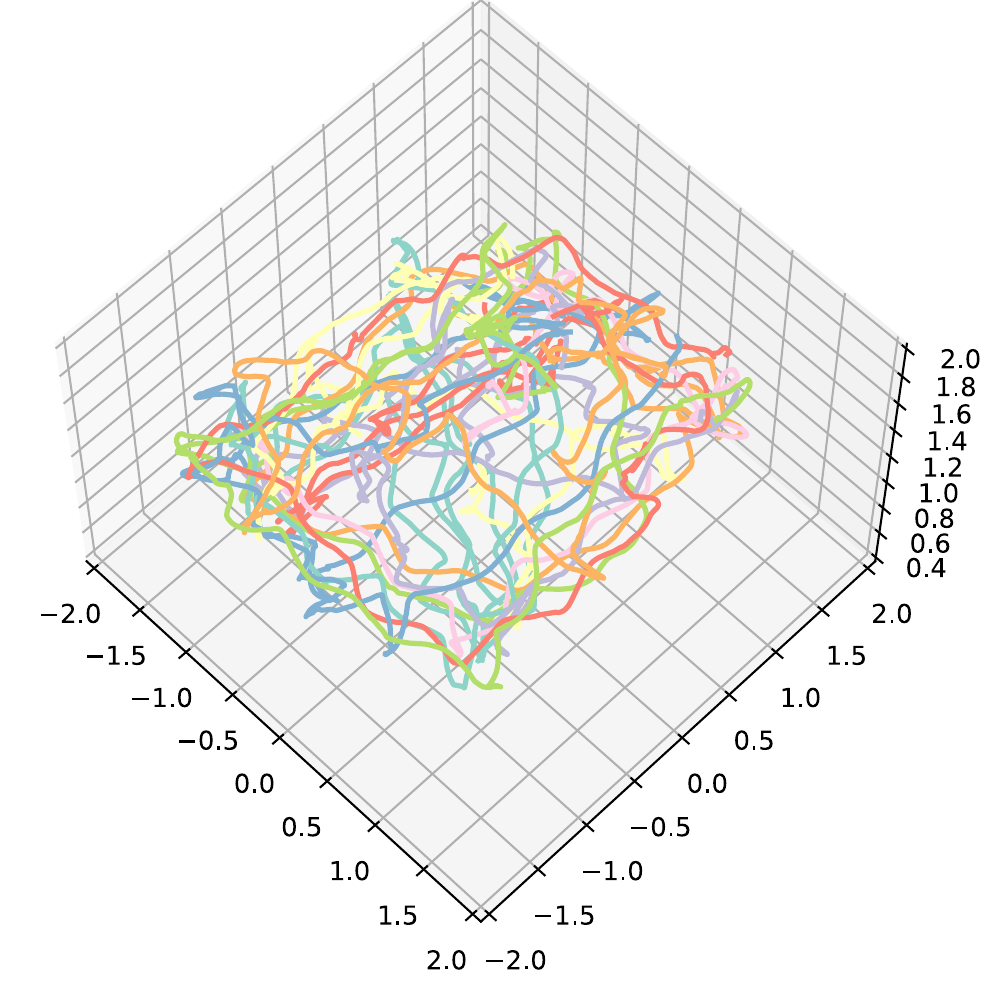}
    }
    \hspace{-10pt}
    \subcaptionbox*{Subject 15}{
        \includegraphics[width=0.05\textwidth]{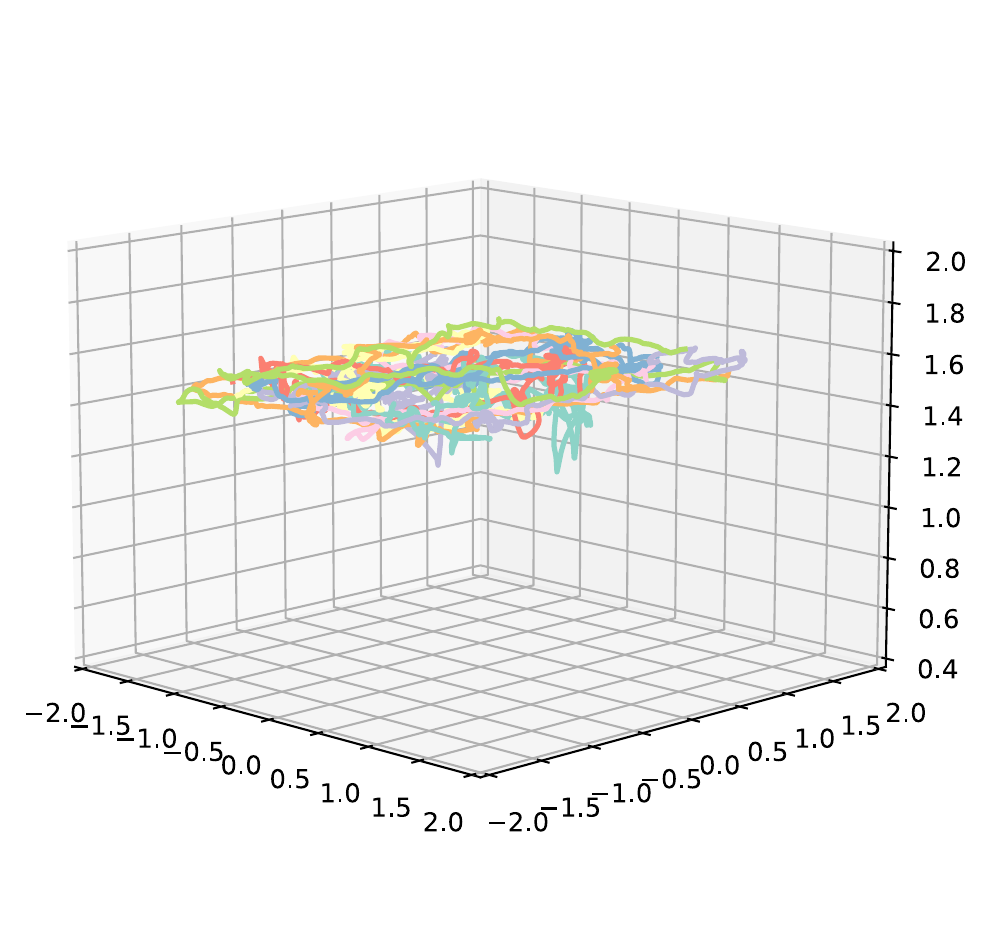}
        \includegraphics[width=0.05\textwidth]{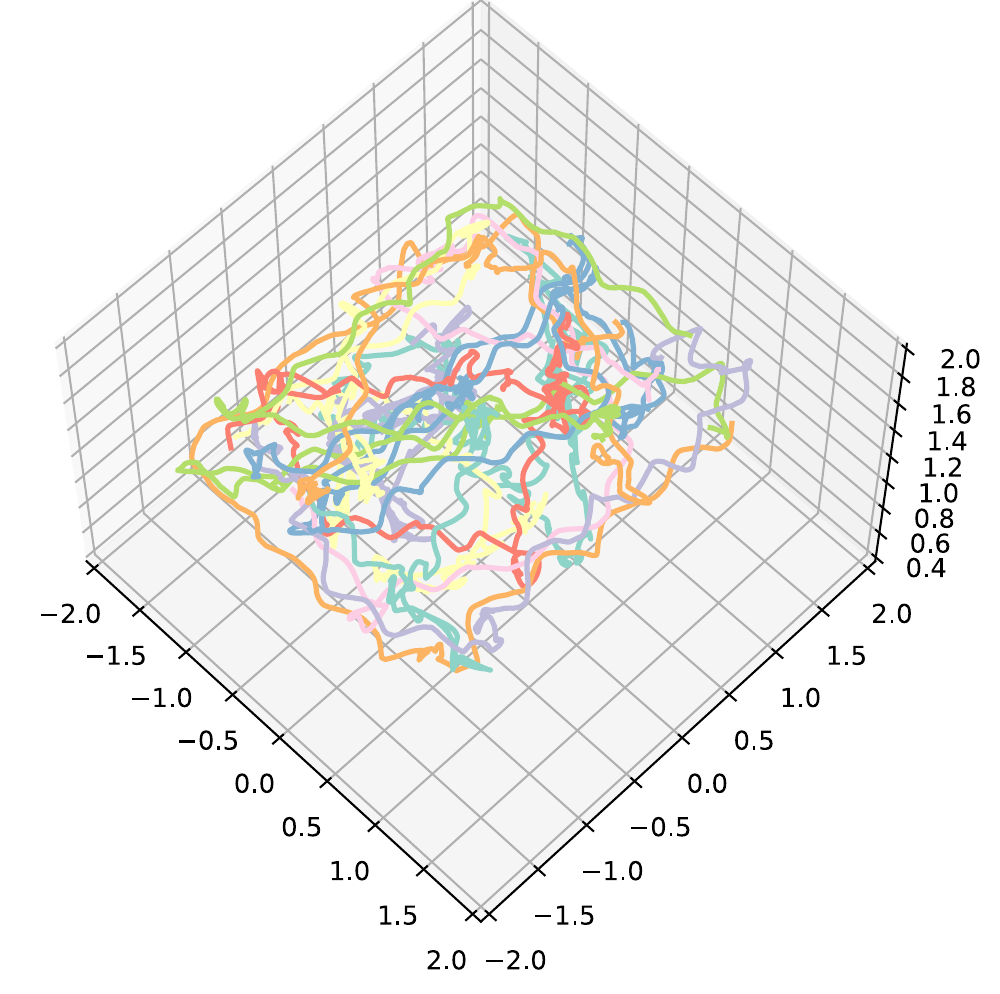}
    }
    \hspace{-10pt}
    \subcaptionbox*{Subject 16}{
        \includegraphics[width=0.05\textwidth]{figures/trajs/sideView16.pdf}
        \includegraphics[width=0.05\textwidth]{figures/trajs/topView16.pdf}
    }

    \subcaptionbox*{Subject 17}{
        \includegraphics[width=0.05\textwidth]{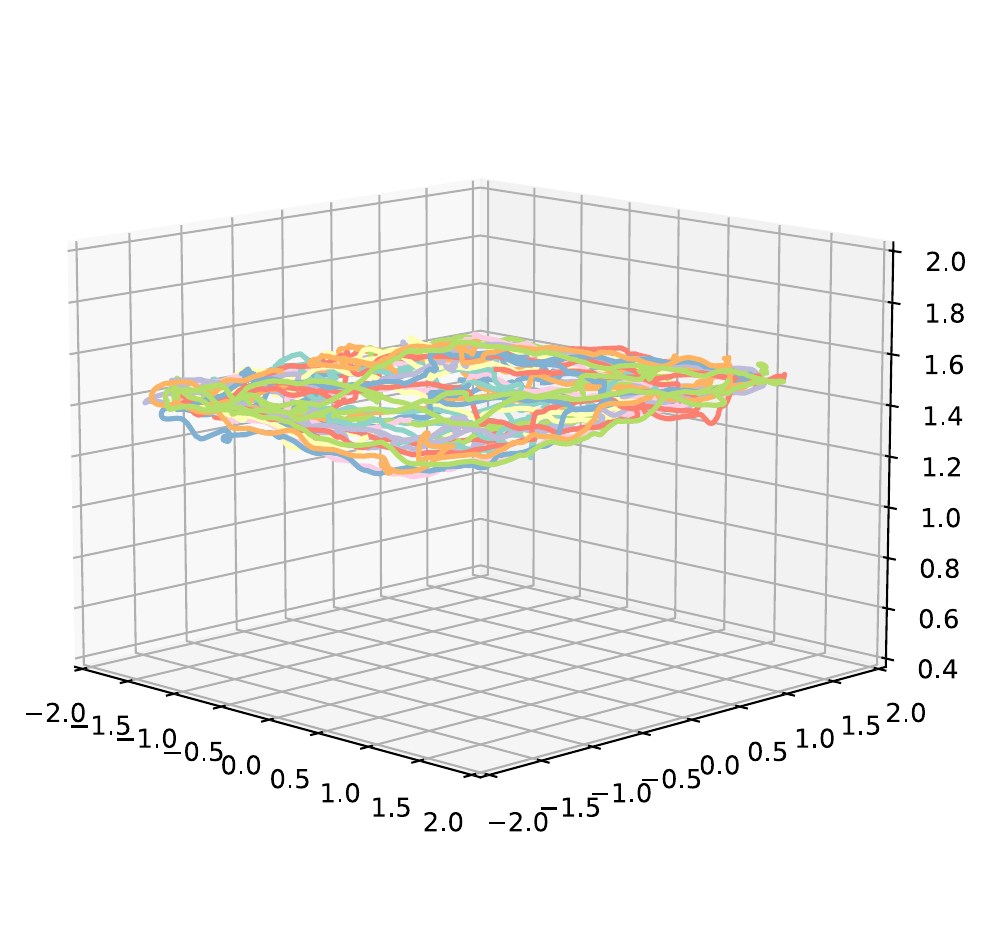}
        \includegraphics[width=0.05\textwidth]{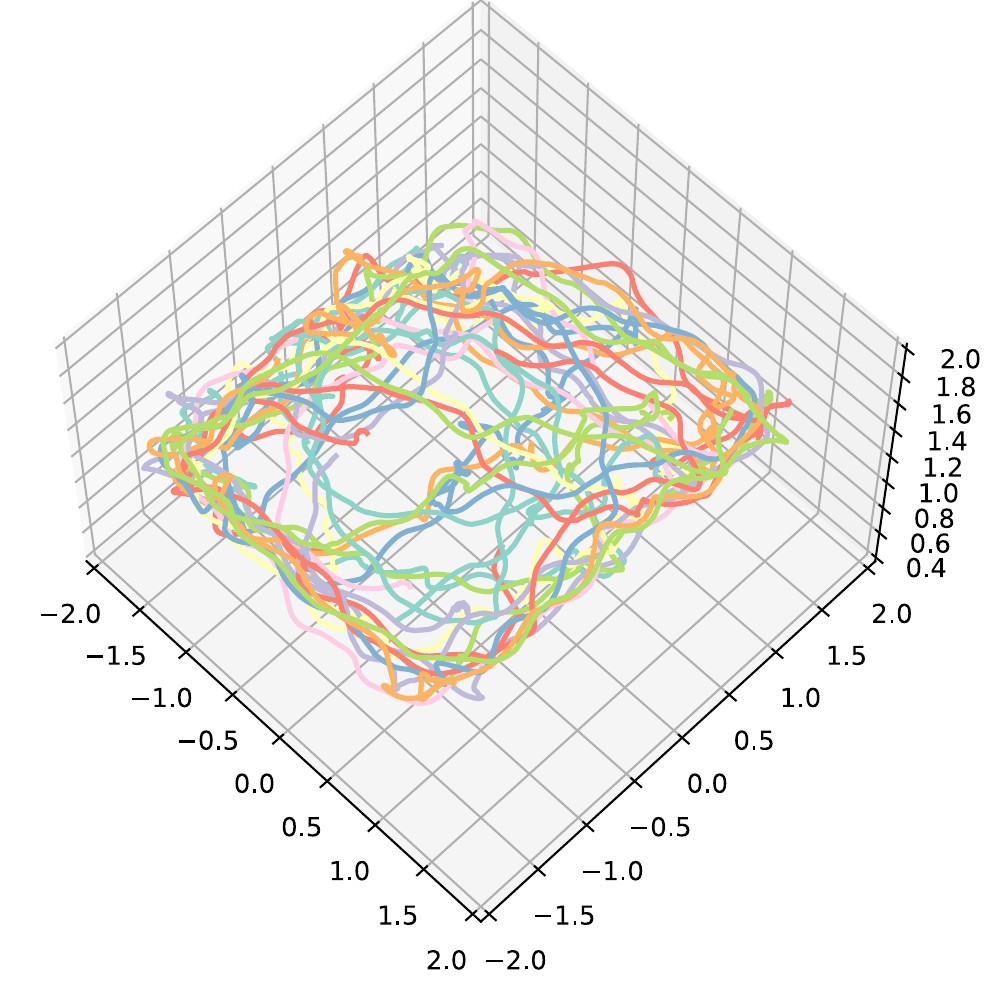}
    }
    \hspace{-10pt}
    \subcaptionbox*{Subject 18}{
        \includegraphics[width=0.05\textwidth]{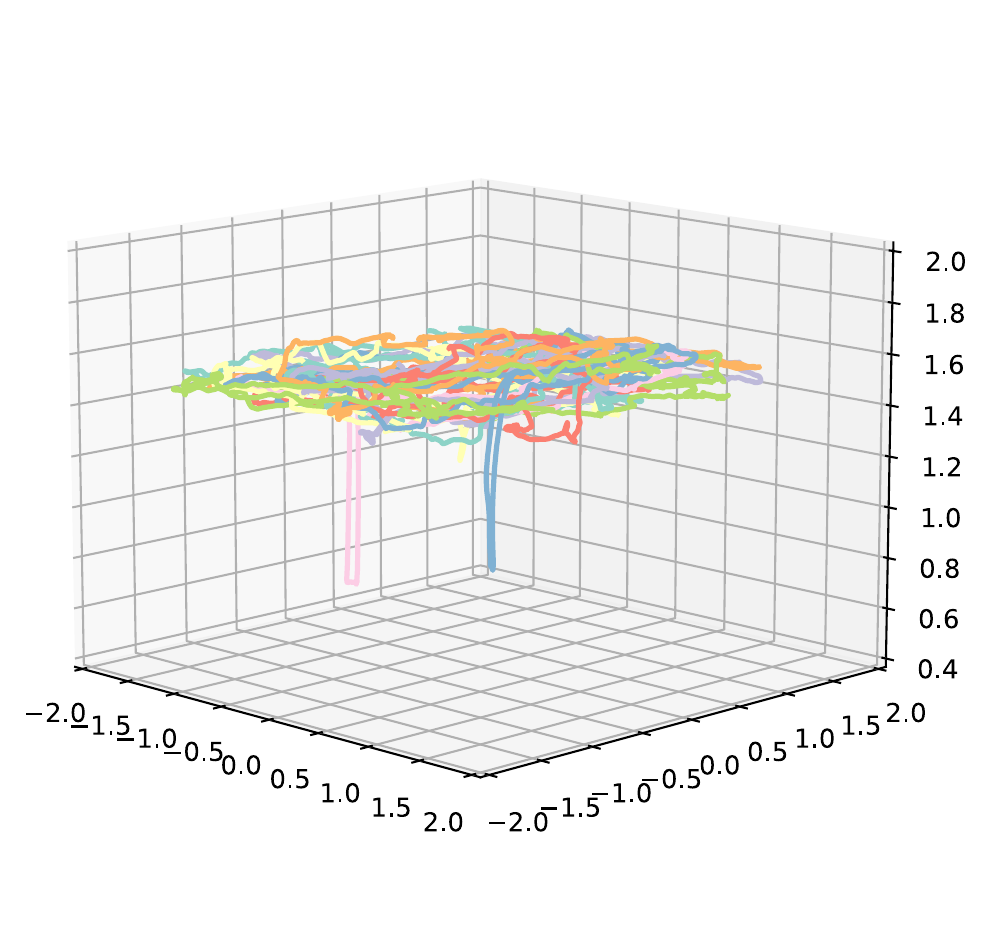}
        \includegraphics[width=0.05\textwidth]{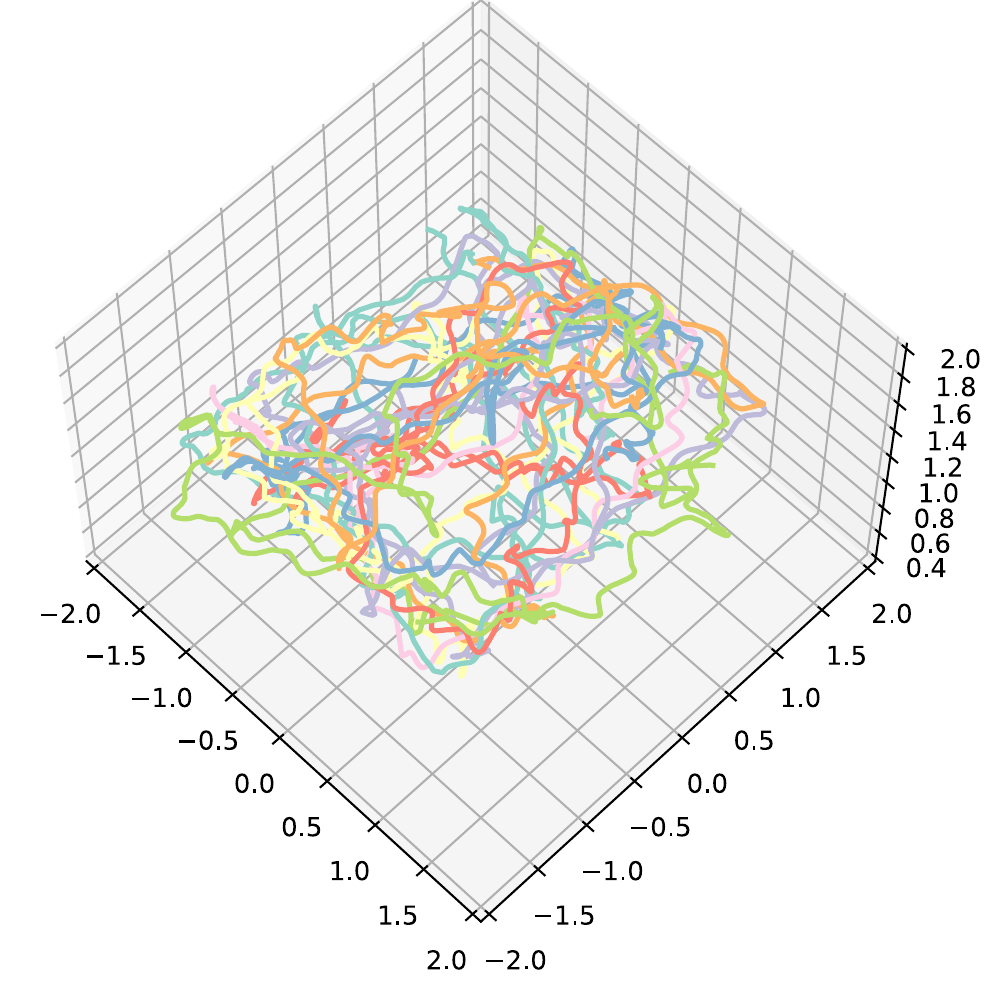}
    }
    \hspace{-10pt}
    \subcaptionbox*{Subject 19}{
        \includegraphics[width=0.05\textwidth]{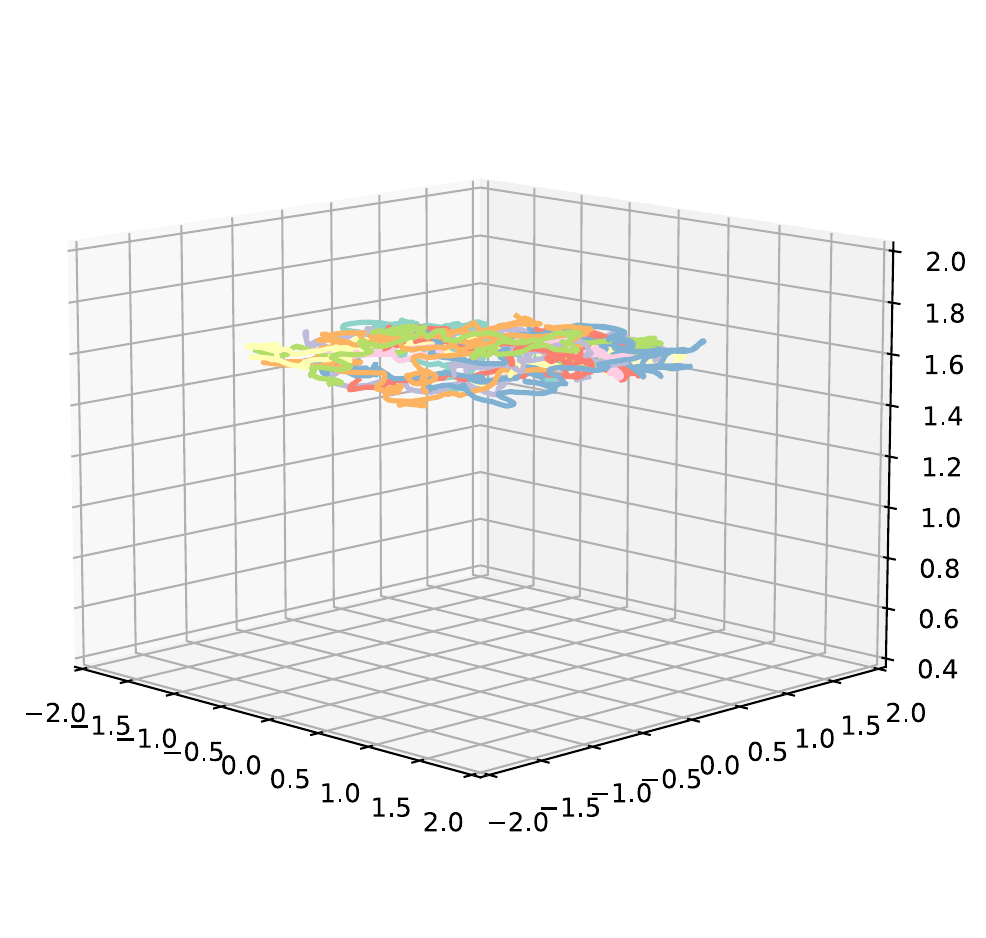}
        \includegraphics[width=0.05\textwidth]{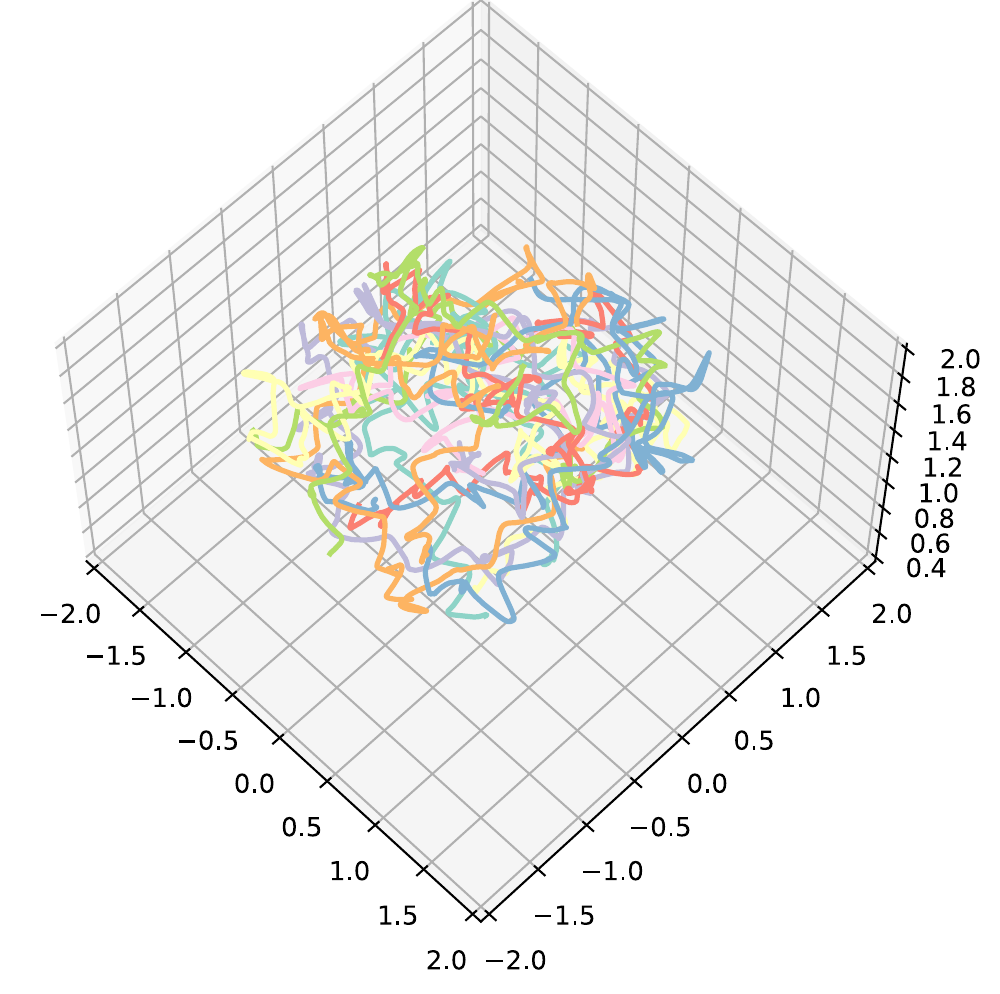}
    }
    \hspace{-10pt}
    \subcaptionbox*{Subject 20}{
        \includegraphics[width=0.05\textwidth]{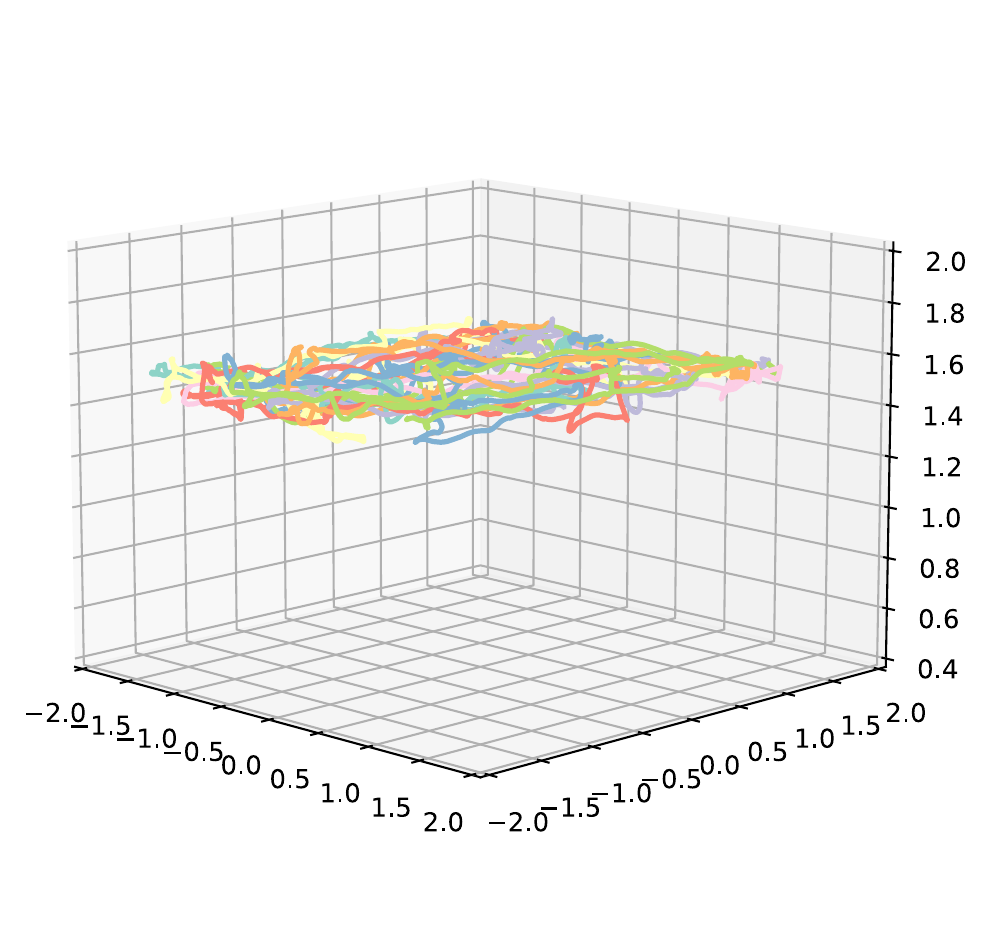}
        \includegraphics[width=0.05\textwidth]{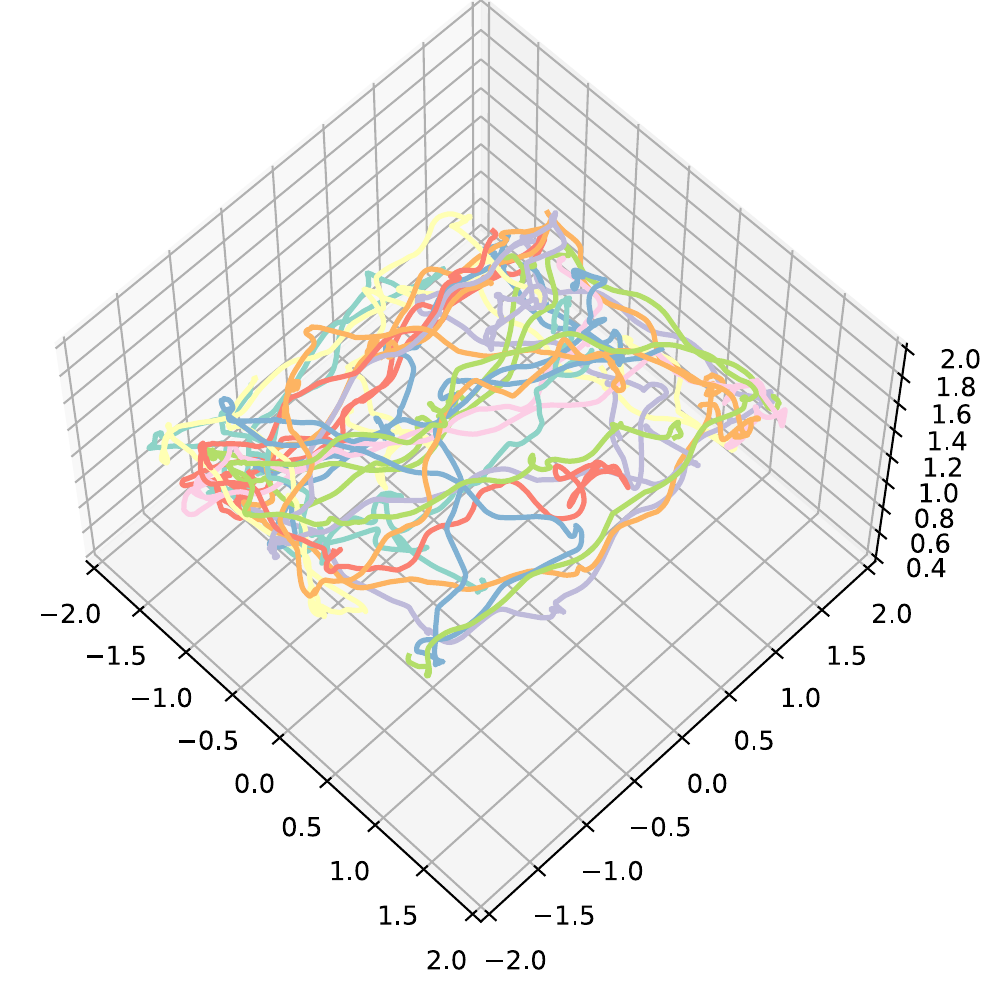}
    }

    \subcaptionbox*{Subject 21}{
        \includegraphics[width=0.05\textwidth]{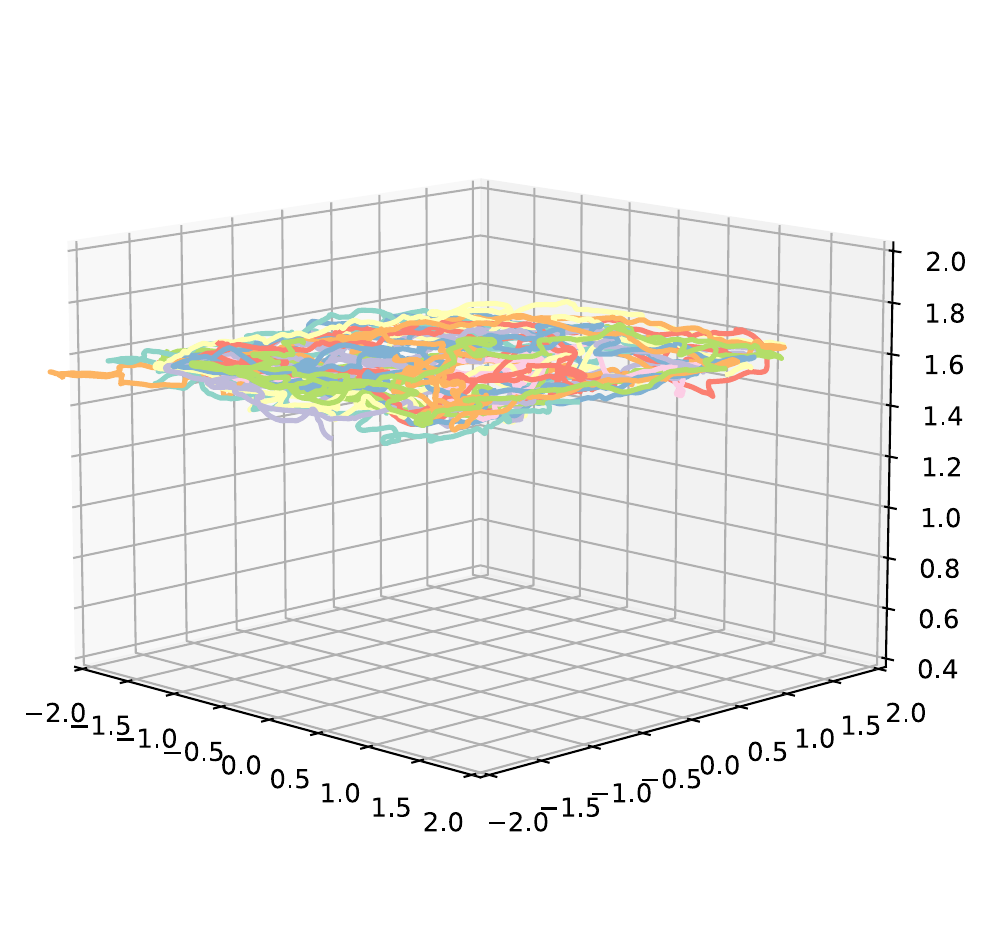}
        \includegraphics[width=0.05\textwidth]{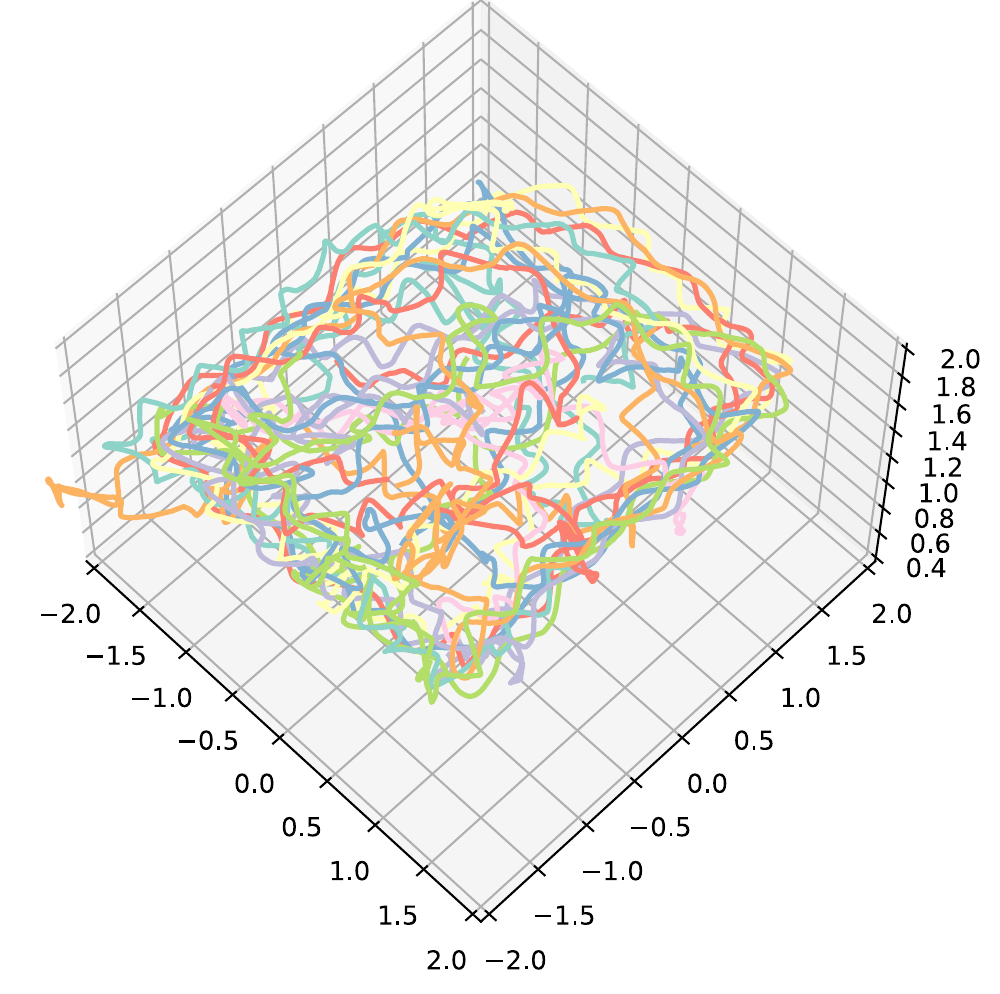}
    }
    \hspace{-10pt}
    \subcaptionbox*{Subject 22}{
        \includegraphics[width=0.05\textwidth]{figures/trajs/sideView22.pdf}
        \includegraphics[width=0.05\textwidth]{figures/trajs/topView22.pdf}
    }
    \hspace{-10pt}
    \subcaptionbox*{Subject 23}{
        \includegraphics[width=0.05\textwidth]{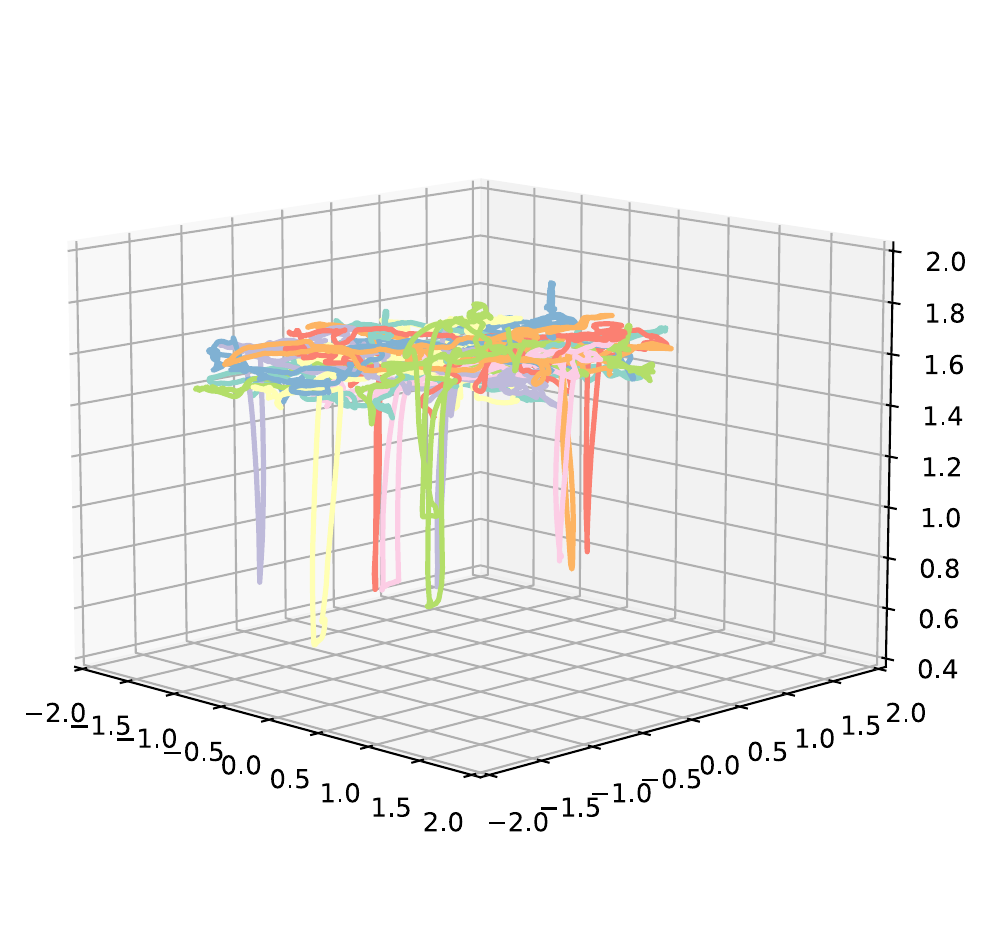}
        \includegraphics[width=0.05\textwidth]{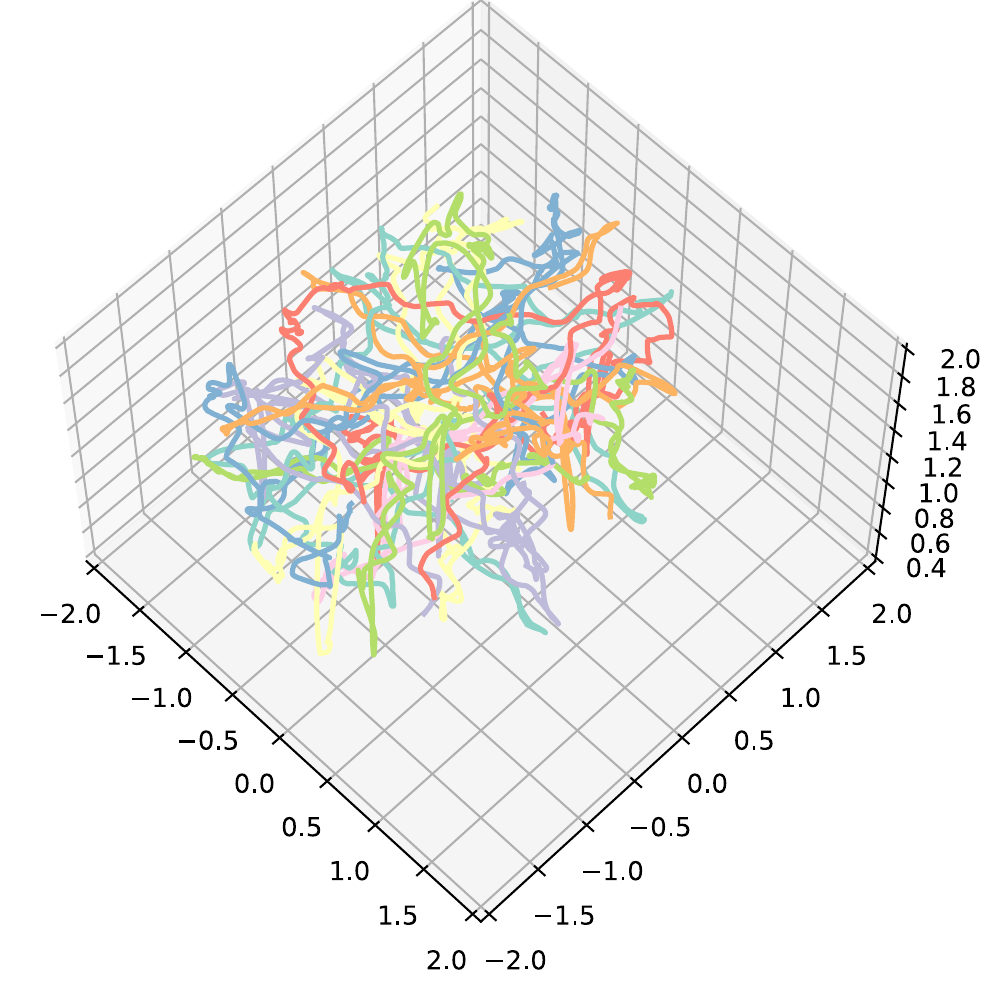}
    }
    \hspace{-10pt}
    \subcaptionbox*{Subject 24}{
        \includegraphics[width=0.05\textwidth]{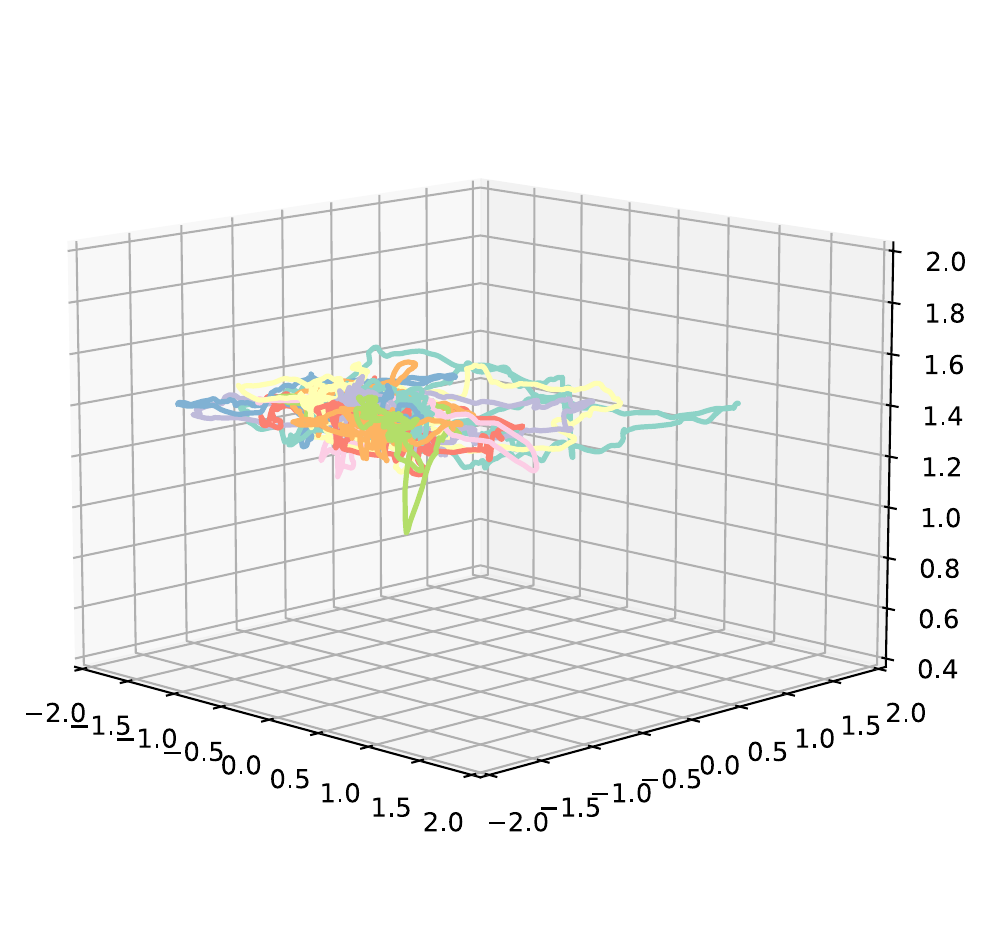}
        \includegraphics[width=0.05\textwidth]{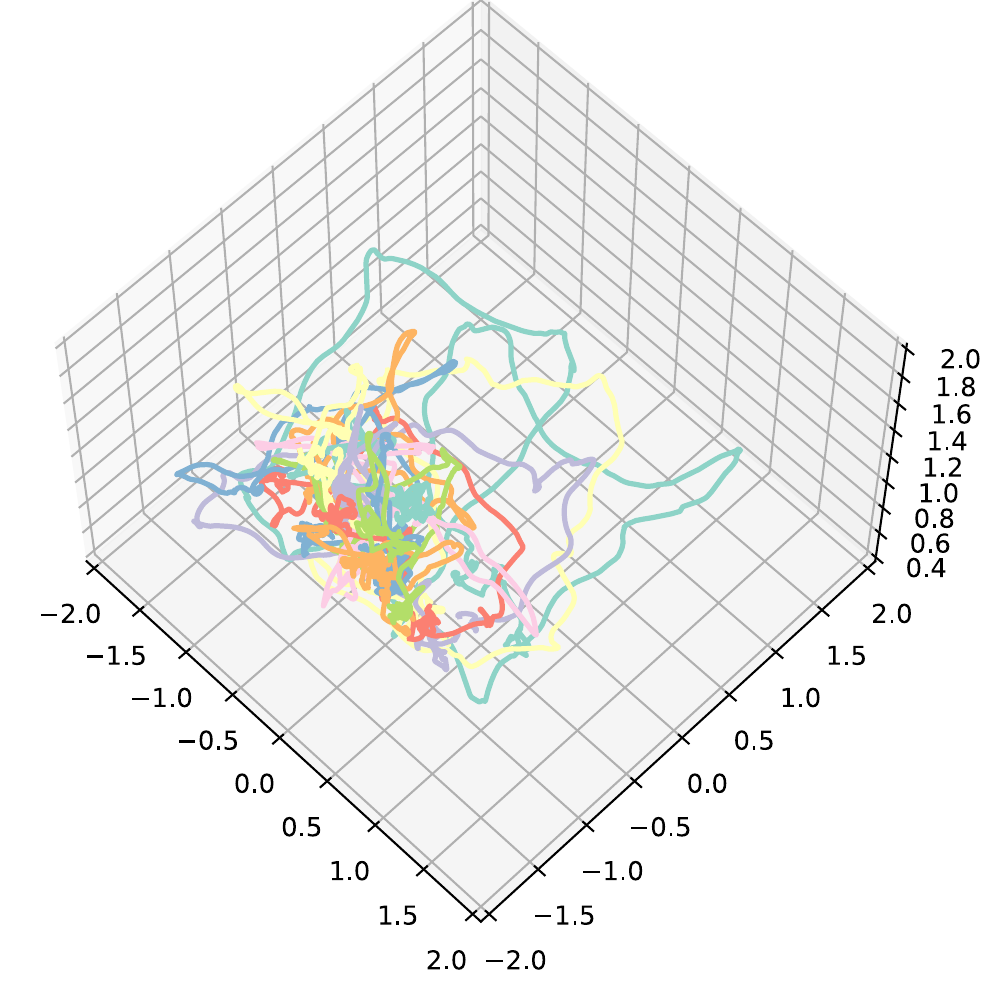}
    }

    \subcaptionbox*{Subject 25}{
        \includegraphics[width=0.05\textwidth]{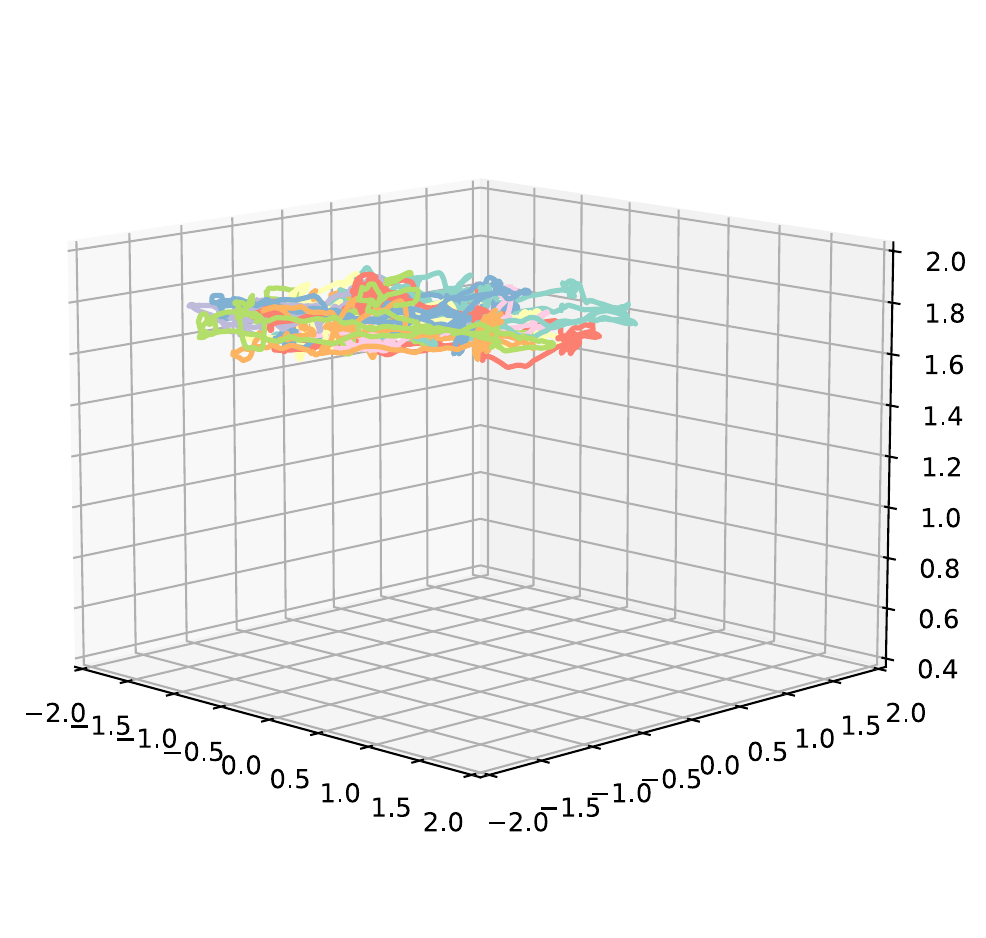}
        \includegraphics[width=0.05\textwidth]{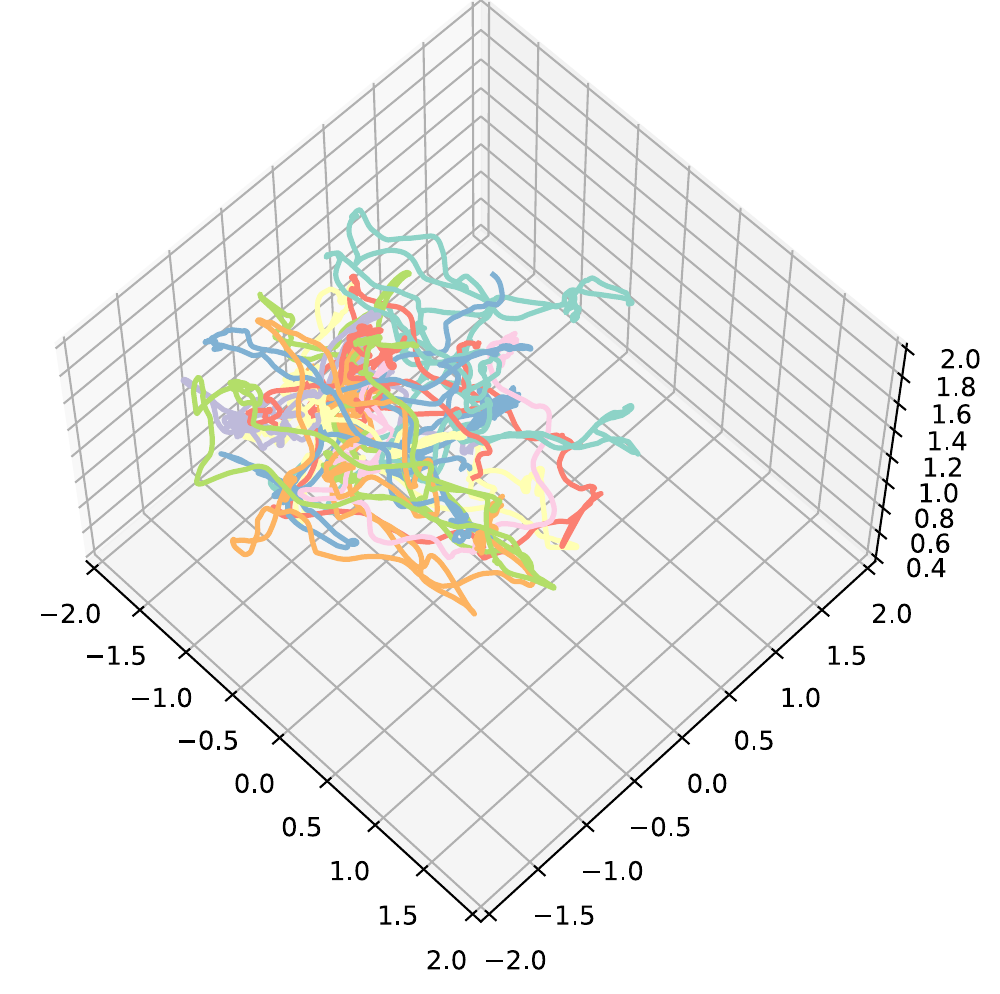}
    }
    \hspace{-10pt}
    \subcaptionbox*{Subject 26}{
        \includegraphics[width=0.05\textwidth]{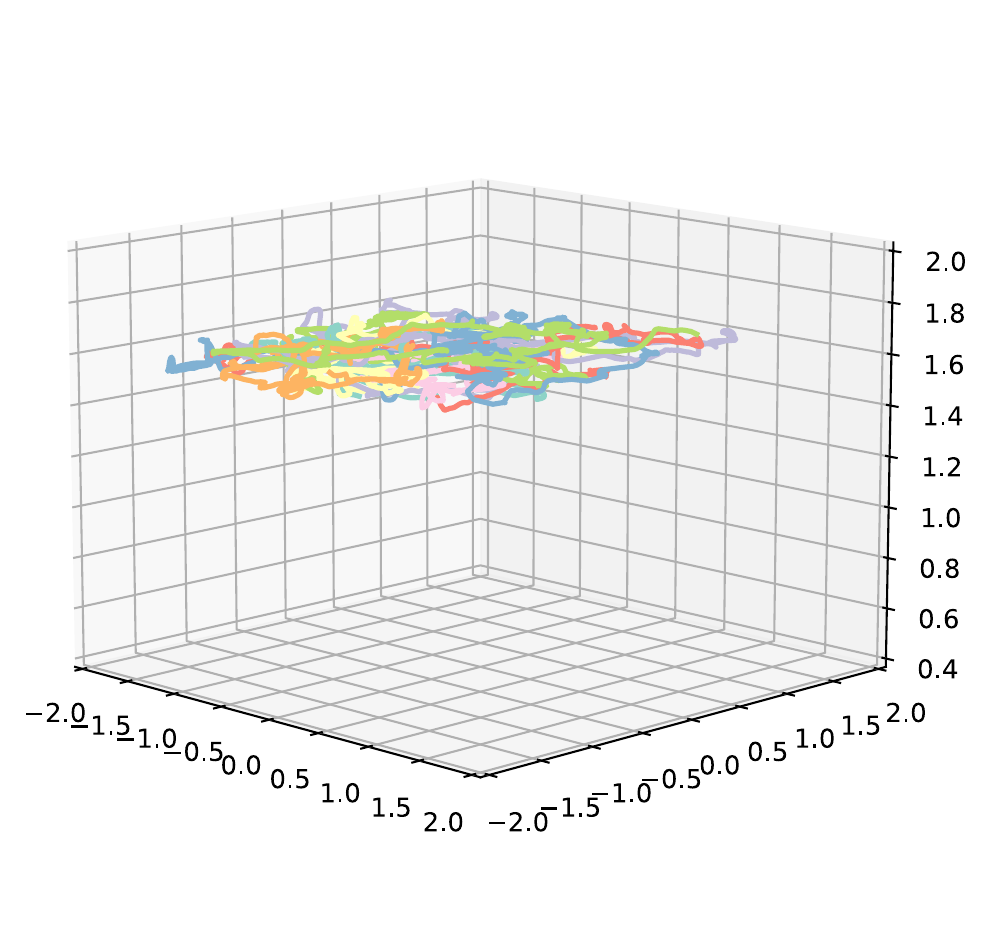}
        \includegraphics[width=0.05\textwidth]{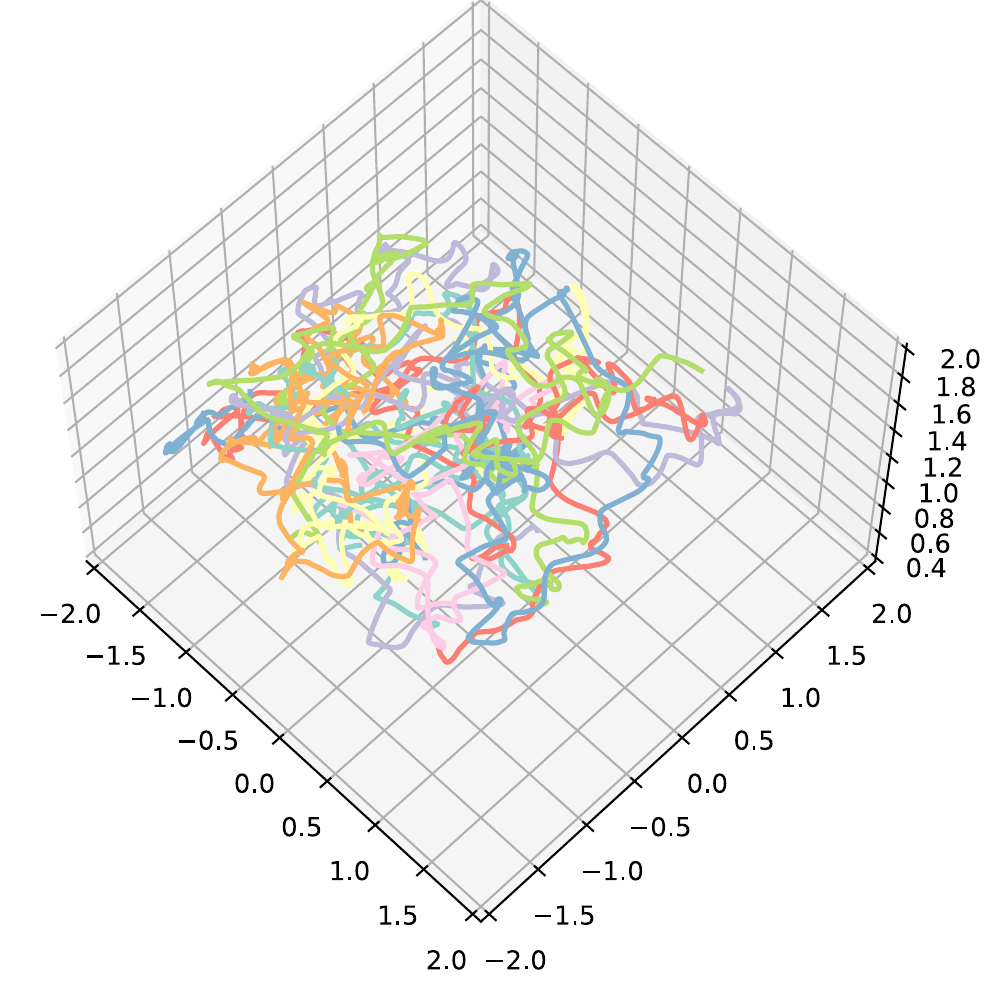}
    }
    \hspace{-10pt}
    \subcaptionbox*{Subject 27}{
        \includegraphics[width=0.05\textwidth]{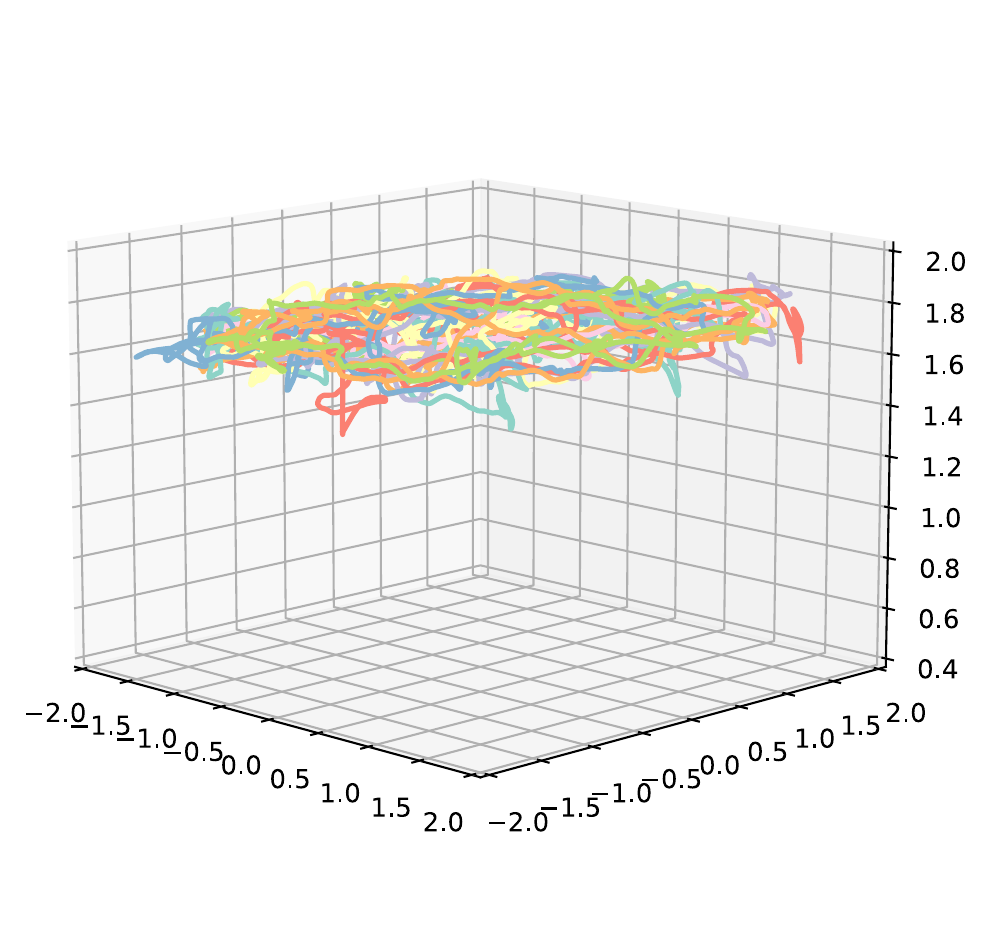}
        \includegraphics[width=0.05\textwidth]{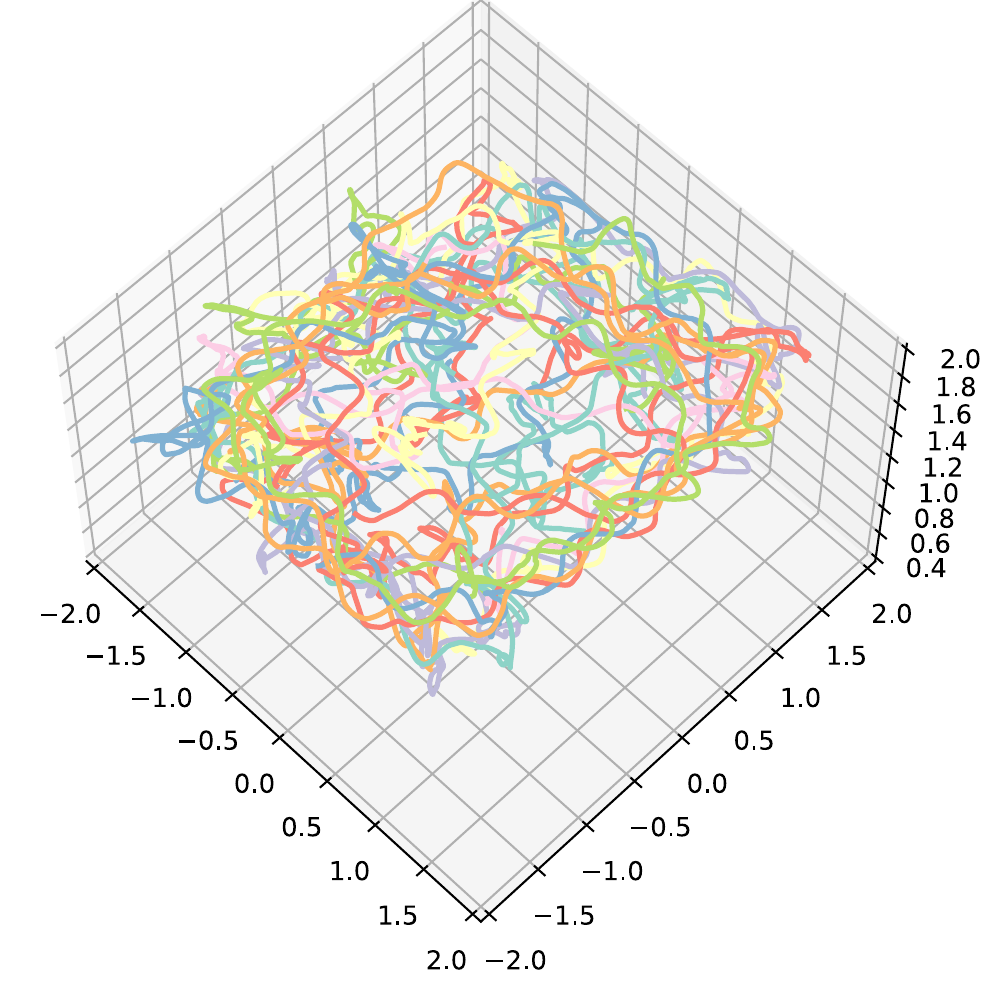}
    }
    \hspace{-10pt}
    \subcaptionbox*{Subject 28}{
        \includegraphics[width=0.05\textwidth]{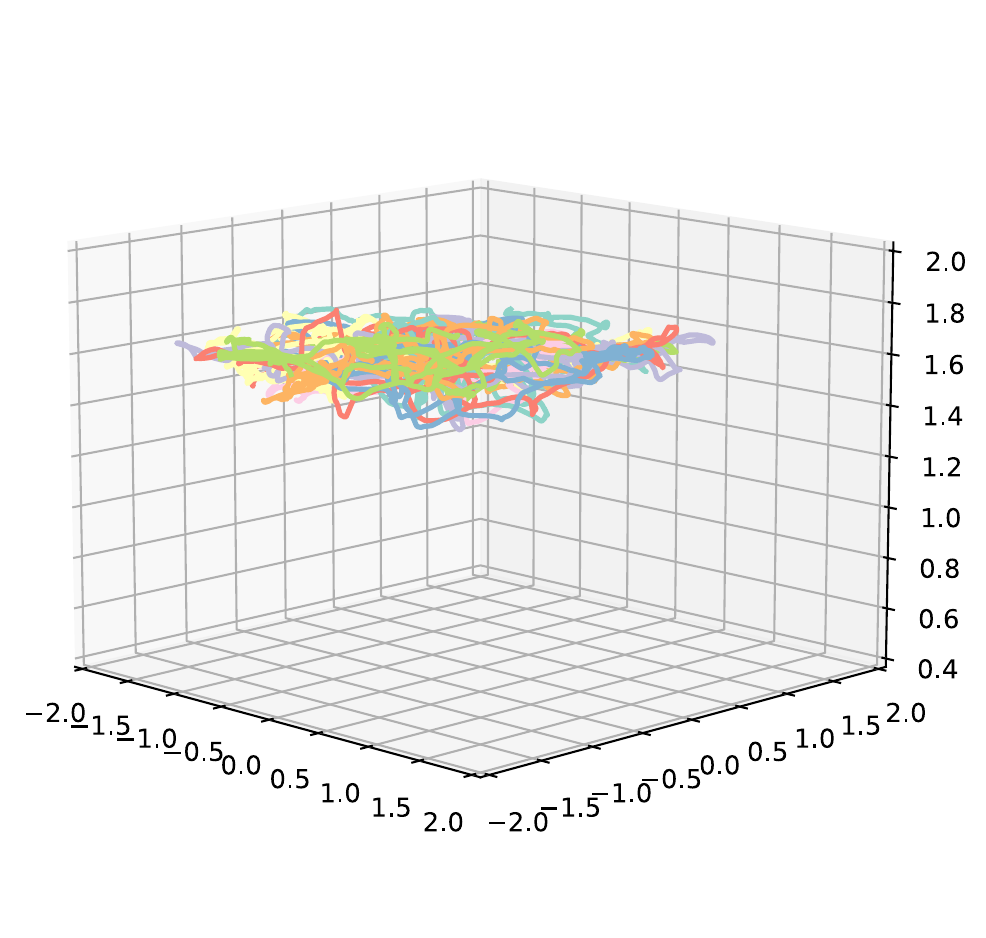}
        \includegraphics[width=0.05\textwidth]{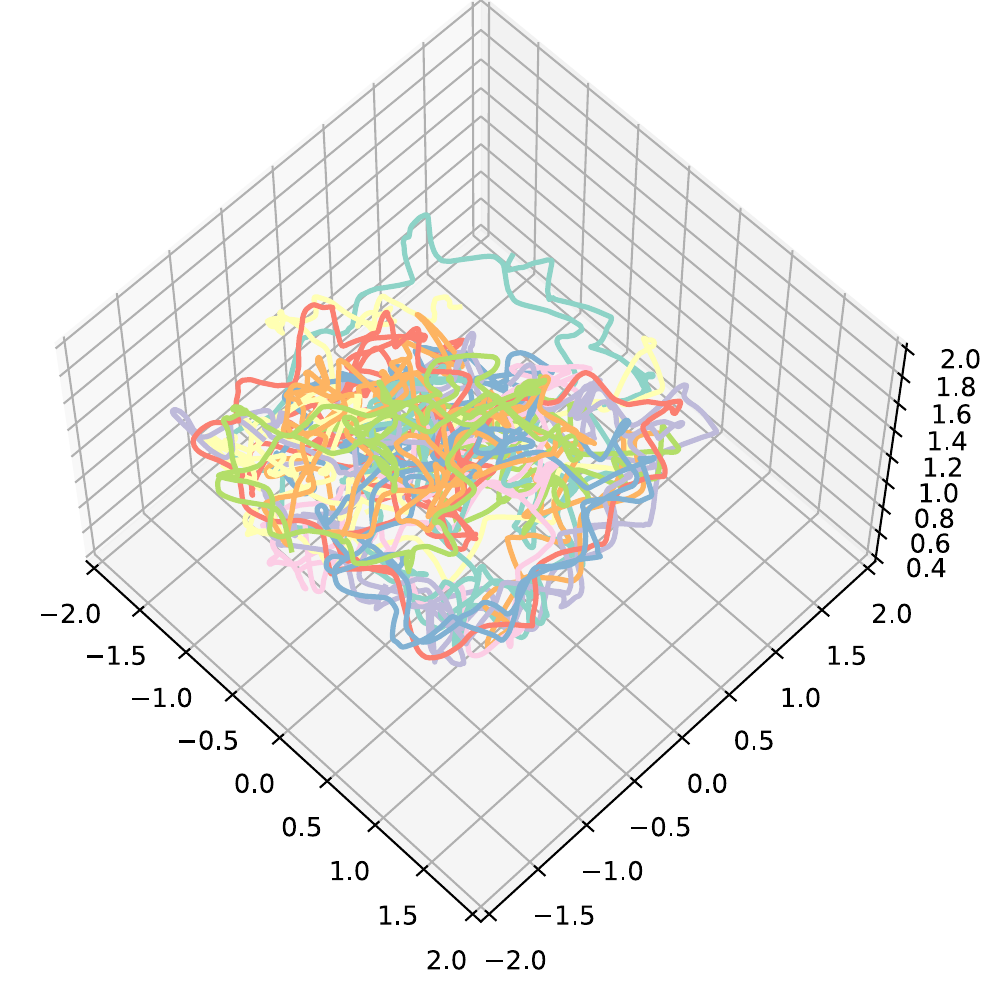}
    }

    \subcaptionbox*{Subject 29}{
        \includegraphics[width=0.05\textwidth]{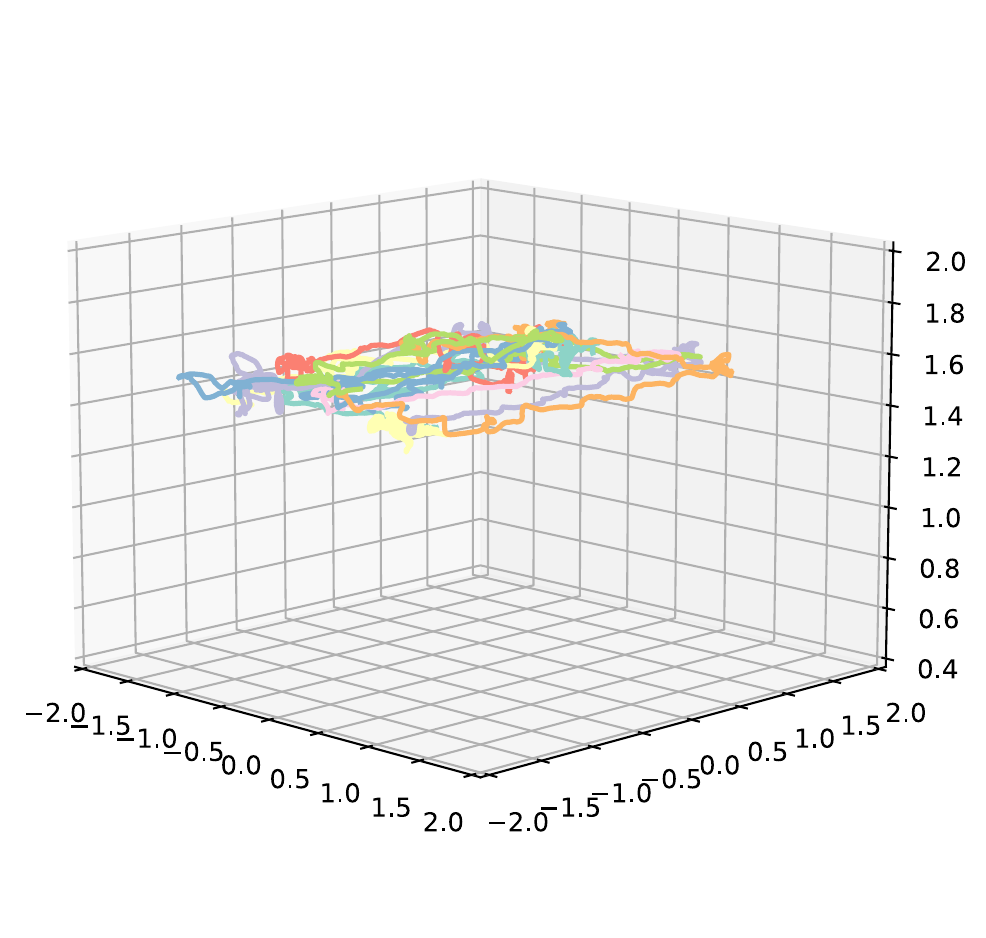}
        \includegraphics[width=0.05\textwidth]{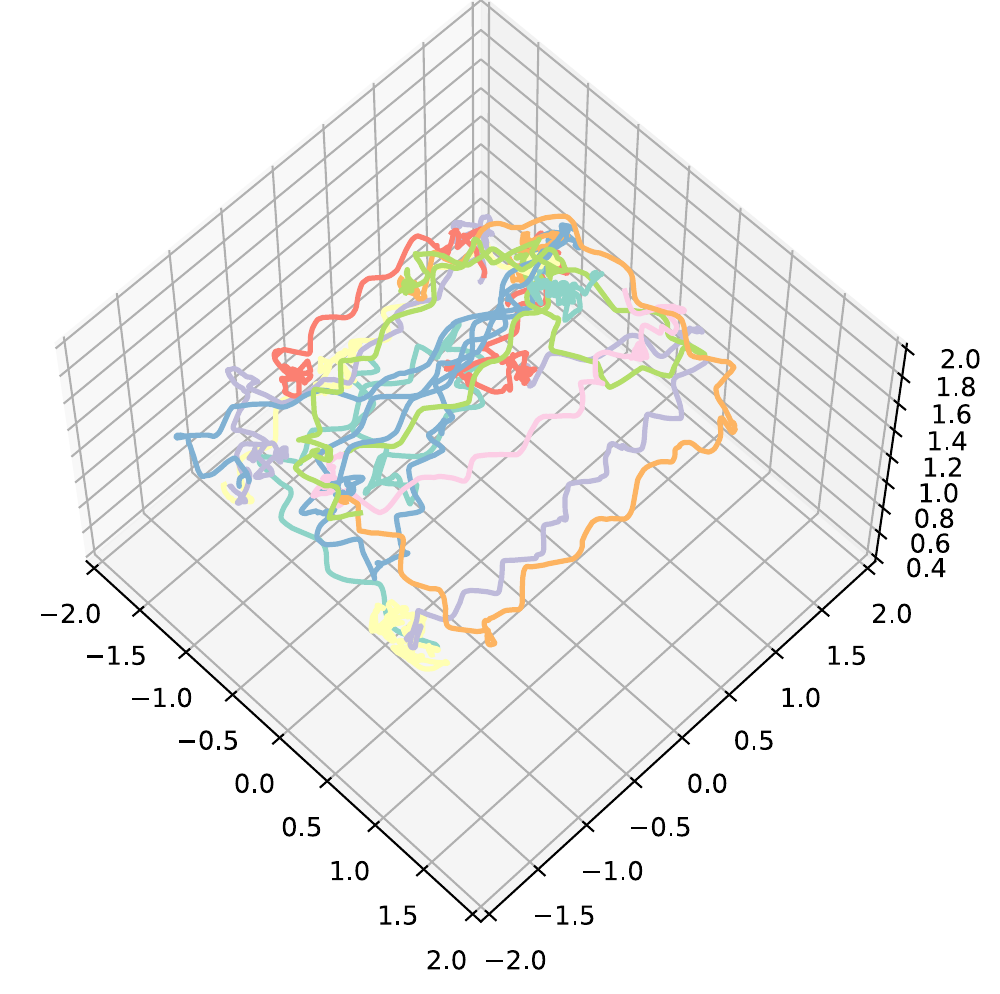}
    }
    \hspace{-10pt}
    \subcaptionbox*{Subject 30}{
        \includegraphics[width=0.05\textwidth]{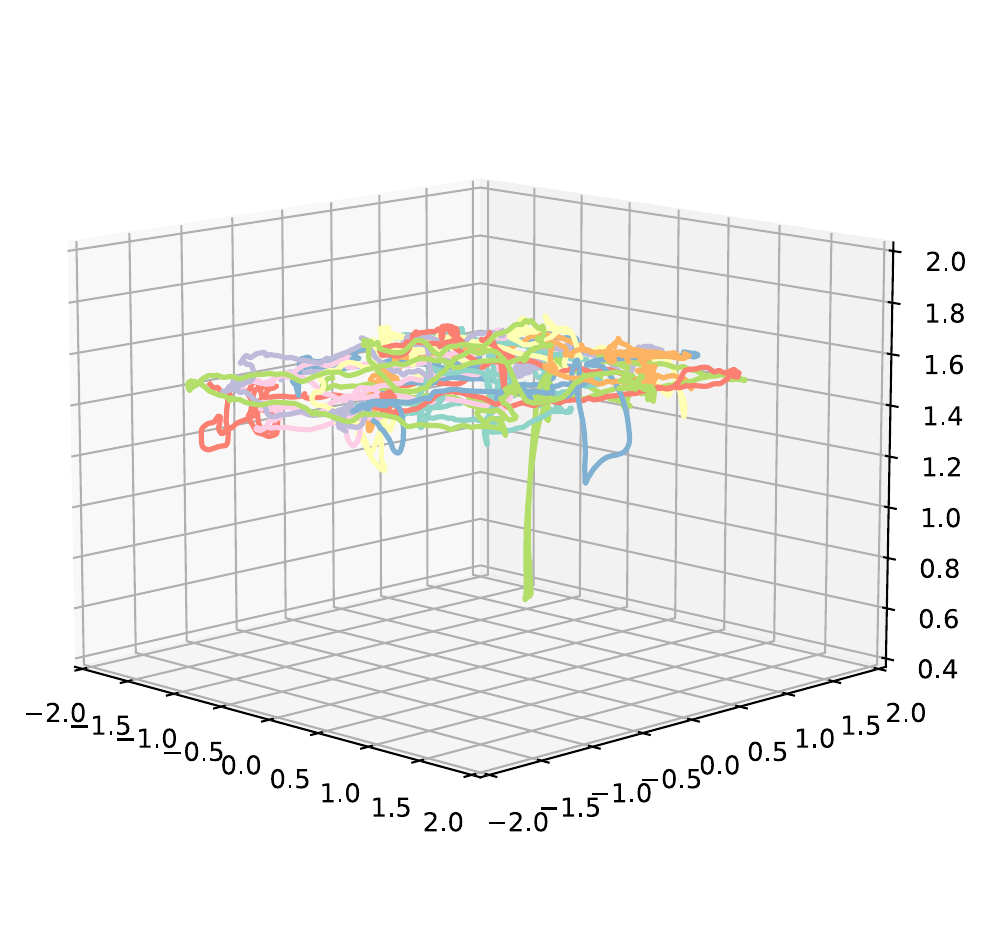}
        \includegraphics[width=0.05\textwidth]{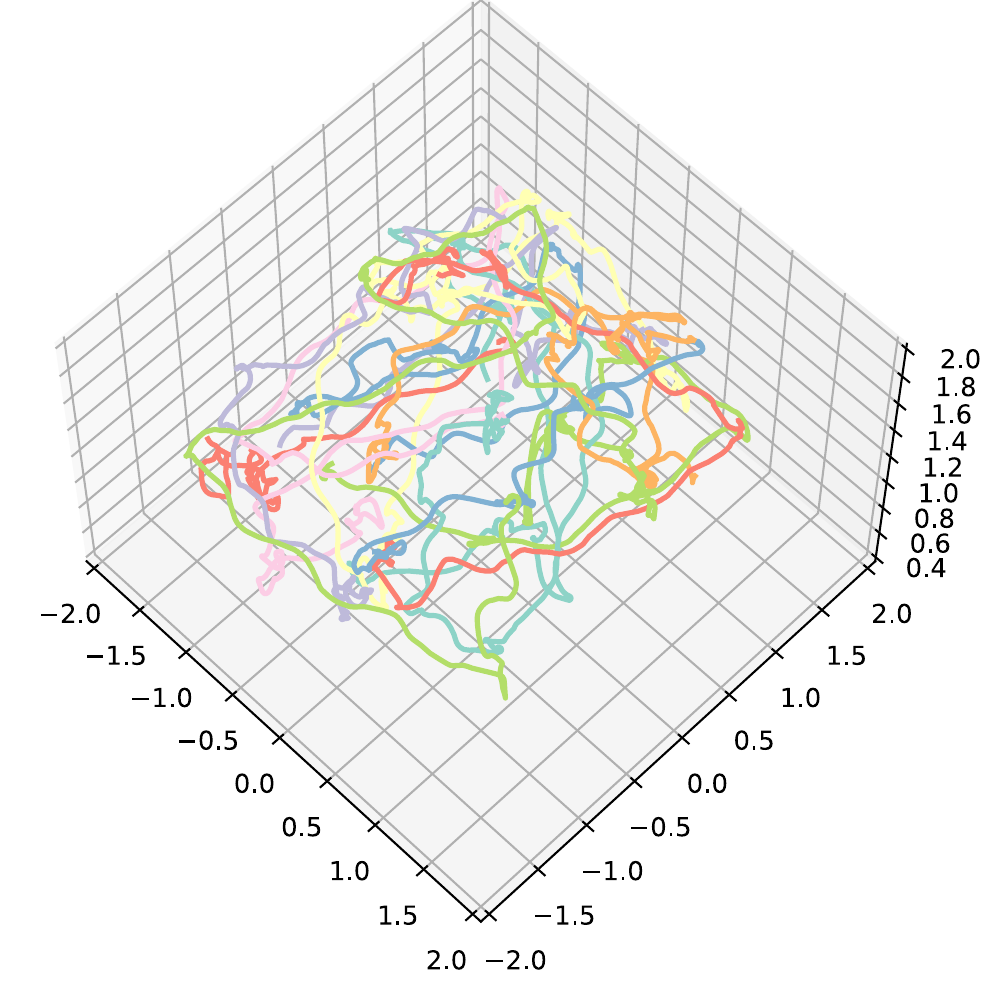}
    }
    \hspace{-10pt}
    \subcaptionbox*{Subject 31}{
        \includegraphics[width=0.05\textwidth]{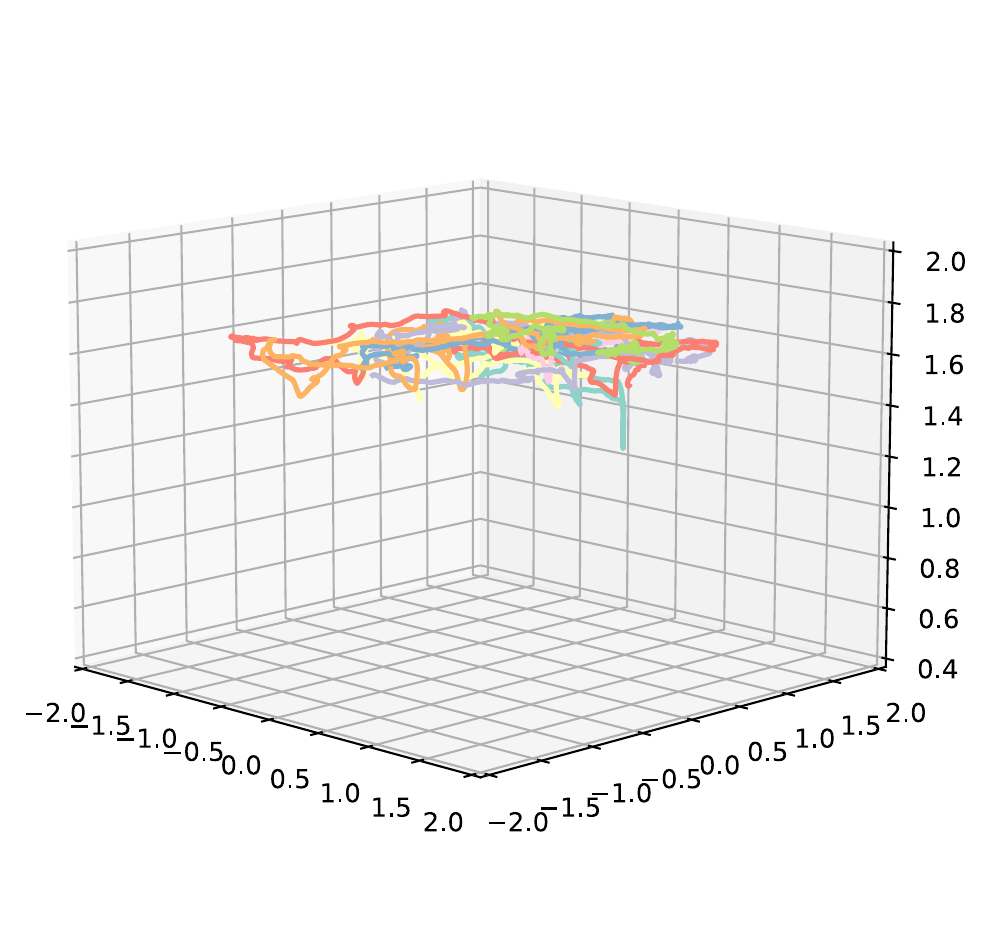}
        \includegraphics[width=0.05\textwidth]{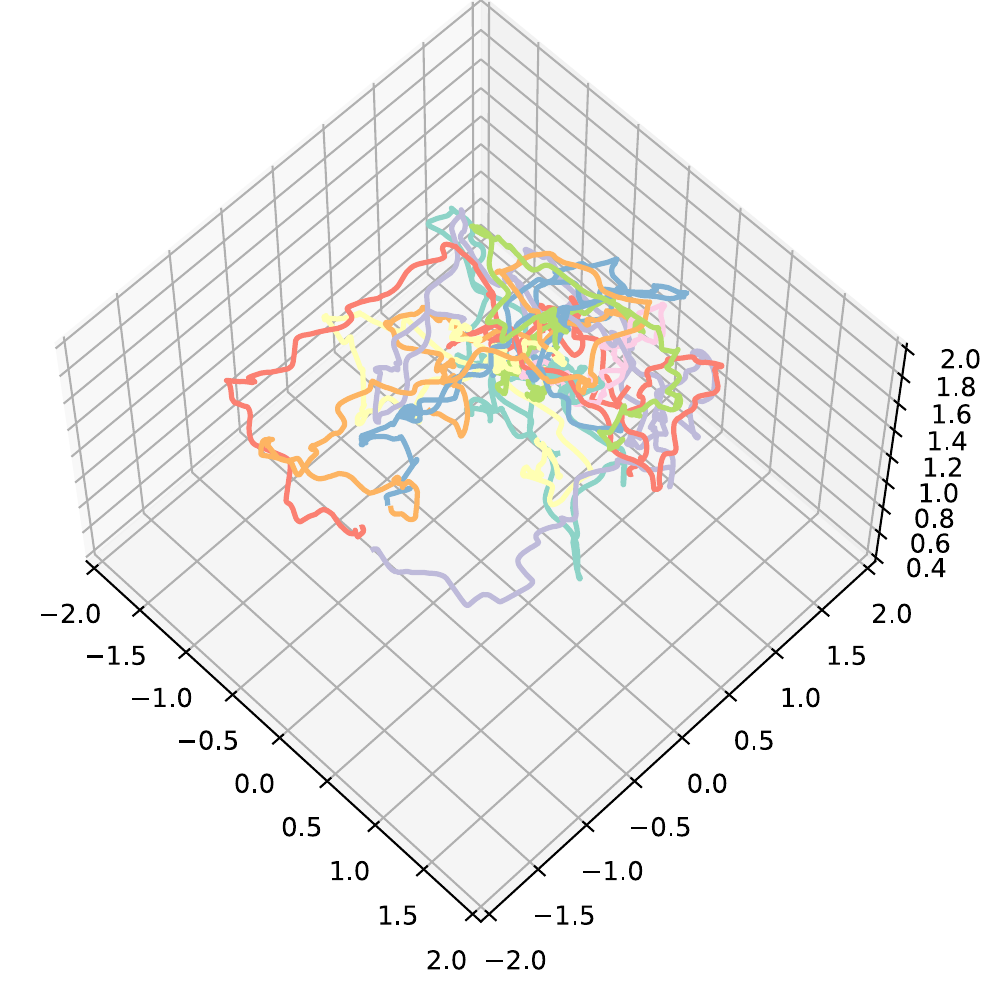}
    }
    \hspace{-10pt}
    \subcaptionbox*{Subject 32}{
        \includegraphics[width=0.05\textwidth]{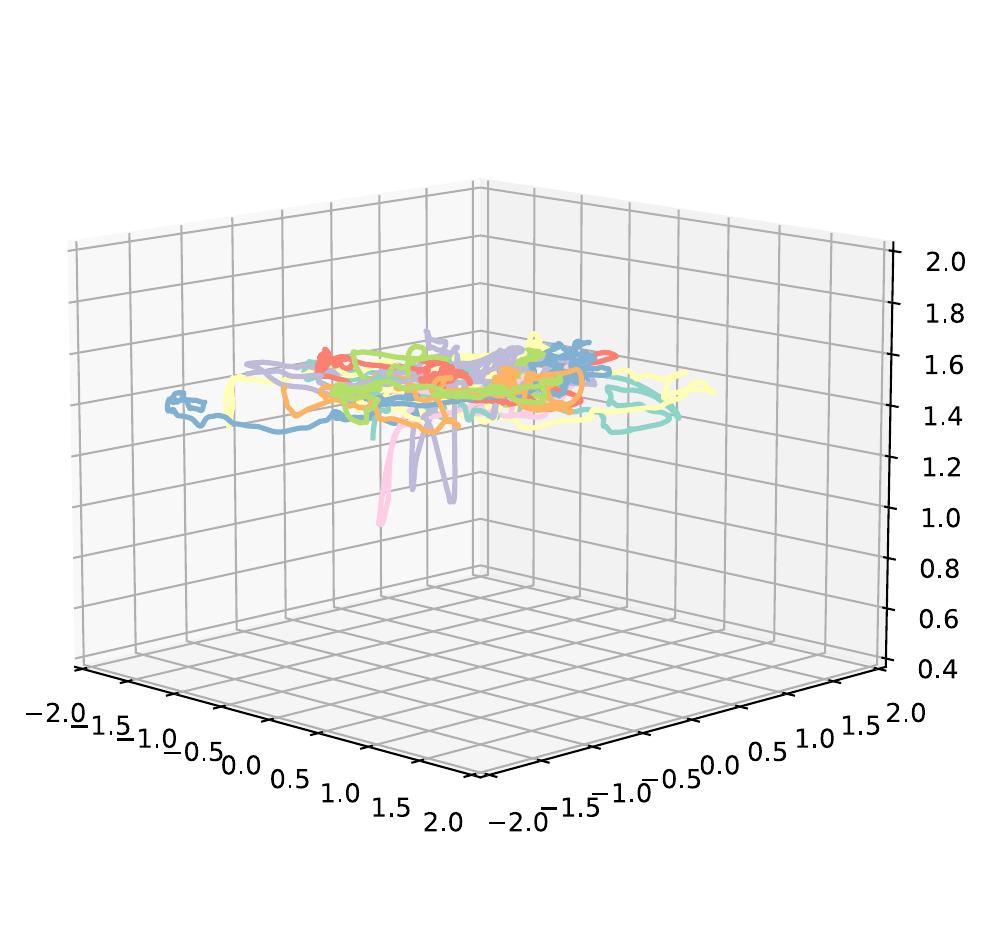}
        \includegraphics[width=0.05\textwidth]{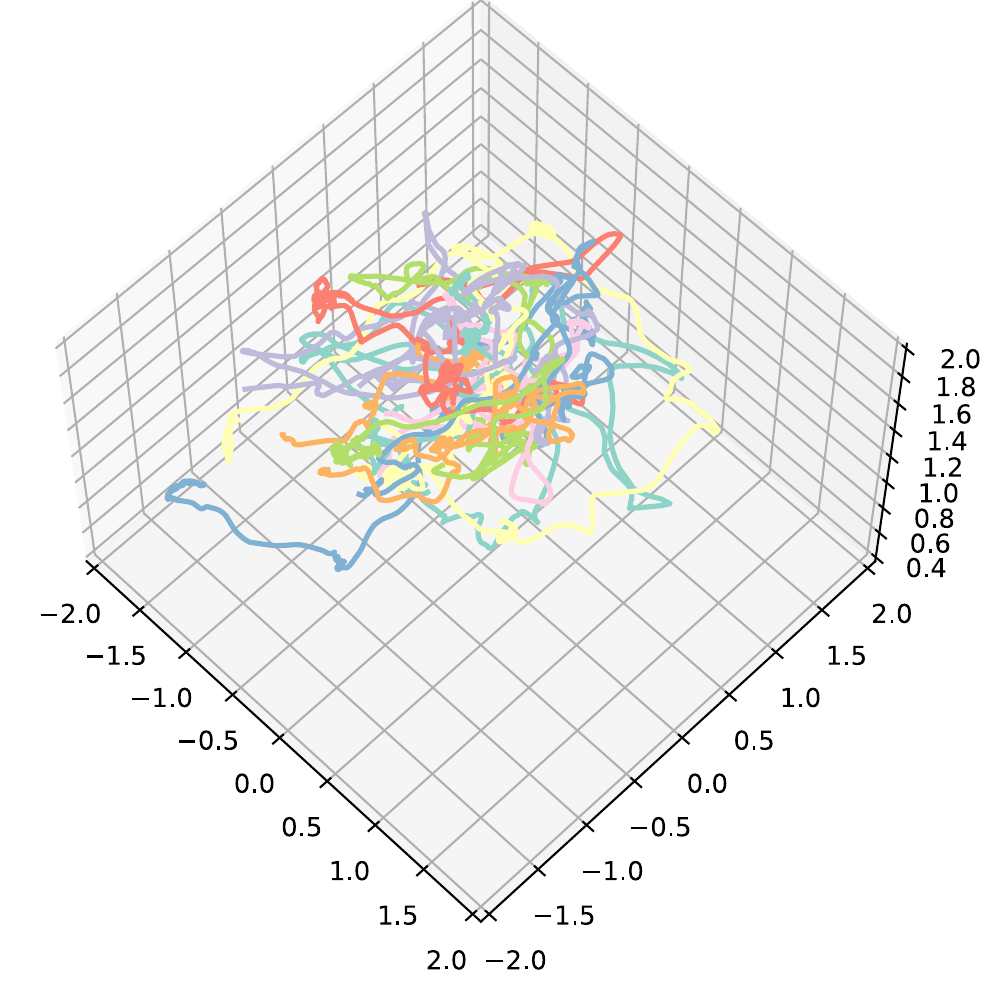}
    }
    \caption{The collected viewport trajectories from 32 subjects.}
    \label{fig:trajs}
\end{figure}

A total of 32 subjects, aged between 18 and 25 with an equal gender distribution, viewed these volumetric scenes for 1 minute each in a controlled 4m$\times$4m area. The dataset is split into training and testing sets, with 12 scenes and their corresponding viewport data forming the training set and the remaining 3 scenes used for testing, as shown in Figure~\ref{fig:thumbnail}. Additionally, Figure~\ref{fig:trajs} displays the viewport trajectories collected from the subjects.

\subsubsection{Dataset Capture}

User viewport data were captured using the HTC VIVE Pro Eye HMD. We developed a custom Unreal Engine 5 (UE5) project to record viewport data from the HMD at a rate of 30 frames per second (every 1/30 seconds).

\subsection{Implementation Details}

During training, the CVP and DBA modules were trained separately before integration into the GSStream system. Key implementation details are as follows:

\subsubsection{Collaborative Viewport Prediction}

The CVP module was set with a prediction horizon $H=30$. Both viewport and user embeddings were 16-dimensional. The attention modules and the integrated iTransformer each use 1 head. The learning rate is set to $1 \times 10^{-5}$.

\subsubsection{DRL-based Bitrate Adaptation}

In the DBA module, the temporal discount factor $\gamma$ is 0.9. The reward function weights are $\omega_1=\omega_2=10$ and $\omega_3=1$. Each minibatch contains 60 samples, with a soft update factor $\tau=0.005$. The time slot duration $\Delta t$ is 1 second.

\subsubsection{Bandwidth Trace Simulation}

To mimic a real-world streaming condition, we utilize bandwidth traces from cellular networks~\cite{vanderHooft2016}. We select 10 traces and scale them to average bandwidths of 40Mbps, 80Mbps, and 120Mbps. During training, bandwidth conditions are randomly selected from these scaled traces to ensure robustness across different network scenarios.

\subsection{Baselines}

To evaluate GSStream's performance, we compare it against two types of baseline methods.

\subsubsection{Baselines for Volumetric Video Streaming}

The first set of baselines includes prominent methods in volumetric video streaming, adapted for static 3DGS scene streaming:

\begin{itemize}
    \item \textbf{ViVo~\cite{han2020vivo}:} ViVo is a rule-based method that selects tile representations based on predicted user viewports. For static scenes, we masked previously selected representations to prevent duplicate transmissions. Originally supporting 5 quality levels, we interpolated to accommodate 6 levels in our setup.
    
    \item \textbf{CaV3~\cite{liu2023cav3}:} CaV3 uses a cache-assisted strategy for efficient streaming with effective tile caching and eviction. For static scenes, we retained only the caching mechanism to focus on tile selection based on priority, ensuring fair comparison with GSStream.
\end{itemize}

\subsubsection{Baseline for 3DGS Scene Streaming}

Due to the limited research on 3DGS scene streaming, we compared GSStream with the open-source 3DGS web viewer GS3D:

\begin{itemize}
    \item \textbf{GS3D~\cite{kellogg2023gaussiansplats3d}:} GS3D streams representations by sorting 3D Gaussians based on their distance from the world origin, prioritizing closer tiles. In a tile-based context, this means selecting tiles closer to the origin first. Unlike GSStream, GS3D relies solely on distance without incorporating viewport predictions or adaptive bitrate strategies.
\end{itemize}

\subsection{Results}

\begin{figure*}[htbp]
    \subcaptionbox*{}{
    \includegraphics[width=0.3304\textwidth]{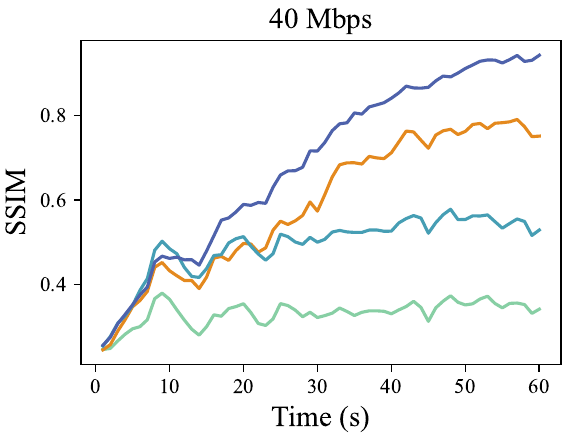}
    }
    \subcaptionbox*{}{
    \includegraphics[width=0.3098\textwidth]{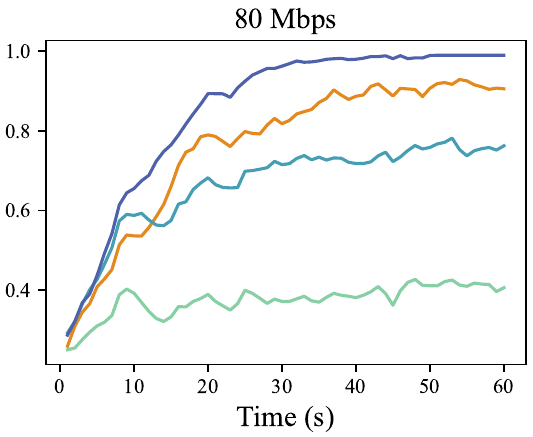}
    }
    \subcaptionbox*{}{
    \includegraphics[width=0.3098\textwidth]{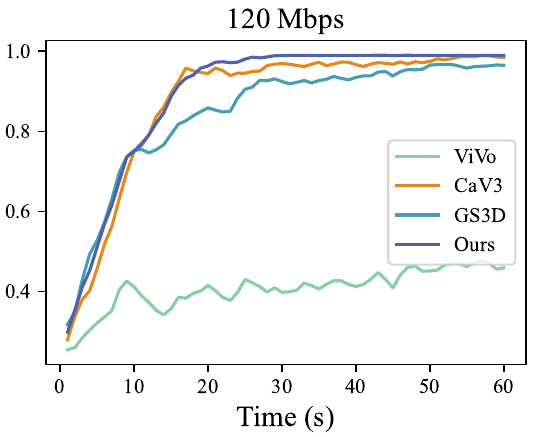}
    }
    \caption{The temporal viewport SSIM of the four volumetric scene streaming systems at each time slot $t$, under three typical network conditions.}
    \label{fig:vssim}
\end{figure*}
\begin{figure*}[htbp]
    \subcaptionbox*{}{
    \includegraphics[width=0.3304\textwidth]{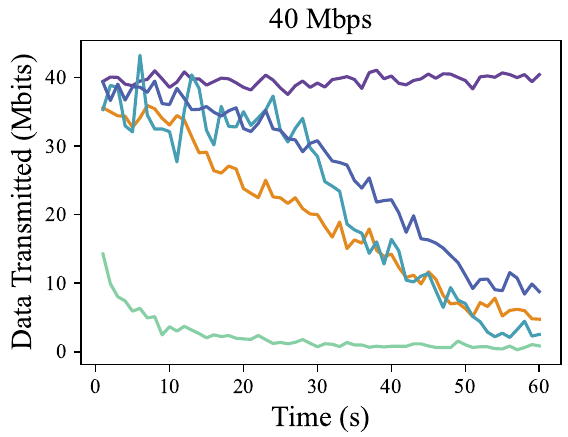}
    }
    \subcaptionbox*{}{
    \includegraphics[width=0.3098\textwidth]{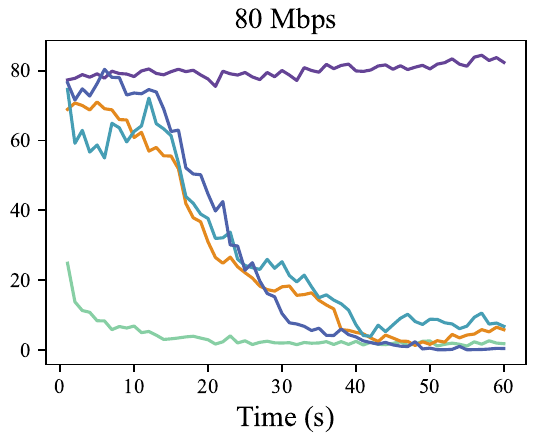}
    }
    \subcaptionbox*{}{
    \includegraphics[width=0.3098\textwidth]{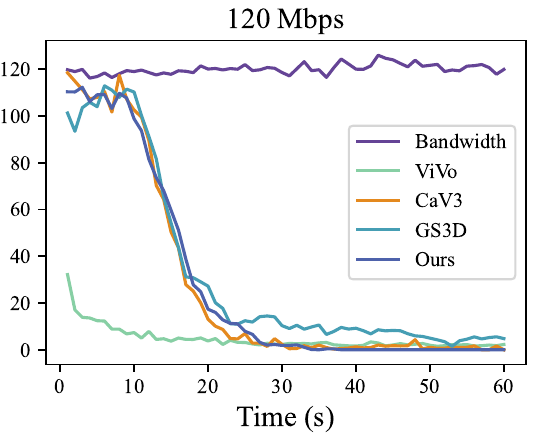}
    }
    \caption{The accumulative transmitted data of the four volumetric scene streaming systems at each time slot $t$, under three typical network conditions.}
    \label{fig:datautil}
\end{figure*}

\begin{table}[htbp]
    \centering
    \begin{tabular}{c|cccc}
    \toprule
        \multirow{2}*{Scene} & \multicolumn{4}{c}{Average Viewport SSIM $\uparrow$} \\
        ~ & System & 40Mbps & 80Mbps & 120Mbps \\
        \midrule
        \multirow{4}*{\makecell[c]{chtf\\187MB}} & ViVo & $0.10\pm0.08$ & $0.15\pm0.12$ & $0.18\pm0.13$\\
        ~ & CaV3 & $0.56\pm0.43$ & $0.77\pm0.38$ & $0.84\pm0.34$\\
        ~ & GS3D & $0.61\pm0.31$ & $0.77\pm0.31$ & $0.85\pm0.28$\\
        ~ & \textbf{Ours} & $\mathbf{0.64\pm0.37}$ & $\mathbf{0.81\pm0.32}$ & $\mathbf{0.84\pm0.31}$\\
        \midrule
        \multirow{4}*{\makecell[c]{garden\\224MB}} & ViVo & $0.49\pm0.33$ & $0.54\pm0.32$ & $0.58\pm0.31$\\
        ~ & CaV3 & $0.67\pm0.31$ & $0.79\pm0.29$ & $0.89\pm0.23$\\
        ~ & GS3D & $0.66\pm0.30$ & $0.74\pm0.29$ & $0.79\pm0.27$\\
        ~ & \textbf{Ours} & $\mathbf{0.72\pm0.29}$ & $\mathbf{0.87\pm0.23}$ & $\mathbf{0.92\pm0.18}$\\
        \midrule
        \multirow{4}*{\makecell[c]{stump\\224MB}} & ViVo & $0.40\pm0.32$	& $0.43\pm0.32$ & $0.47\pm0.31$\\
        ~ & CaV3 & $0.54\pm0.34$ & $0.72\pm0.33$ & $0.90\pm0.23$\\
        ~ & GS3D & $0.60\pm0.29$ & $0.75\pm0.29$ & $0.82\pm0.26$\\
        ~ & \textbf{Ours} & $\mathbf{0.71\pm0.29}$ & $\mathbf{0.89\pm0.22}$ & $\mathbf{0.91\pm0.21}$\\
    \bottomrule
    \end{tabular}
    \caption{The average viewport SSIM of different methods.}
    \label{tab:AvSSIM}
\end{table}

\subsubsection{Comparison on Viewport Prediction Task}

\begin{figure}
    \subcaptionbox{The MAE of the future position prediction.}{
        \includegraphics[width=0.45\linewidth]{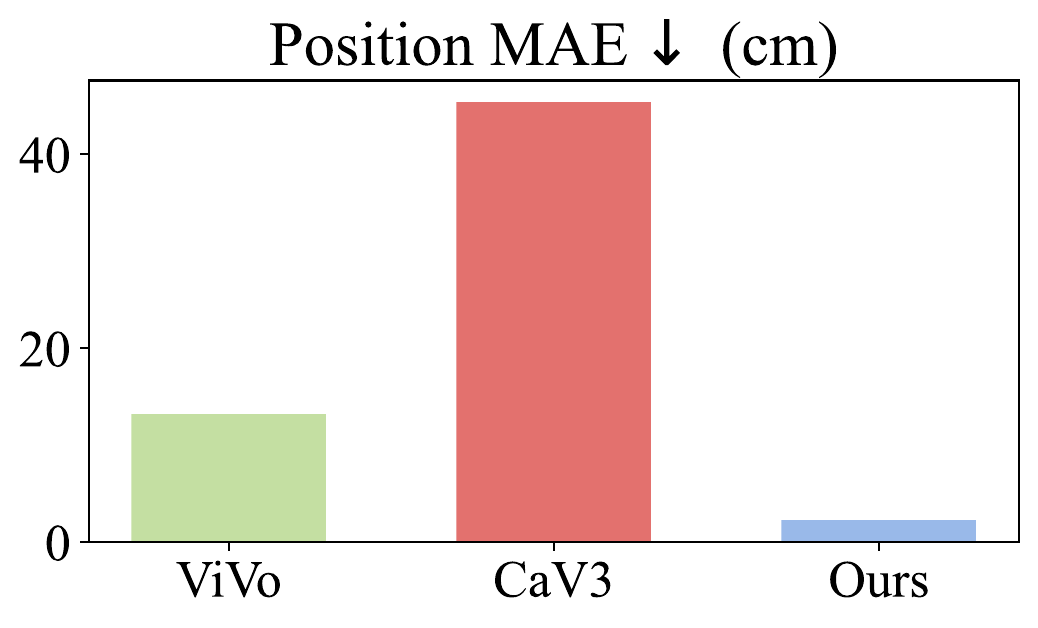}
    }
    \subcaptionbox{The MAE of the future rotation prediction.}{
        \includegraphics[width=0.45\linewidth]{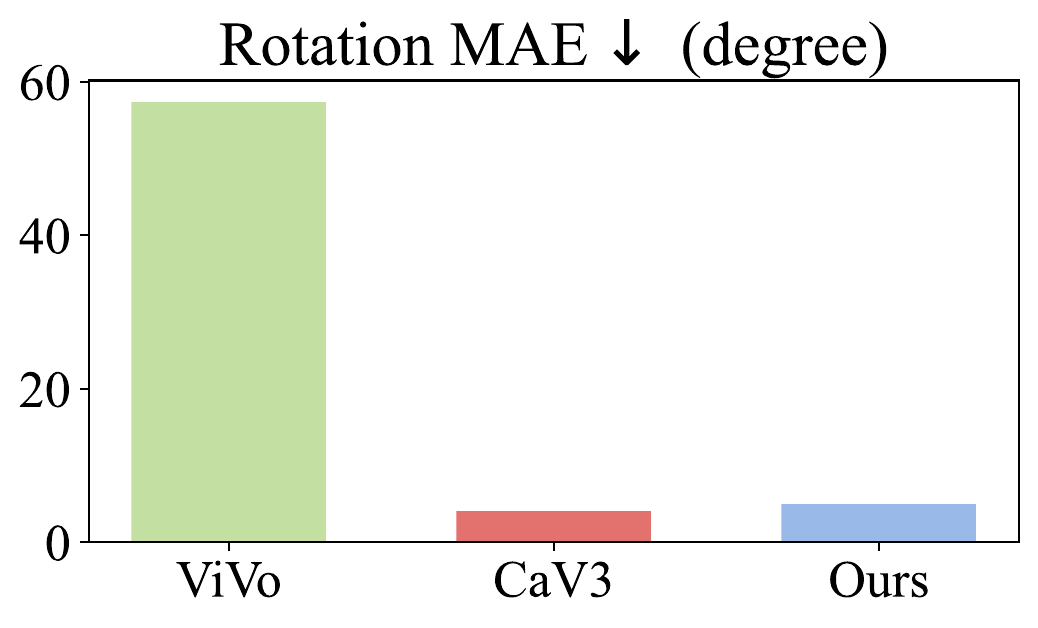}
    }
    \caption{Results of the comparison experiments on viewport prediction task. Note that the compared baselines have different lengths in the prediction time window; therefore, the experimental results have limited significance.}
    \label{fig:vpCompExp}
\end{figure}

We compared the viewport prediction accuracy (measured in MAE) of the proposed CVP module against ViVo~\cite{han2020vivo}, CaV3~\cite{liu2023cav3} and GS3D~\cite{kellogg2023gaussiansplats3d}. As illustrated in Figure~\ref{fig:vpCompExp}, the proposed method achieves SOTA among the compared representative systems.

\subsubsection{Comparison on Bitrate Adaptation Task}

We compare GSStream with three representative systems: ViVo~\cite{han2020vivo}, CaV3~\cite{liu2023cav3} and GS3D~\cite{kellogg2023gaussiansplats3d} with the above-mentioned bandwidth simulation. To evaluate the performance of each streaming system comprehensively, we construct comparison experiments from three perspectives.

\textbf{Average Viewport SSIM:} 
As quantified in Table~\ref{tab:AvSSIM}, the proposed GSStream framework achieves 118.9\%, 9.4\%, and 10.9\% higher mean viewport SSIM than ViVo, CaV3, and GS3D respectively, where viewport SSIM measures structural similarity between original and transmitted volumetric content. This superiority derives from overcoming three key limitations: ViVo's rigid rule-based transmission, CaV3's binary quality selection restricting adaptation to bandwidth extremes, and GS3D's origin-anchored spatial prioritization that degrades early-stage quality during viewpoint exploration. By synergizing continuous quality spectrum selection with temporal-aware adaptation, GSStream demonstrates comprehensive advantages in both perceptual quality maintenance and adaptive efficiency, conclusively outperforming all baseline methods for progressive volumetric streaming.

\textbf{Temporal Visual Quality:} Figure~\ref{fig:vssim} depicts the temporal visual quality evolution across four streaming systems under stochastic network dynamics, revealing three critical observations. (1) Systemic adaptability: While all systems exhibit progressive quality enhancement over time under bandwidth-constrained scenarios, ViVo's static prioritization rules constrain its responsiveness to dynamic network conditions. (2) Phase-sensitive divergence: Though CaV3 temporarily achieves marginal SSIM advantages during initial exploration phases through RL-based discretized adaptation, it experiences early-stage quality fluctuation from binary quality decisions. (3) Spatial-temporal optimization: GS3D's origin-biased transmission strategy unintentionally penalizes mid-stage viewing experiences during coordinate-distant viewpoint transitions, whereas GSStream continuously refines reconstruction fidelity through temporal prediction-accelerated LOD switching, ultimately achieving 25.9\%/80.3\% higher late-stage visual detail retention than CaV3/GS3D respectively. These empirical measurements confirm our framework's capability to synergize structural preservation with adaptive streaming efficiency, fundamentally overcoming the spatial-temporal decoupling limitations inherent in baseline architectures.

\textbf{Bandwidth Throughput Usage:} Figure~\ref{fig:datautil} reveals distinct bandwidth utilization patterns across four volumetric streaming architectures. The results show that the proposed GSStream system consistently outperforms ViVo, CaV3, and GS3D in terms of data transmission efficiency and bandwidth utilization. While other methods like ViVo and CaV3 exhibit significant fluctuations in throughput and fail to effectively utilize available bandwidth, GSStream maintains stable and high data throughput across all tested bandwidth conditions. In contrast, GS3D, although improving throughput to some extent, still lags behind Ours in efficiently utilizing the available bandwidth. Overall, Ours demonstrates superior performance in static 3D scene streaming by ensuring stable and optimal bandwidth utilization.

\subsubsection{Ablation Experiments}

\begin{figure}
    \subcaptionbox{The MAE of the future position prediction.}{
        \includegraphics[width=0.45\linewidth]{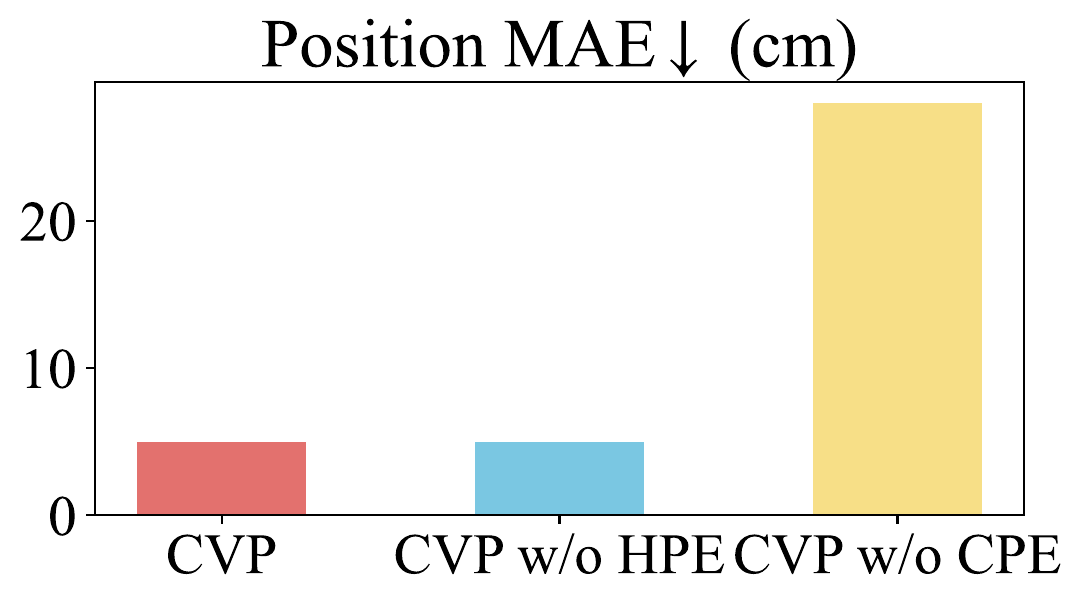}
    }
    \subcaptionbox{The MAE of the future rotation prediction.}{
        \includegraphics[width=0.45\linewidth]{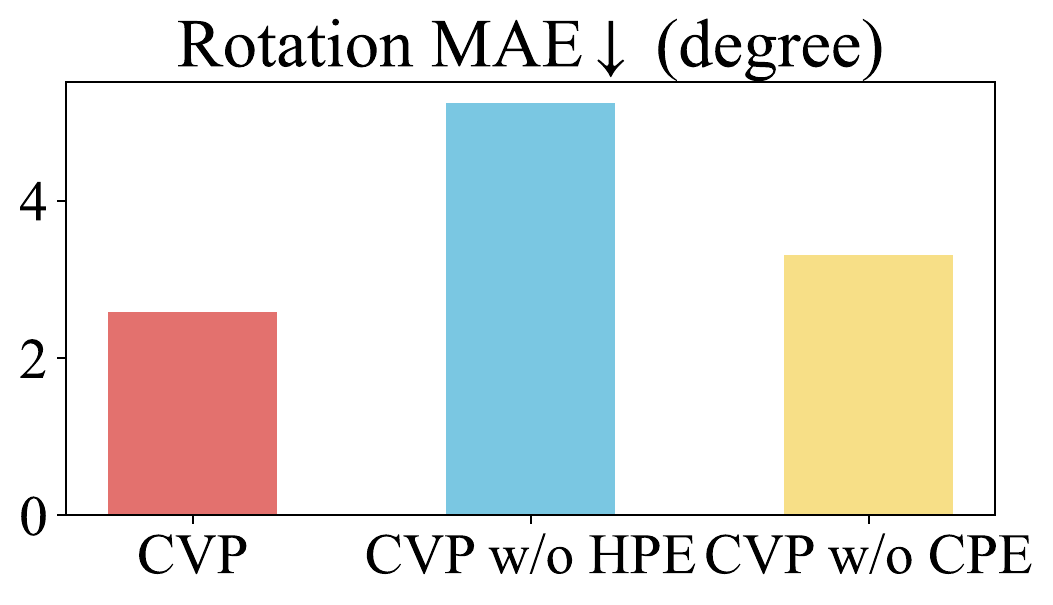}
    }
    \caption{Results of the ablation experiments.}
    \label{fig:vpAblaExp}
\end{figure}

We construct ablation experiments for our proposed CVP module by masking the historical prior information and the collaborative prior information respectively, denoted as \textbf{CVP w/o HPE} and \textbf{CVP w/o CPE} in Figure~\ref{fig:vpAblaExp}, where \textbf{CVP} denotes the proposed collaborative viewport prediction algorithm. As we can see, \textbf{CVP} fuses the two signals and achieves the lowest average MAE in both position and rotation prediction. Without the collaborative priors, \textbf{CVP w/o HPE} still achieves higher performance than \textbf{CVP w/o CPE} in position prediction, which reflects that the viewing routes of different users when watching volumetric scenes are affected by their inherent habits significantly. On the other hand, without historical priors, \textbf{CVP w/o HPE} performs weaker than \textbf{CVP w/o CPE} in rotation prediction, which indicates that the rotation angles changed within the time window $H$ have a stronger sequential nature. However, the proposed collaborative viewport prediction algorithm succeeded in fusing all two signals, achieving the best performance among all three settings.

\section{Limitations}

\textbf{Dynamic media support:} The proposed GSStream system currently only considers the streaming of volumetric scenes. However, with the rapid development of virtual reality technology, the topic of volumetric video streaming is also widely developed. Recently, pioneer works have explored the possibility of using 3DGS-related technologies as a novel form of volumetric video implementation, and have also made amazing breakthroughs~\cite{sun20243dgstream,xu20234k4d,wu20234d,10757420}. In the future, one of our key work directions is to expand GSStream to support dynamic volumetric data streaming.

\textbf{More optimization:} Recently, 3DGS has attracted a lot of interest from researchers. Numerous cutting-edge works are exploring whether a more compact representation of 3DGS can be given to save the data scale of volumetric scene representation~\cite{lee2023compact,morgenstern2023compact}. Besides, some researchers are committed to developing codecs for 3DGS data to generate representations in different quality levels, to achieve efficient and progressive encoding and decoding~\cite{navaneet2023compact3d,girish2023eagles} as developed in the field of point cloud compression~\cite{9496619}. However, in our proposed GSStream system, representations are downsampled from the original tiles. One of our future directions is to introduce efficient 3DGS codecs, achieving greater savings on bandwidth throughput.

\section{Conclusion}

In this paper, we propose GSStream, a novel 3DGS-based volumetric scene-streaming system. We utilize 3DGS as a new form of volumetric scene representation, greatly enhancing the display visual quality, compared with existing volumetric scene streaming systems. The proposed GSStream system integrates a CVP module and a DBA module. The CVP module utilizes the user-specific embeddings learned from multiple users collaboratively alongside the historical priors learned from the historical viewport sequences, promoting the viewport prediction accuracy. The DBA module tackles the variability challenge of state and action space in the bitrate adaptation problem by introducing the SA and FP modules, then solves it via a DDPG-based algorithm. Experiments demonstrate the efficiency and effectiveness of the proposed volumetric scene streaming system among existing SOTA streaming systems.

{
\bibliographystyle{IEEEtran}
\bibliography{references}

\begin{IEEEbiography}[{\includegraphics[width=1in,height=1.25in,clip,keepaspectratio]{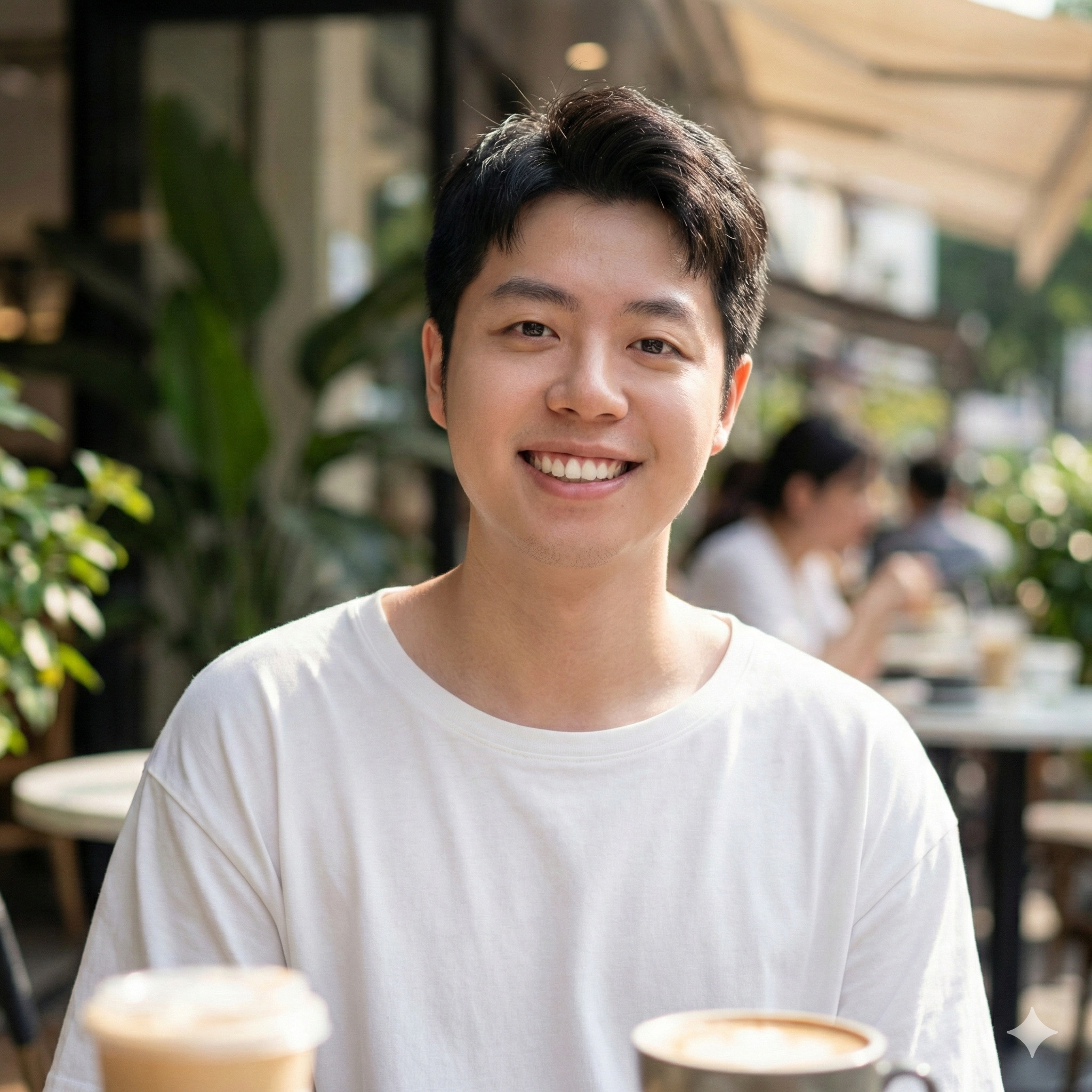}}]{Zhiye Tang} received the B.E. degree from Shenzhen University, Shenzhen, China, in 2022. He is currently working toward the M.S. degree with the College of Computer Science and Software Engineering, Shenzhen University, China. His research interests include compact representation and efficient transmission for 3D Gaussian Splatting.
\end{IEEEbiography}
\vfill{-0.5in}

\begin{IEEEbiography}[{\includegraphics[width=1in,height=1.25in,clip,keepaspectratio]{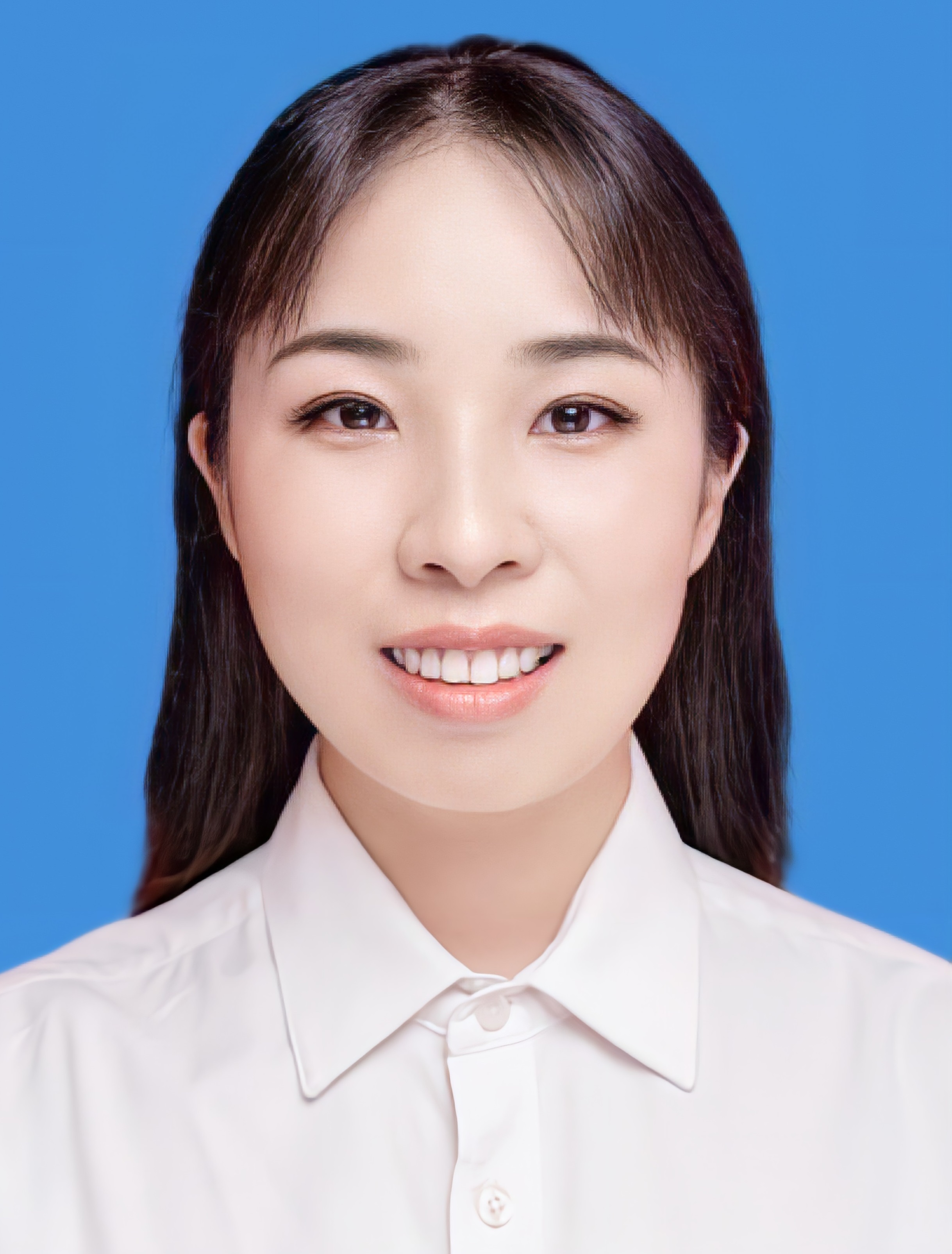}}]{Qiudan Zhang} received the B.E. and M.S. degrees in the College of Computer Science and Software Engineering from Shenzhen University, China in 2015 and 2018, respectively. She received her Ph.D. degree from the Department of Computer Science, City University of Hong Kong, China (Hong Kong SAR) in 2021. She is currently an Assistant Professor in the College of Computer Science and Software Engineering, Shenzhen University, China. Her research interests include computer vision, visual attention, 3D vision and deep learning.
\end{IEEEbiography}
\vspace{-9 mm}
\begin{IEEEbiography}[{\includegraphics[width=1in,height=1.25in,clip,keepaspectratio]{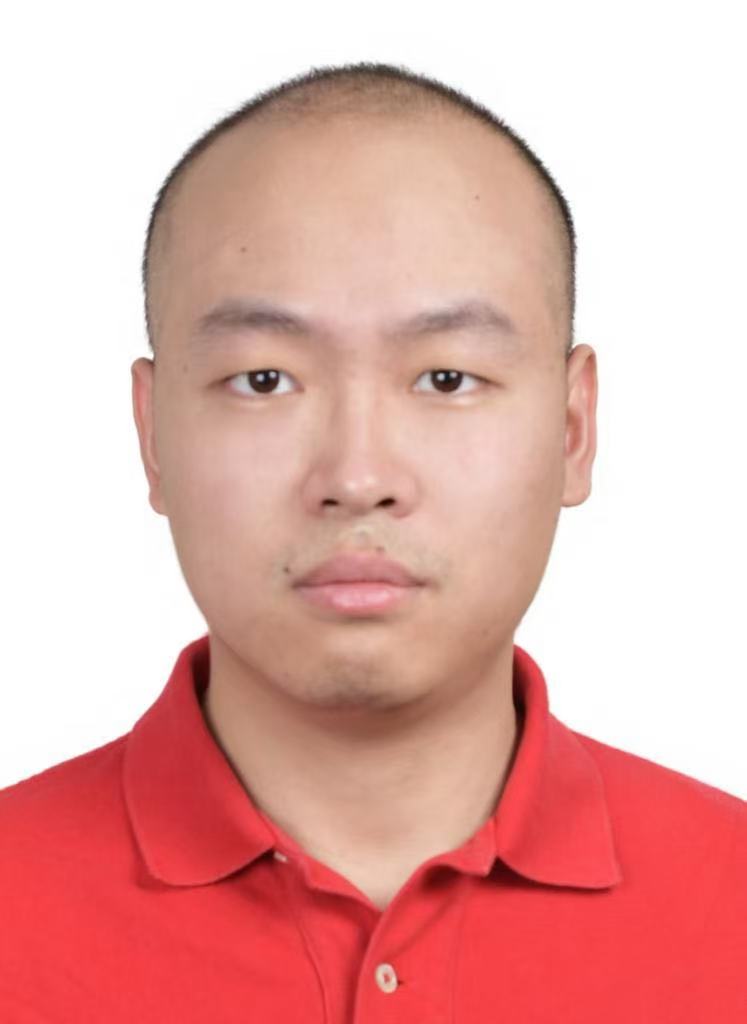}}]{Lei Zhang} (Member, IEEE) received the B.Eng. degree from the Advanced Class of Electronics and Information Engineering, Huazhong University of Science and Technology, Wuhan, China, in 2011, and the M.S. and Ph.D. degrees from Simon Fraser University, Burnaby, BC, Canada, in 2013 and 2019, respectively. He is currently an Assistant Professor with the College of Computer Science and Software Engineering, Shenzhen University, Shenzhen, China. His research interests include multimedia systems and applications, mobile cloud computing, edge computing, social networking, and the Internet of Things. He was the recipient of the
C.D. Nelson Memorial Graduate Scholarship (2013) and Best Paper Finalist at IEEE/ACM IWQoS (2016).
\end{IEEEbiography}
\vspace{-9 mm}
\begin{IEEEbiography}[{\includegraphics[width=1in,height=1.25in,clip,keepaspectratio]{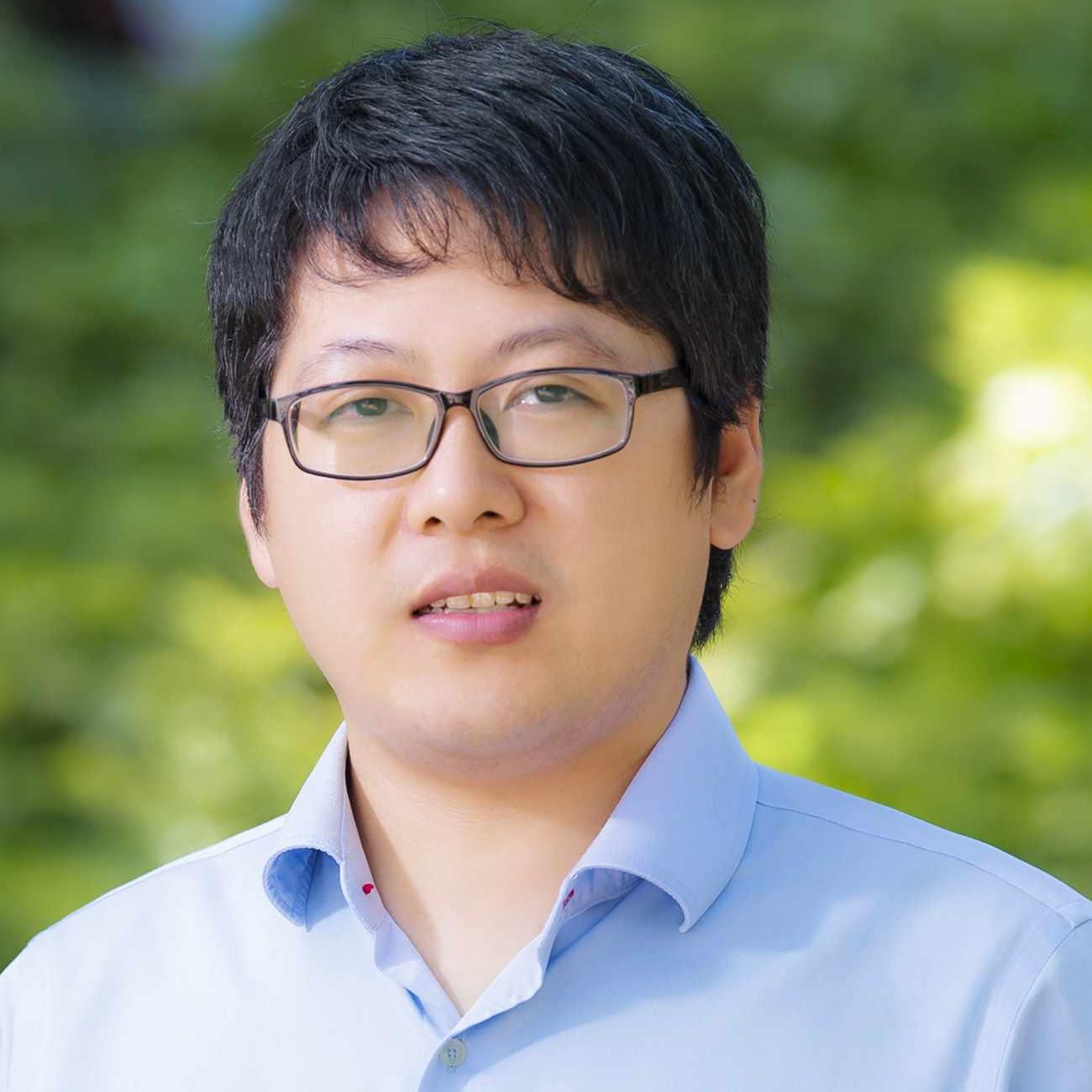}}]{Junhui Hou} (Senior member, IEEE) is an Associate Professor with the Department of Computer Science, City University of Hong Kong. He holds a B.Eng. degree in information engineering (Talented Students Program) from the South China University of Technology, Guangzhou, China (2009), an M.Eng. degree in signal and information processing from Northwestern Polytechnical University, Xi’an, China (2012), and a Ph.D. degree from the School of Electrical and Electronic Engineering, Nanyang Technological University, Singapore (2016). His research interests are multi-dimensional visual computing. Dr. Hou received the Early Career Award (3/381) from the Hong Kong Research Grants Council in 2018 and the NSFC Excellent Young Scientists Fund in 2024. He has served or is serving as an Associate Editor for IEEE Transactions on Visualization and Computer Graphics, IEEE Transactions on Image Processing, IEEE Transactions on Multimedia, and IEEE Transactions on Circuits and Systems for Video Technology.
\end{IEEEbiography}
\vspace{-9 mm}
\begin{IEEEbiography}[{\includegraphics[width=1in,height=1.25in,clip,keepaspectratio]{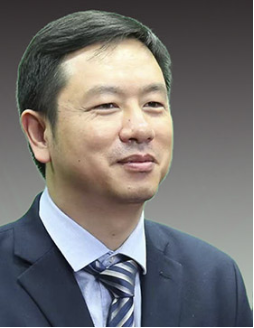}}]{You Yang} (M'10-SM'16) received the Ph.D. degree in computer science from the Institute of Computing Technology, Chinese Academy of Sciences, Beijing, China, in 2009. He worked as a Postdoctoral Fellow with the Automation Department, Tsinghua University, from 2009 to 2011. He was a Senior Research Scientist in Sumavision Research from 2011 to 2013. He currently heads the Department of Information Engineering, Huazhong University of Science and Technology, Wuhan, China. He has authored and co-authored more than 60 technical papers and authorized 17 patents. His research interests include three-dimensional (3D) vision system and its applications, including the multi-view imaging system, 3D/VR/AR content processing and broadcasting, human-machine interaction techniques, and interactive visual applications.

Dr. Yang has been an Academic Editor of the PLoS ONE since 2015, the Guest Editor of Neurocomputing in 2014, the Committee/TPC Member or Session Chair of over 30 international conferences, and Reviewers of 28 prestigious international journals from IEEE, ACM, OSA and other associations. He was elected to be a Fellow of the Institute of Engineering and Technology in 2018.
\end{IEEEbiography}
\vspace{-9 mm}
\begin{IEEEbiography}[{\includegraphics[width=1in,height=1.25in,clip,keepaspectratio]{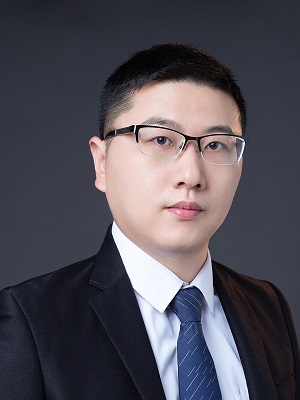}}]{Xu Wang}(M'15) received the B.S. degree from South China Normal University, Guangzhou, China, in 2007, and M.S. degree from Ningbo University, Ningbo, China, in 2010. He received his Ph.D. degree from the Department of Computer Science, City University of Hong Kong in 2014. In 2015, he joined the College of Computer Science and Software Engineering, Shenzhen University, where he is currently an Associate Professor. His research interests are video coding and 3D vision.
\end{IEEEbiography}

}
\newpage

\vfill

\end{document}